\documentclass[11pt,a4paper,logo]{googledeepmind}

\setleftlogo[120pt]{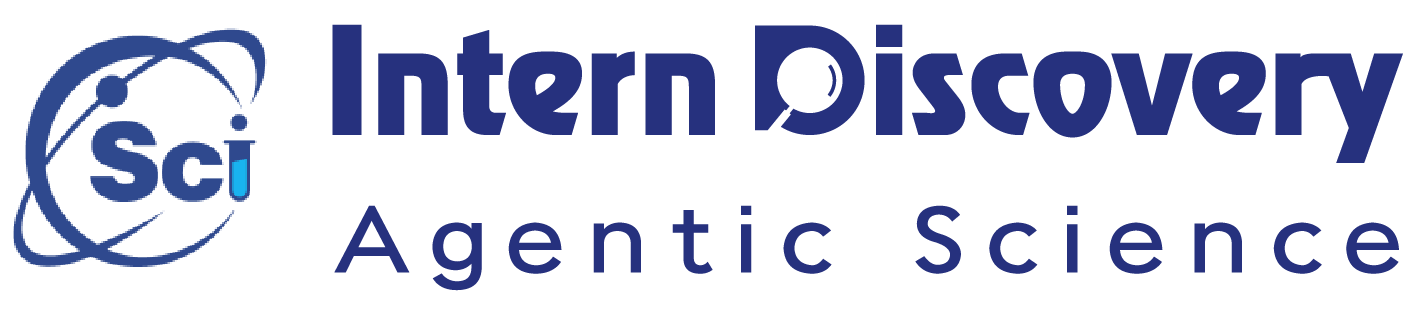}
\setrightlogo[180pt]{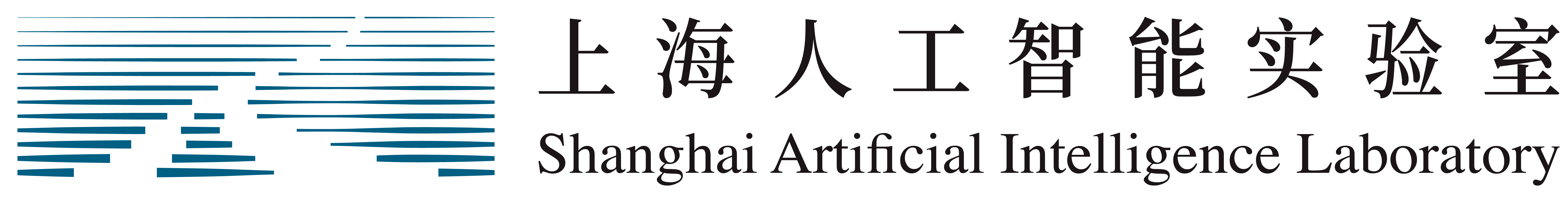}

\usepackage[T1]{fontenc}
\usepackage{pifont}
\usepackage[
    natbib=true,
    backend=biber,
    style=numeric, 
    sorting=none 
]{biblatex}
\AtEveryBibitem{\clearfield{month}}
\AtEveryBibitem{\clearfield{day}}
\usepackage{csquotes}
\usepackage{hyperref}
\usepackage{xcolor}
\usepackage{fontawesome5} 
\usepackage{graphicx}
\usepackage{tocloft}

\usepackage{float}

\title{Agents-K1: Towards Agent-native Knowledge Orchestration}

\correspondingauthor{%
$^{*}$Please send correspondence regarding this report to zhangbo@pjlab.org.cn and bailei@pjlab.org.cn%
}

\makeatletter
\renewcommand\AB@authnote[1]{}
\makeatother

\author{\small\centering Zongsheng Cao$^{\spadesuit}$, Bihao Zhan$^{\spadesuit,\clubsuit}$, Jinxin Shi$^{\spadesuit,\clubsuit}$, Jiong Wang$^{\spadesuit,\diamondsuit}$, Fangchen Yu$^{\spadesuit}$, Zhijie Zhong$^{\spadesuit}$, Yingnan Han$^{\spadesuit}$, Zijie Guo$^{\spadesuit,\diamondsuit}$, Tianshuo Peng$^{\spadesuit}$, Zhuo Liu$^{\spadesuit}$, Yi Xie$^{\spadesuit}$, Xiang Zhuang$^{\spadesuit}$, Shengji Tang$^{\spadesuit}$, Yue Fan$^{\spadesuit}$, Runmin Ma$^{\spadesuit}$, Shiyang Feng$^{\spadesuit}$, Xiangchao Yan$^{\spadesuit}$, Anran Liu$^{\spadesuit}$, Peng Ye$^{\spadesuit}$, Wenlong Zhang$^{\spadesuit}$, Xiaosong Wang$^{\spadesuit}$, Shufei Zhang$^{\spadesuit}$, Chunfeng Song$^{\spadesuit}$, Fenghua Ling$^{\spadesuit}$, Jie Zhou$^{\clubsuit,\spadesuit}$,\\ Liang He$^{\clubsuit,\spadesuit}$, Bo Zhang$^{\spadesuit}$$^{*}$, Lei Bai$^{\spadesuit}$$^{*}$}

\affil[$\spadesuit$]{\small Shanghai Artificial Intelligence Laboratory}
\affil[$\clubsuit$]{\small East China Normal University}
\affil[$\diamondsuit$]{\small Fudan University}

\usepackage{pdflscape}
\usepackage{amsmath, amssymb}                              
\usepackage{booktabs,array,multirow}         
\usepackage{algorithm,algpseudocode}
\usepackage{textcomp}
\usepackage{rotating}
\usepackage{setspace}
\usepackage{microtype} 
\usepackage{soul} 
\usepackage{subcaption}
\usepackage{caption}

\usepackage[table]{xcolor}
\sethlcolor{green!14}

\usepackage{array}

\usepackage{bbm}
\usepackage{siunitx}

\usepackage{enumitem}
\usepackage{float}
\usepackage{seqsplit}
\usepackage{framed}
\usepackage{tikz}
\usepackage{url}
\usepackage{listings}
\usepackage{colortbl}  
\usepackage{wrapfig}
\usepackage[most]{tcolorbox}
\tcbuselibrary{listingsutf8}

\definecolor{mycolor}{RGB}{50,80,150}
\definecolor{boxbg}{RGB}{245,247,255}

\usepackage{ragged2e}

\colorlet{punct}{red!60!black}
\definecolor{background}{HTML}{F6F8FA} 
\definecolor{delim}{RGB}{20,105,176}
\colorlet{numb}{magenta!60!black}

\lstdefinelanguage{json}{
    basicstyle=\ttfamily\scriptsize, 
    numbers=left,                    
    numberstyle=\tiny\color{gray},   
    stepnumber=1,
    numbersep=8pt,
    showstringspaces=false,
    breaklines=true,                 
    frame=single,                    
    rulecolor=\color{gray!40},       
    backgroundcolor=\color{background}, 
    literate=
     *{0}{{{\color{numb}0}}}{1}
      {1}{{{\color{numb}1}}}{1}
      {2}{{{\color{numb}2}}}{1}
      {3}{{{\color{numb}3}}}{1}
      {4}{{{\color{numb}4}}}{1}
      {5}{{{\color{numb}5}}}{1}
      {6}{{{\color{numb}6}}}{1}
      {7}{{{\color{numb}7}}}{1}
      {8}{{{\color{numb}8}}}{1}
      {9}{{{\color{numb}9}}}{1}
      {:}{{{\color{punct}{:}}}}{1}
      {,}{{{\color{punct}{,}}}}{1}
      {\{}{{{\color{delim}{\{}}}}{1}
      {\}}{{{\color{delim}{\}}}}}{1}
      {[}{{{\color{delim}{[}}}}{1}
      {]}{{{\color{delim}{]}}}}{1},
}
\lstdefinestyle{jsonblock}{
    basicstyle=\ttfamily\scriptsize,
    breakatwhitespace=false,
    columns=fullflexible,
    keepspaces=true,
    showstringspaces=false,
    xleftmargin=0.5em,
    xrightmargin=0.5em,
    linewidth=\dimexpr\linewidth-1em\relax
}

\newtcolorbox{examplebox}{
    colback=white,
    colframe=black!25,
    boxrule=0.4pt,
    sharp corners,
    left=1mm,
    right=1mm,
    top=1mm,
    bottom=1mm,
    fontupper=\normalfont\scriptsize
}

\usepackage{makecell}
\usepackage{adjustbox}
\usepackage[symbol]{footmisc}
\newcolumntype{Y}{>{\RaggedRight\arraybackslash}X}

\newcolumntype{C}{>{\centering\arraybackslash}X} 
\newtheorem{definition}{Definition}

\newtheorem{proposition}{Proposition}
\usepackage{xspace}
\usepackage{bbding}
\usepackage{tabularx}
\newcolumntype{L}{>{\raggedright\arraybackslash}X}
\newcolumntype{Z}{>{\raggedleft\arraybackslash}X}  

\providecommand{\absfont}{\normalfont\small}

\definecolor{AbstractBgColor}{HTML}{F4F7FB}

\hypersetup{
    colorlinks=true,
    linkcolor=blue, 
    citecolor=blue,  
    filecolor=black,
    urlcolor=blue    
}

\usepackage{cuted}

\usepackage{etoolbox}
\usepackage{arydshln}
\usepackage{ulem} 
\makeatletter
\patchcmd{\@tocline}
    {\hfil}
    {\leaders\hbox{\hfil}\hfil}
    {}{}
\makeatother

\begin{document}

\def\theabstract{} 
\begin{tcolorbox}[
    colback=AbstractBgColor, 
    colframe=AbstractBgColor, 
    arc=5pt,                  
    auto outer arc,
    boxrule=0pt,              
    left=8pt, right=8pt, 
    top=4pt, bottom=4pt,      
    parbox=false,
    width=\textwidth,
    before skip=-800pt,        
    after skip=0pt, 
    enlarge top by=-12pt
]

\maketitle

\vspace{-3.0em}

\noindent \textbf{Abstract}\quad Current LLM-based research agents have advanced through agent orchestration, yet largely overlook scientific knowledge orchestration. Existing works often reduce papers to abstracts, surface mentions, and flat \texttt{cites} edges, omitting key entities, claims, evidence, mechanisms, and method lineages essential for scientific reasoning. To this end, we introduce \textbf{Agents-K1}, an end-to-end knowledge orchestration pipeline that converts raw documents into agent-native scientific knowledge graphs. Agents-K1 integrates three components under a unifying theoretical foundation: a multimodal parser whose five-module schema captures entities, multimodal evidence, citations, and typed inter-entity relations across the full paper rather than abstracts alone; a 4B information-extraction backbone trained with GRPO under a rule-based reward; and a graphanything CLI, a tri-source agent interface that unifies web search, multimodal graph retrieval, and cross-document traversal. On top of this, we process 2.46 million scientific papers across six subjects to produce \textbf{Scholar-KG}, of which we release a one-million-paper subset, and the full Scholar-KG is accessible via the SCP link below. The same pipeline can be extended to general-domain corpora and to schema-conformant data synthesis. Extensive experiments demonstrate that Agents-K1 achieves superior performance in scientific information extraction, knowledge graph construction, and multi-hop scientific reasoning. 

\vspace{-1em} 

\begin{center}
    \href{https://scphub.intern-ai.org.cn/detail/42}{\textcolor{black}{\faGithub~\textbf{SCP}}} \qquad
    \href{https://github.com/InternScience/GraphAnything}{\textcolor{black}{\faGithub~\textbf{Code}}} \qquad
    \href{https://huggingface.co/datasets/InternScience/Scholar-kg}{\raisebox{-0.15\height}{\includegraphics[height=1em]{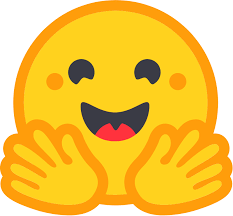}}~\textcolor{black}{\textbf{Data}}} \qquad
    \href{https://huggingface.co/InternScience/Agents-K1-LLM}{\raisebox{-0.15\height}{\includegraphics[height=1em]{imgs/huggingface.png}}~\textcolor{black}{\textbf{Model}}}
\end{center}

\end{tcolorbox}


\begin{figure}[H]
    \centering
    \includegraphics[width=0.95\textwidth, height=\textheight, keepaspectratio]{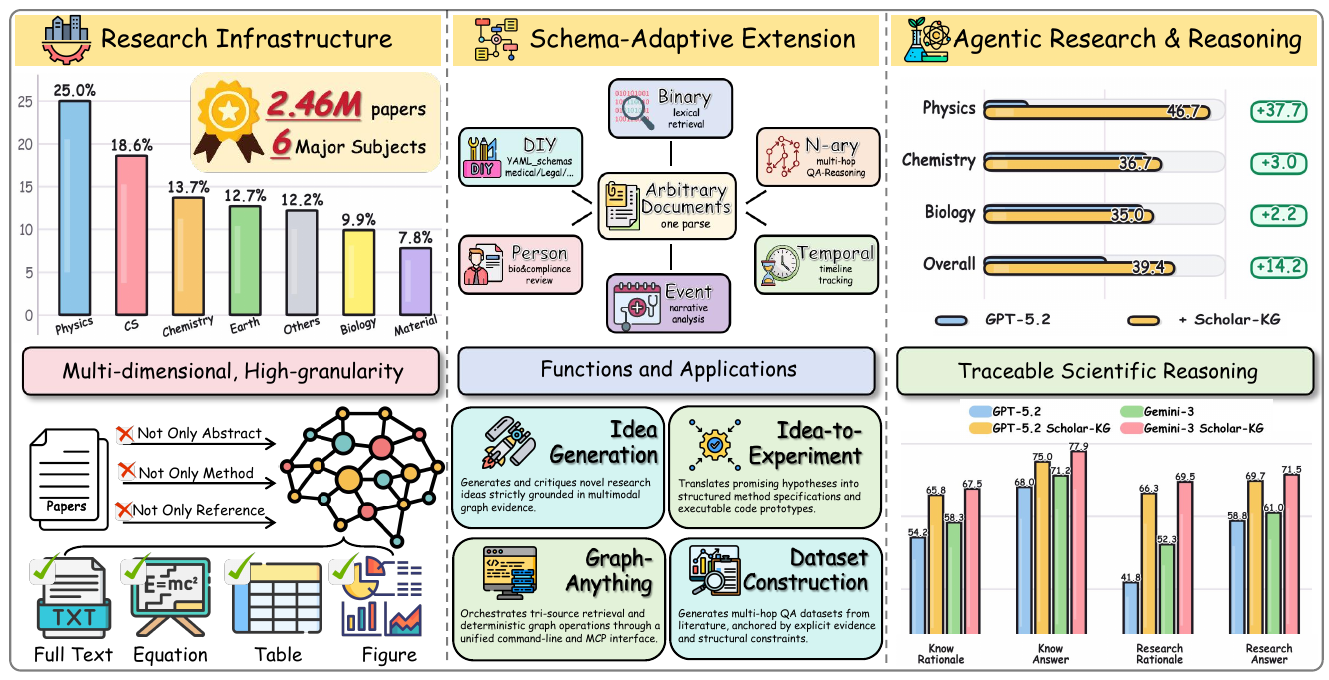}
    \caption{\textbf{Agents-K1}: Architecture and Capabilities. \textbf{Left}: Extracting multimodal knowledge from scientific papers. \textbf{Middle}: Schema-adaptive extensions for core research tasks. \textbf{Right}: Enhancing LLM reasoning and verifiable knowledge tracing.}
    \label{fig:framework}
\end{figure}

\newpage
\begingroup
\hypersetup{linkcolor=black} 
\tableofcontents
\endgroup

\newpage
\section{Introduction}
\label{sec:introduction}

LLM-based research agents have moved from prototype to deployment over the past two years. Advanced systems such as AI-Scientist~\cite{yamada2025ai}, InternAgent~\cite{team2025novelseek,feng2026internagent}, and AI Co-Scientist~\cite{gottweis2025towards} can now plan experiments, retrieve literature, write code, and draft papers in a single loop. Specifically, their progress depends on two parts: \textit{agent orchestration}, which decides how agents plan and act, and \textit{knowledge orchestration}, which decides what knowledge they can use and how that knowledge is organized. Most recent work has focused on agent orchestration. The knowledge counterpart, however, remains much less developed.

This gap matters because a research agent needs more than a list of relevant papers. It needs structured knowledge that preserves what each paper claims, what evidence supports the claim, how the claim relates to prior work, and where the evidence can be checked. Current scientific knowledge infrastructure falls short of this requirement in three ways. First, modern graph-augmented retrieval pipelines, including LightRAG~\cite{guo2025lightragsimplefastretrievalaugmented}, HippoRAG~\cite{jimenez2024hipporag}, HippoRAG2~\cite{gutiérrez2025ragmemorynonparametriccontinual}, GFM-RAG~\cite{luo2025gfm}, E$^2$GraphRAG~\cite{zhao2025e2graphragstreamlininggraphbasedrag}, RAPTOR~\cite{sarthi2024raptorrecursiveabstractiveprocessing}, and KGP~\cite{wang2023knowledgegraphpromptingmultidocument}, usually build generic text-only triples. They capture little beyond abstracts and directly mentioned terms, so key entities, claims, mechanisms, and method links remain buried in the full paper. Figures, tables, and equations, which often carry key evidence, are usually reduced to captions. Second, scholarly citation graphs usually use a flat \texttt{cites} edge. This shows that one paper references another, but not whether it extends a method, challenges a claim, or only cites a baseline. Third, LLM-based research agents~\cite{yamada2025ai,feng2026internagent,gottweis2025towards} often read raw PDFs or short summaries at runtime. This repeats extraction for each query and makes it hard to trace an answer back to exact evidence. These limitations are not just retrieval errors; they show that the knowledge infrastructure itself is not yet designed for agent reasoning.

These limitations point to a different design goal: agent-native \textit{knowledge orchestration}. It should meet three requirements. First, it should cover the full paper and treat text, figures, tables, and equations as connected evidence rather than separate artifacts. Second, it should turn scientific content into typed knowledge, including entities, claims, mechanisms, method links, citation roles, and inter-entity relations. Third, it should support auditable retrieval, so that an agent can trace each answer or decision back to stable graph identifiers and exact evidence. Meeting these requirements calls for a unified pipeline that combines full-paper graph construction, affordable structured extraction, and an agent-facing retrieval interface. This raises a natural question: \textit {Can we build such a unified pipeline to turn raw scientific papers into agent-ready scientific knowledge graphs and support reliable research-agent reasoning at scale?}

To this end, we present \textbf{Agents-K1}, an end-to-end knowledge orchestration pipeline that converts raw documents into agent-native multimodal knowledge graphs at scale. Agents-K1 integrates a multimodal parser, a five-module extraction schema, a reinforcement-learned extraction model, and a tri-source agent CLI into a single framework, and the same pipeline can be replayed on arbitrary corpora. The first stage is multimodal parsing and schema (Section~\ref{subsec:hierarchical_kg}). It uses a MinerU-based offline parser to ingest PDFs and a five-module schema (metadata, explicit mentions, implicit abstractions, citation intent, and fine-grained inter-entity relations) that organises the entire paper rather than its abstract alone, with figures, tables, and equations admitted as first-class evidence alongside textual entities. The second stage is the extraction backbone (Section~\ref{sec:ie-grpo}). It is a 4B-parameter model trained with Group Relative Policy Optimization and a rule-based reward jointly supervising format compliance, JSON validity, and task-conditioned F1 on named entity recognition, relation extraction, and long-form structured extraction. It surpasses an 8B open-source reference across ten benchmarks and matches a 32B base on NER while remaining inexpensive to retrain on new domains. The third stage is the agent CLI (Section~\ref{sec:agent}). It fuses real-time web search, multimodal graph retrieval, and cross-document network traversal into a single tri-source interface and orchestrates a closed-loop workflow of idea generation, method specification, and code synthesis. On the \textsc{FrontierScience-Research} benchmark~\cite{wang2026frontierscience}, it lifts Gemini-3 overall accuracy from 7.9\% to 24.6\% and GPT-5.2 from 25.2\% to 39.4\%; on geoscience research questions, it lifts Gemini-3 rationale accuracy from 52.3\% to 69.5\%; and on multi-hop QA, it reaches state-of-the-art performance on HotpotQA~\cite{yang2018hotpotqa}, 2WikiMultiHopQA~\cite{ho2020constructing}, and MuSiQue~\cite{trivedi2022musique} against nine graph-augmented retrieval baselines. Theoretical analysis further supports this design by explaining why organizing evidence in one connected graph makes cross-source reasoning more reliable than searching separate text fragments.

We instantiate Agents-K1 on scientific literature by processing 2.46 million papers across six disciplines (computer science, chemistry, biology, earth science, physics, and materials) to produce \textbf{Scholar-KG}, and release a one-million-paper subset for community research. Beyond scientific papers, the same parser and extraction backbone can be retargeted through a schema-adaptive variant to build \textbf{General-KG} over arbitrary document corpora, without bespoke per-domain engineering. The structured JSON outputs can also serve as schema-conformant data for training downstream extraction and reasoning models.

The main contributions of this paper are summarised as follows.
\begin{itemize}[leftmargin=*]
    \item \textbf{Unified scientific knowledge infrastructure.} We introduce \textbf{Agents-K1}, an agent-native knowledge orchestration pipeline that unifies KG, LLM, and CLI in a unified framework. Unlike static scholarly knowledge graphs, Agents-K1 is designed around the full workflow of research agents: parsing papers, extracting structured knowledge, building reusable graphs, and exposing the resulting knowledge through agent-facing interfaces.

    \item \textbf{Million-scale full-paper multimodal knowledge graphs.} Agents-K1 constructs \textbf{Scholar-KG} from 2.46 million scientific papers across six disciplines, with a one-million-paper subset released for community research. The pipeline models the full paper rather than only titles, abstracts, metadata, or citation graphs, extracting entities, claims, evidence, motivations, mechanisms, method lineages, citation intent, and inter-entity relations. 

    \item \textbf{Reinforcement-learned extraction backbone.} We train a 4B-parameter information-extraction model with GRPO and a rule-based reward that jointly supervises format compliance, JSON validity, and task-conditioned F1 on NER, relation extraction, and long-form structured extraction. 

    \item \textbf{GraphAnything CLI for executable graph-based research.} We provide GraphAnything CLI as an agent-facing interface that turns the graph from a static data asset into an executable research tool. It supports web search, multimodal graph retrieval, cross-document traversal, graph operations, and graph evolution through CLI/MCP/API access, enabling research agents to directly use the graph in end-to-end research workflows. 
\end{itemize}

\section{Related Work}
\textbf{Retrieval-Augmented Generation}.  As AI systems are increasingly expected to solve knowledge-intensive tasks \cite{NaiveRAG,huang2026radar,du2026mlevolve}, Retrieval-Augmented Generation (RAG) \cite{cao2025tv,RQ-RAG} has become an important paradigm for extending large language models (LLMs) beyond their internal parametric knowledge. Instead of relying solely on information memorized during pretraining, RAG introduces an external knowledge source into the generation pipeline. In general, it first organizes documents or data into a retrievable knowledge base, then identifies evidence relevant to a given query, and finally conditions the LLM on the retrieved content to support generation and reasoning. By allowing models to access fresh, domain-specific, and task-related information, RAG significantly improves their ability to produce responses that are both factually grounded and contextually appropriate~\cite{LightRAG, MemoRAG}. This makes RAG especially suitable for real-world scenarios where accurate knowledge access and reliable reasoning are required at the same time.

Existing RAG studies have developed along multiple directions, mainly driven by the need to handle increasingly complex knowledge sources. Chunk-based retrieval methods~\cite{NaiveRAG, ChunkRAG, cao2026vig} focus on dividing documents into effective retrieval units and improving the matching between queries and textual segments through better embeddings and segmentation strategies. In contrast, graph-based RAG systems~\cite{GraphRAG, LightRAG, SubgraphRAG} attempt to move beyond independent text chunks by introducing explicit relational structures, thereby enabling models to retrieve and reason over connected knowledge more effectively. Meanwhile, multimodal RAG~\cite{ColPali} further expands the retrieval space beyond plain text to heterogeneous information such as images, audio, and video, enabling richer evidence integration across modalities and application settings.

\textbf{Graph-based Information Understanding. }  
To provide more structured supporting evidence for LLM reasoning, GraphRAG~\citep{GraphRAG} marks a representative step toward incorporating knowledge graph structures \cite{cao2024diffusione} into retrieval-augmented generation. Following this line of work, many approaches~\citep{MedGraphRAG, OG-RAG, KAG, MiniRAG, PIKE-RAG} have adapted graph-based RAG frameworks to different domains and tasks, showing that graph structures can improve how external knowledge is organized, retrieved, and used by LLMs. For example, LightRAG~\citep{LightRAG} enhances retrieval efficiency by introducing graph indexing and update mechanisms, while PathRAG~\citep{PathRAG} and HippoRAG2~\citep{HippoRAG2} further improve graph retrieval through path pruning and Personalized PageRank, respectively.
Nevertheless, the effectiveness of these methods is still largely constrained by how the graph itself is constructed. Most existing graph-based RAG approaches mainly extract knowledge from textual content and encode it as fragmented triples or other local relational units. Although such representations are useful for capturing isolated facts, they often lose the broader semantic continuity of the original corpus, especially when the source documents contain complex structures, long-range dependencies, figures, tables, and hierarchical arguments. As a result, the constructed graphs may fail to reflect the complete information organization of papers and other knowledge-intensive corpora. This motivates the need for an end-to-end information extraction method that can model document-level knowledge more comprehensively and preserve richer semantic relations during graph construction.

\textbf{Deep Research Agents. }  
The Deep Research Agents~\citep{hu2025flowsearch,du2025automlgen,team2025tongyi,li2025webthinker,shi2025dualresearch} advance a long-horizon research paradigm for knowledge-intensive tasks. A typical system follows a recurring loop of planning, retrieval, reading, verification, and re-retrieval, which autonomously fills knowledge gaps and progressively reduces uncertainty. Both OpenAI~\citep{openai_deepresearch} and Gemini~\citep{gemini_deepresearch} emphasize first constructing an adaptable research plan around a complex question, then conducting multiple rounds of exploration and cross-validation across heterogeneous sources to support deeper, research-oriented problem-solving. Another line of work focuses on dynamic reasoning and tool use. For example, InternAgent~\citep{team2025novelseek, feng2026internagent} and WebThinker~\citep{li2025webthinker} make deep research explicit as continual search and information extraction, and improve tool-use quality via preference optimization. Complementary to multi-agent paradigms, the single-agent direction also evolves rapidly. Tongyi DR~\citep{team2025tongyi} trains a self-regulated reasoning agent with reinforcement learning to learn action selection.

However, most existing solutions still treat webpages or plain-text retrieval as the primary gateway to knowledge, making it difficult to fully exploit key evidence in scientific literature carried by figures, tables, and formulas. In contrast, this paper uses semantic anchors to integrate text, images, tables, and formulas into a unified, retrievable knowledge representation and to fuse three sources: web retrieval, multimodal knowledge graph retrieval, and cross-paper knowledge network traversal. This design provides a more structure-aware and evidence-traceable knowledge infrastructure for deep research workflows, reduces information blind spots caused by text-centric retrieval, and opens a new path to better control the solution space and improve reproducibility.

\section{The Overall Agents-K1 Framework}
\label{sec:solution}

\begin{figure}[H] 
    \centering
    \vspace{-4mm}
    \includegraphics[width=1\textwidth]{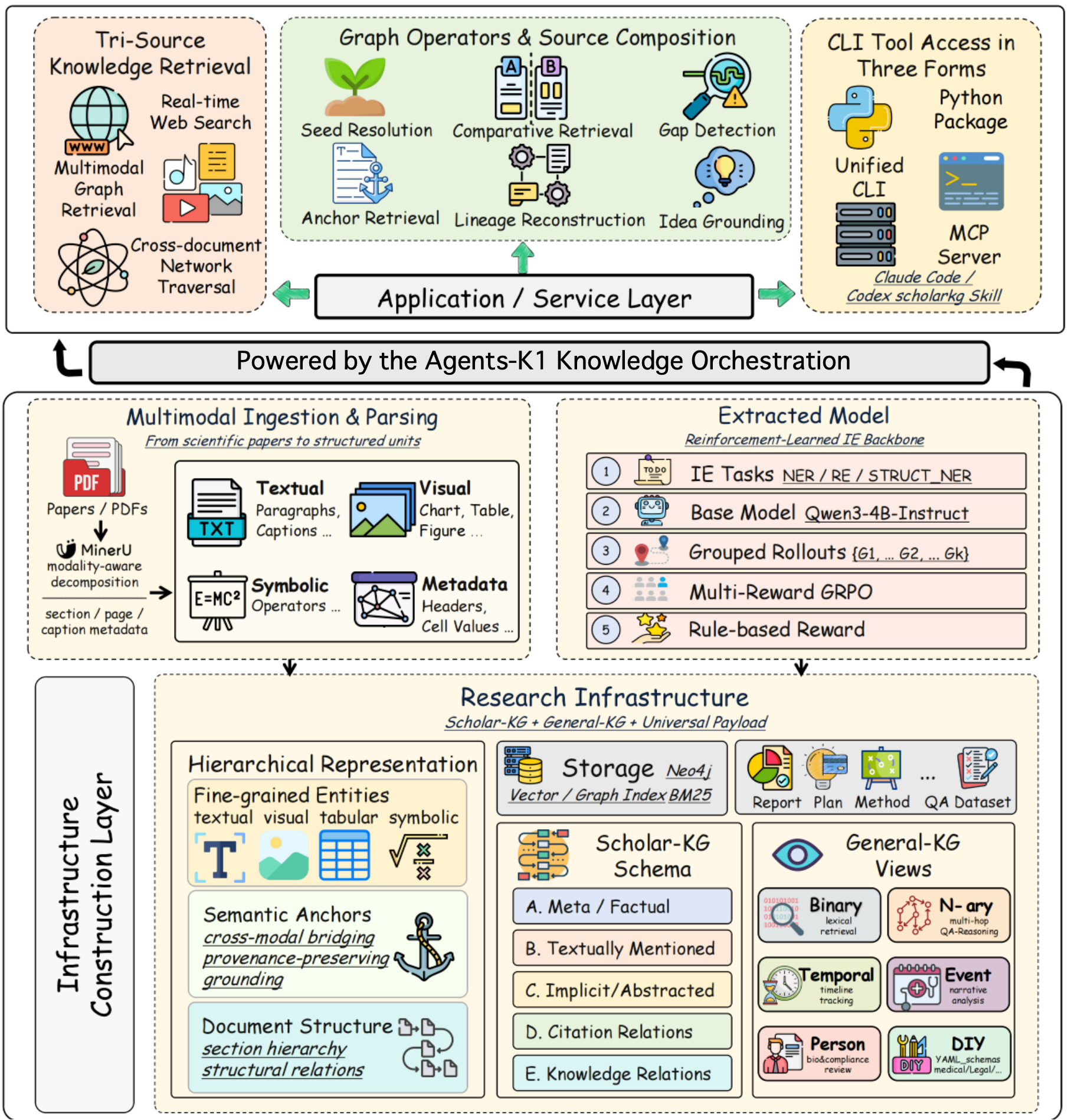}
    \caption{\textbf{Overall framework of Agents-K1.} Agents-K1 features three parts. The \textbf{infrastructure construction layer} parses multimodal documents into structured knowledge graphs (\textbf{Scholar-KG} and \textbf{General-KG}). Utilizing this foundation, the \textbf{application/service layer} offers a \textbf{tri-source agent CLI} as the main interface, enabling comprehensive and auditable research workflows such as comparative analysis and idea grounding.}
    \label{fig:framework}
\end{figure}

Scientific research agents need more than a powerful reasoning model. To answer complex research questions, an agent must identify relevant papers, inspect evidence across text, figures, tables, and equations, connect claims across documents, and keep the final answer traceable to its sources. However, many existing scholarly infrastructures expose only abstracts, metadata, and coarse citation links. Important information such as methods, assumptions, evidence, limitations, and method lineages therefore remains buried in raw papers, forcing the agent to repeatedly recover this structure at query time.

Agents-K1 is designed as a reusable knowledge infrastructure for this setting. Instead of asking the agent to reason directly over scattered PDFs and flat search results, Agents-K1 first converts documents into structured multimodal knowledge graphs, then exposes these graphs through retrieval and tool interfaces that an agent can use during research. In this way, the framework separates two jobs that are often mixed together: offline construction of reliable knowledge representations, and online use of this knowledge for evidence-grounded reasoning.

As shown in Figure~\ref{fig:framework}, Agents-K1 addresses this problem through three coordinated components. First, the KG layer (Section~\ref{subsec:hierarchical_kg}) parses the full paper rather than only the abstract, extracts multimodal content units, and organizes metadata, explicit concepts, implicit claims, citation intent, and fine-grained relations into graph structures. This produces \textbf{Scholar-KG} for scientific literature and extends to \textbf{General-KG} for other document collections through a schema-adaptive design. Second, the LLM layer (Section~\ref{sec:ie-grpo}) supplies the extraction backbone used to populate these graphs. We train a compact 4B model with reinforcement learning and rule-based rewards so that large-scale extraction can remain accurate, structured, and affordable to adapt. Third, the agent CLI layer (Section~\ref{sec:agent}) makes the constructed knowledge usable by research agents. Its tri-source CLI combines web search, multimodal graph retrieval, and cross-document traversal, allowing the agent to retrieve evidence, follow provenance, compare related works, and ground new ideas in existing literature.

\section{KG Layer: An end-to-end Pipeline for Scientific Knowledge Construction}
\label{subsec:hierarchical_kg}

A central challenge in scientific knowledge representation lies in bridging the semantic gap between heterogeneous content types. Traditional approaches either flatten all modalities into text (losing visual structure) or maintain separate representations (complicating cross-modal retrieval). We observe that scientific documents exhibit a natural hierarchical organization: fine-grained entities (methods, datasets, metrics) are semantically grounded within coarse-grained content units (figures, tables, paragraphs), which together compose the document's narrative structure. This observation motivates our semantic anchor design, an intermediate abstraction layer that serves as modality-agnostic bridges, enabling entities from different modalities to connect through shared semantic anchors rather than requiring direct cross-modal alignment.

\noindent\textbf{Multimodal Content Decomposition}
The first stage transforms raw documents into a canonical representation suitable for unified processing. Each knowledge source $k_i \in \mathcal{K}$ undergoes modality-aware decomposition:
\begin{align}
k_i \;\xrightarrow{\;\text{Parse}\;}\; \mathcal{C}_i = \{c_j = (t_j, x_j, m_j)\}_{j=1}^{n_i},
\end{align}
where each content unit $c_j$ comprises a modality type $t_j \in \{\text{text}, \text{figure}, \text{table}, \text{equation}\}$, raw content $x_j$, and structural metadata $m_j$ (e.g., section hierarchy, page location, caption associations). Specialized parsers handle each modality: text is segmented into semantically coherent paragraphs; figures are extracted with captions and cross-references; tables preserve cell structure with headers; equations retain symbolic representations alongside surrounding context. This decomposition maintains the document's inherent organization while enabling modality-specific processing.

\noindent\textbf{Unified Heterogeneous Graph with Semantic Anchors.}
For the given multimodal document, instead of constructing modality-specific graphs and aligning them post-hoc, we represent all multimodal content within a single heterogeneous graph $\mathcal{G}=(\mathcal{V},\mathcal{E},\phi_v,\phi_e)$ organized into three layers. The lowest layer contains fine-grained entities extracted from text, figures, tables, and equations (e.g., named entities, visual concepts, table entries, symbolic variables). To avoid brittle cross-modal entity alignment, we introduce a middle semantic anchor layer: for each content unit $c_j$, we generate an abstract anchor node
\begin{equation}
a_j = f_{\text{MLLM}}(c_j,\mathcal{N}_j),
\end{equation}
where $\mathcal{N}_j$ provides local contextual grounding. Each anchor captures a modality-agnostic semantic summary together with salient entities and relations, serving as a stable bridge across modalities while preserving fine-grained provenance.

The top layer models document structure (sections and documents), enabling hierarchical retrieval and global context. Cross-layer edges link entities to anchors via \texttt{grounded\_in} relations, and anchors to structure via \texttt{belongs\_to} edges. Relationships among anchors are induced through explicit references, shared canonical entities, and semantic similarity in embedding space. By routing multimodal interactions through semantic anchors rather than direct entity matching, the graph achieves improved robustness and retrieval reliability while remaining lightweight and extensible.

\subsection{Scientific Knowledge Network Construction}
\label{subsec:knowledge_network}
To effectively utilize the abundant literature and papers, we collect them and leverage MinerU to conduct multimodal understanding. In this way, we construct a comprehensive scientific knowledge network that captures both explicit and implicit knowledge from research papers. This knowledge network serves as a structured repository that organizes entities, relationships, and scientific claims extracted from the literature, enabling efficient retrieval and reasoning for downstream applications. The network construction process involves defining a hierarchical schema, implementing multimodal knowledge extraction, and organizing the extracted knowledge into a queryable graph structure.

\begin{figure*}[t] 
    \centering
    \includegraphics[width=1\textwidth]{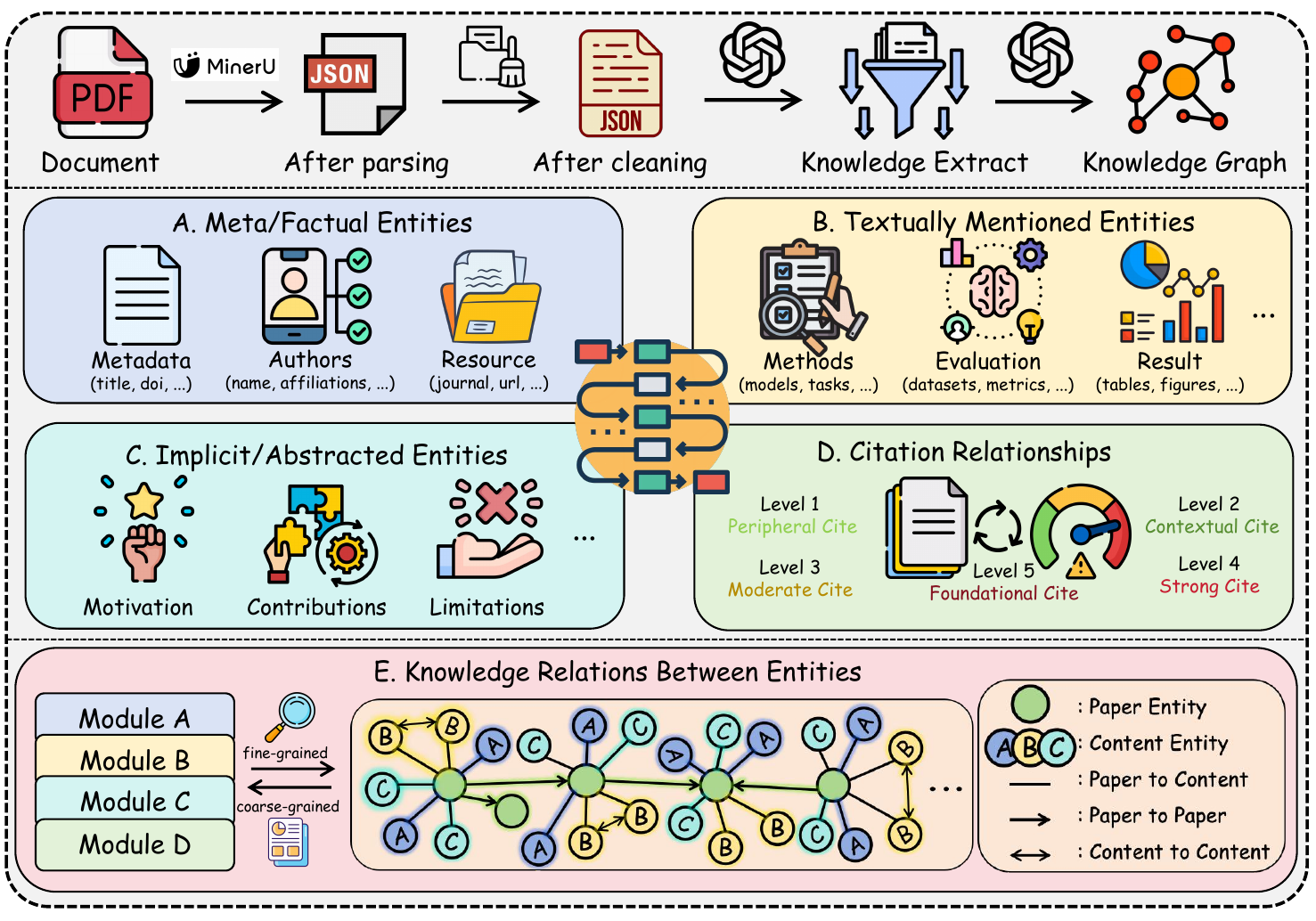}
    \caption{Framework for Scientific Knowledge Network Construction. (a) Meta/factual entities provide a stable backbone for deduplication and provenance tracking through standardized metadata, author canonicalization, and resource linking. (b) Textually mentioned entities capture explicit scientific objects (methods, datasets, metrics) with synonym recognition and core-focus recall for fair benchmarking. (c) Implicit/abstracted entities synthesize high-level knowledge, including motivations, contributions, and limitations, through rhetorical-role tagging and discourse parsing. (d) Citation relationships encode argumentative intent (support/contrast/extend) and strength scores, enabling lineage tracing and causal mapping of claim propagation across the literature. (e) Knowledge relations between entities refine coarse abstractions into fine-grained, queryable triples, turning the graph into a navigable reasoning surface.}
    \label{fig:kg}
\end{figure*}

As shown in the Figure~\ref{fig:kg}, our knowledge graph schema is designed to capture the multi-faceted nature of scientific literature, encompassing verifiable metadata, explicitly mentioned concepts, implicitly abstracted knowledge, and citation relationships. We organize entities into five categories:

\begin{figure*}[t] 
    \centering
    \includegraphics[width=1\textwidth]{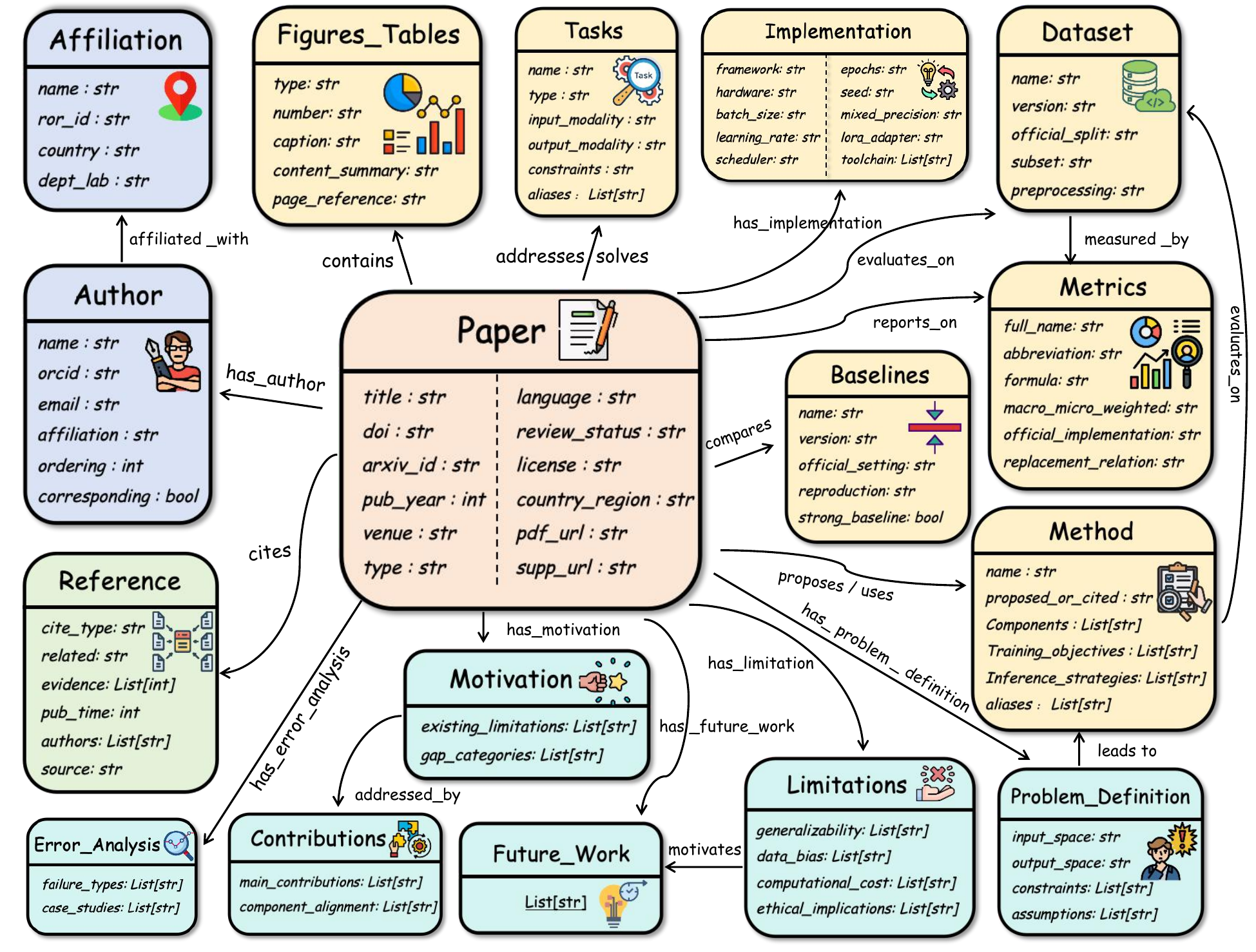}
    \caption{Disaggregated Schema of our KG. This schema illustrates the structural decomposition of a multimodal scientific document into a unified, queryable knowledge graph. By explicitly modeling these entities and their interconnected relationships , the schema dismantles unstructured text to furnish AI agents with a rigorous reasoning infrastructure. For detailed examples, please refer to Appendix \ref{sec:appendix_json}.}
    \label{fig:schema}
\end{figure*}

\paragraph{A. Meta/Factual Entities.}
We define a stable backbone of low-variance, verifiable metadata that supports deduplication, disambiguation, and provenance across the corpus. For each Paper, we normalize title, doi/arXiv\_id, pub\_year, venue (conf/journal/workshop), type (long/short), language, peer\_review\_status, license, country/region, and resolvable pdf/supplement URLs. Authors are canonicalized (Surname, Given) with ORCID/email (hashed if required), ordered positions, corresponding flags, and linked Affiliations harmonized to ROR IDs with country and dept/lab fields. We capture process times (submission, accept, camera-ready) and open-science Resources, including repository URLs pinned to commits, model artifact hashes, and dataset releases with names/versions. Each field is stored with provenance $\langle\text{doc, section/page, span}\rangle$ and a calibrated confidence score, providing a trustworthy spine for downstream reasoning and global entity linking.

\paragraph{B. Textually Mentioned Entities.}
We lift all explicitly referenced scientific objects into normalized nodes to enable high-precision retrieval and reproducible comparisons. Span detectors and sequence labelers identify noun phrases for \textit{Tasks/Problems} (names, modalities, settings such as zero/few-shot and common aliases), \textit{Methods/Models/Algorithms} (proposed/cited names, modular components like encoders/experts/routers, training objectives such as cross-entropy/KL/GRPO, and inference strategies), \textit{Datasets/Splits/Modalities} (name, year/version, official splits, subsets, preprocessing such as tokenizers or image sizes), \textit{Metrics} (long names, acronyms, formula semantics, macro/micro/weighted variants and alignment with official implementations, including deprecations like BLEU$\rightarrow$ChrF), \textit{Baselines} (versioned and tagged as official/reproduced/strong), \textit{Implementation/Training} details (framework, hardware and VRAM class, batch/lr/scheduler, epochs, seeds, AMP, adapters, and toolchain such as Apex/Deepspeed/FlashAttn), formal \textit{Theorems/Definitions/Lemmas} (IDs, premises, conclusions), and structured references to \textit{Figures/Tables/Equations/Examples} (IDs, captions, page anchors). Mentions are canonicalized through controlled vocabularies and embedding-based entity linking into the ontology; all nodes carry evidence spans and confidence, ensuring that each textual reference is reconciled to a single, queryable concept that supports fair benchmarking and faithful reproduction of setups.

\paragraph{C. Implicit/Abstracted Entities.}
Beyond what is explicitly named, we synthesize high-value abstractions that capture \textit{what the paper claims, assumes, and finds}, which are indispensable for cross-paper reasoning. Using rhetorical-role tagging and discourse parsing, we aggregate sentences across sections to construct a formal \textit{Problem Definition} $(\mathcal{X},\mathcal{Y},\mathcal{C},\mathcal{A})$ specifying input/output spaces, constraints, and assumptions (e.g., independence/stationarity); codify \textit{Motivations/Gaps} into a taxonomy (data efficiency, long-range dependence, tool-integration failure, etc.); align bulletized \textit{Contributions} to concrete components so that each improvement is attributable; register \textit{Hypotheses/Assumptions} with testability flags and scope; extract \textit{Findings/Claims} with quantitative effect sizes and qualitative patterns (consistency, robustness, scaling behavior); record \textit{Explanations/Mechanisms} (causal, mechanistic, or heuristic rationales) that account for observed results; and document \textit{Limitations/Threats} (external validity, biases, compute cost, ethics), \textit{Design Rationales}, \textit{Future Work/Open Questions}, and \textit{Error Analyses/Case Studies}. These abstracted nodes are evidence-linked and confidence-calibrated, merged across documents when semantically equivalent, and provide a structured infrastructure for causal-aware synthesis, systematic reviews, and meta-analyses.

\paragraph{D. Citation Relationships.}
We represent how the literature positions a focal work by encoding \textit{who cites whom, how strongly, and to what argumentative effect}. Each \textit{Citation} edge stores the cited paper's identity (title, authors, year, DOI/venue), \textit{Cite Type} (strong/weak, direct support vs.\ indirect mention, core reference flag), a one-sentence \textit{Relation} (support/contrast/extend/background) with its argumentative role, fine-grained \textit{Evidence} (section/paragraph indices, in-text spans, frequency), temporal information (publication date and timeline ordering), and enriched \textit{Author/Team} and \textit{Source} signals (affiliations, collaborations, venue prestige). We compute a citation-strength score from frequency, rhetorical-zone coverage, lexical cues (e.g., ``we adopt'', ``we improve upon''), and proximity to key entities, and threshold it to label strong vs.\ weak citations; relation roles are predicted with cue-based classifiers and verified for high-impact cases via human-in-the-loop. The resulting citation layer, together with Modules A–C, yields a heterogeneous, provenance-rich knowledge graph that supports evidence-grounded retrieval, lineage tracing of datasets/metrics, identification of pivotal works, and causal mapping of how claims propagate and evolve across disciplines. For the detailed citation context classification schema, please refer to Appendix \ref{subsec:citation_schema}.

\paragraph{E. Knowledge Relations Between Entities.}
Modules B and C already capture scientific knowledge, but at a coarse, paper-level granularity: B records explicitly mentioned objects and their attributes, while C aggregates motivations, mechanisms, and limitations as high-level nodes attached to each paper. Module E refines this into fine-grained triples between specific entities, so that knowledge previously encoded as paragraph-level abstraction becomes directly queryable. We organize the triples into two complementary families. Controlled relations require head and tail to already exist in Module B, preserving ontological discipline over the method, task, and dataset space; they include \textit{BUILDS\_ON}, \textit{USES\_COMPONENT}, \textit{ALTERNATIVE\_TO}, \textit{SOLVES}, \textit{APPLIED\_TO}, and \textit{TARGETS}. Open relations admit new concepts when a mechanism lacks a canonical name, spanning four semantic zones: causal (\textit{CAUSES}, \textit{ENABLES}, \textit{INHIBITS}, \textit{MODULATES}, \textit{CORRELATED\_WITH}), internal composition (\textit{USES\_TECHNIQUE}, \textit{CONSISTS\_OF}, \textit{IMPLEMENTS}, \textit{COMBINES}, \textit{REQUIRES}), methodological comparison (\textit{DERIVED\_FROM}, \textit{DIFFERS\_FROM}, \textit{HAS\_LIMITATION}, \textit{ADDRESSES\_PROBLEM}, \textit{MOTIVATED\_BY}), and domain structure (\textit{HAS\_PROPERTY}, \textit{SUBSET\_OF}). By design, this layer admits only durable knowledge: numerical results, hyperparameters, and experimental setups are excluded, because they either already reside under Module B or vary across replications and would dilute the graph with ephemeral noise. Triples arrive through two reinforcing channels. Structural edges are materialized deterministically from Modules A and B, for instance \textit{AFFILIATED\_WITH} built from author metadata or \textit{USES\_COMPONENT} from declared submodules. Semantic edges are mined with section awareness, harvesting causal claims primarily from Introduction and Discussion, internal composition from Methods, comparative gaps from Related Work, and failure modes from Limitations, with alias reconciliation against Module B to prevent entity fragmentation. Each triple records $\langle\text{head}, \text{head\_type}, \text{relation}, \text{tail}, \text{tail\_type}\rangle$ with a verbatim evidence span, calibrated confidence, and a source tag distinguishing semantic from structural provenance; semantically equivalent triples are merged across documents via canonicalization over entity embeddings. Together with Modules A through D, this relational layer refines coarse abstractions into a structure that can be traversed and reasoned over, turning the graph from a catalog of entities into a navigable reasoning surface that supports multihop causal tracing, architectural provenance, methodological lineage comparison, and cross-paper synthesis of limitations and design rationales that would otherwise remain buried in prose.

\subsection{Dataset Construction via LLM-Guided Multi-Hop QA Generation.}
Beyond constructing Scholar-KG, the same structured paper representation can be reused to synthesize evidence-grounded multi-hop QA data, which serves as an additional downstream artifact of the KG construction pipeline.

Given a corpus of academic papers $\mathcal{P}=\{p_j\}_{j=1}^{M}$, where each paper is represented as a structured JSON object containing metadata and full text, our objective is to automatically construct a high-quality multi-hop reasoning QA dataset $\mathcal{D}$. Each QA instance is designed to probe deep paper understanding and consists of a complex question requiring 3--5 explicit reasoning steps, an exact and unambiguous answer that is a concrete term or method stated in the paper, an ordered chain of factual statements sufficient to derive the answer, explicit textual evidence from the source paper, and a structured entity--relation graph. To make the generation tractable within the model context window while preserving sufficient information, we first transform each paper $p$ into a compact prompt representation
\[
S(p) \triangleq \text{concat}(\texttt{metadata}, \texttt{abstract}, \texttt{content}[:L]),
\]
where $L$ is a fixed truncation threshold. This summary is embedded into a unified prompt $\Pi(p)$ composed of a reusable schema-constrained template and task-specific instructions, ensuring that the model produces exactly five QA instances per paper with consistent structure and explicit evidence grounding.

\paragraph{Structured Generation, Validation, and Quality Control.}
We formulate QA construction as a conditional text-to-structure generation problem using a large language model $\mathcal{M}_\theta$, which maps the paper-specific prompt to a structured JSON output:
\[
\hat{Y}(p) \sim \mathcal{M}_\theta(\Pi(p); \tau),
\]
where $\tau$ denotes the decoding temperature. The model is strictly instructed to output a valid JSON array of five objects, each containing a question, an exact answer, a multi-hop chain of key facts, source evidence excerpts, and an entity tree encoding entities and typed relations. To ensure reliability and suitability for benchmark construction, we apply a robust post-processing and validation pipeline that enforces structural correctness, non-emptiness of all fields, answer faithfulness to the paper text, logical sufficiency of the reasoning chain, and well-formedness of the entity graph. This combination of constrained prompting and deterministic validation yields a scalable and reproducible procedure for generating high-quality, evidence-grounded multi-hop QA data from scientific literature, while remaining agnostic to the underlying inference backend (API-based or locally hosted models).



\subsection{General-KG: A Multi-View, Schema-Adaptive Extension to Arbitrary Documents}
\label{sec:general_kg}

Our KG schema introduced above specialises to peer-reviewed scientific literature and commits to the five-module Module A through E decomposition. Many downstream deployments require the same infrastructure to run over heterogeneous corpora such as medical reports, legal filings, and financial disclosures, and require different graph granularities over the same document: lexical retrieval prefers compact binary triples, multi-hop reasoning prefers $n$-ary hyperedges that keep co-occurring arguments bundled, and event tracking prefers edges carrying explicit temporal qualifiers. General-KG addresses these requirements with a multi-view, schema-adaptive extension that reuses the OCR front-end, the semantic-anchor layer, and our model extraction backbone, while replacing the fixed schema with a family of pluggable views and a cold-start procedure that adapts to new verticals without weight updates.

\paragraph{Multi-View Mode Family.}
General-KG exposes the same document through six complementary views, each implemented as an independently registered mode that produces a uniform payload with typed nodes $\mathcal{V}$, hyperedges $\mathcal{E}_{h}$, evidence spans, and source-chunk provenance. Table~\ref{tab:gkg_modes} lists the modes and their target tasks. The modes are compositional rather than exclusive: a typical deployment activates several at once and lets the agent layer route per query. The set is open; registering a new mode requires a decorator in a new module directory, with no change to the shared pipeline.

\begin{table}[h]
\centering
\caption{The six graph views exposed by General-KG. All modes share a common output schema so that downstream retrieval and agent code remain mode-agnostic. Projection modes derive their output deterministically from the core artifact; upgrade modes issue one additional LLM pass per chunk on top of the core.}
\small
\setlength{\tabcolsep}{5pt}
\begin{tabular}{@{}p{0.10\linewidth} p{0.22\linewidth} p{0.37\linewidth} p{0.21\linewidth}@{}}
\toprule
\textbf{Mode} & \textbf{Graph form} & \textbf{Best-fit downstream task} & \textbf{LLM calls / chunk} \\
\midrule
\texttt{binary}   & triple $(h,r,t)$         & lexical retrieval, path queries             & $0$ (projection) \\
\texttt{nary}     & hyperedge, $k \ge 3$     & multi-hop QA with bundled arguments          & $1$ (upgrade)    \\
\texttt{temporal} & edge $+$ time qualifier  & timeline reasoning, change tracking           & $1$ (upgrade)    \\
\texttt{person}   & person-centric subgraph  & biography, compliance, author graph          & $0$ (projection) \\
\texttt{event}    & event node $+$ roles     & narrative, FrameNet-style analysis            & $1$ (upgrade)    \\
\texttt{diy}      & YAML-declared schema     & medical, legal, financial verticals            & $1$ (upgrade)    \\
\bottomrule
\end{tabular}

\label{tab:gkg_modes}
\end{table}

\paragraph{Core-then-Modes Architecture.}
Running $M$ independent extraction pipelines would multiply LLM cost by $M$ and would defeat the multi-view story on corpora of realistic size. We instead factor the extraction into a shared core stage and a family of mode stages. Given a document segmented into chunks $\{c_i\}_{i=1}^{n}$, the core stage produces a canonical typed entity set and a binary relation skeleton,
\begin{equation}
  \big(\mathcal{V}_{\text{doc}},\; \mathcal{E}_{\text{skel}}\big) = f_{\text{core}}\!\left(\{c_i\}_{i=1}^{n}\right),
  \label{eq:gkg_core}
\end{equation}
using two LLM passes per chunk: the first extracts typed entities and merges them across chunks by case-insensitive canonical name, and the second is conditioned on the merged entity set and extracts strictly binary relations with evidence quotes, rejecting any triple whose head or tail falls outside $\mathcal{V}_{\text{doc}}$. Each mode then consumes $(\mathcal{V}_{\text{doc}}, \mathcal{E}_{\text{skel}}, \{c_i\})$ and dispatches to one of two strategies. Projection modes (\texttt{binary}, \texttt{person}) derive their output deterministically by filtering and relabelling the skeleton, and cost zero additional LLM calls. Upgrade modes (\texttt{nary}, \texttt{temporal}, \texttt{event}, \texttt{diy}) issue one LLM pass per chunk that takes the skeleton as a structural anchor and lifts it into the target form, for example by merging co-occurring binary edges into hyperedges of arity $k \ge 3$, or by attaching time qualifiers drawn from $\{\textsc{point\_time}, \textsc{start\_time}, \textsc{end\_time}, \textsc{before}, \textsc{after}\}$. Because the entity set is canonicalised once in the core, every upgrade pass operates on a strictly smaller hypothesis space than an end-to-end extractor would face, which both shortens the prompt and empirically reduces hallucinated participants. Letting $n_{\text{up}}$ denote the number of activated upgrade modes and $M$ the total number of activated modes, the per-document cost satisfies
\begin{equation}
  \mathcal{C}_{\text{ours}} = n \cdot (c_{\text{core}} + n_{\text{up}}) \;\le\; \mathcal{C}_{\text{naive}} = c_{\text{core}} \cdot n \cdot M,
  \label{eq:gkg_cost}
\end{equation}
with $c_{\text{core}} = 2$ in our implementation. For a representative all-views configuration with $n = 8$, $n_{\text{up}} = 4$, and $M = 6$, the factoring reduces LLM calls from $96$ to $48$, a fifty percent saving, while producing the full six-view output. A second structural benefit is that every view references the same node identifiers, so cross-view joins such as pivoting from a \texttt{binary} retrieval hit to the \texttt{nary} hyperedges that contain the same head entity reduce to identifier lookup rather than fuzzy string matching. This property is what later allows the CLI layer to treat different views as interchangeable retrieval sources.

The mode family above answers how a document is viewed; it does not answer what the extractor should look for in a vertical that our model has never seen. Hand-authoring a prompt for every vertical is slow, does not scale across customers, and loses ground as the corpus drifts. We therefore attach a weight-frozen self-improvement loop that distils a reusable skill library from a small seed of gold annotations. Given $10$ to $20$ hand-curated gold documents per vertical, the loop proceeds in four phases per gold document. In the rollout phase, the current extractor is sampled $K$ times with non-zero decoding temperature to elicit genuine sampling variance, producing candidate edge sets $\{R_k\}_{k=1}^{K}$. In the classify phase, every gold edge $e^{\star}$ is scored against $\{R_k\}$ under a matcher that normalises the relation to \textsc{upper\_snake\_case} and treats participants as a case-insensitive set, yielding the label
\begin{equation}
  \text{label}(e^{\star}) = \begin{cases}
    \textsc{stable}   & \text{if } e^{\star} \text{ matches for all } k \in [K], \\
    \textsc{unstable} & \text{if } e^{\star} \text{ matches for some but not all } k, \\
    \textsc{miss}     & \text{otherwise}.
  \end{cases}
  \label{eq:skill_label}
\end{equation}
\textsc{stable} edges are already handled reliably and are skipped. \textsc{unstable} edges drive path induction, in which the LLM is given the evidence span of $e^{\star}$ and asked to verbalise a minimal trigger pattern that would have produced the edge on every rollout. \textsc{miss} edges drive hindsight reasoning, in which the LLM is given both the gold edge and the extractor's failing output, asked to diagnose why the edge was plausibly missed, and asked to propose a corrective pattern. Both paths yield candidate skills anchored by concrete evidence quotes, which we find essential for downstream retrievability. A deterministic controller then folds each candidate skill into the library under one of four actions (\textsc{add}, \textsc{modify}, \textsc{merge}, \textsc{keep}), using Jaccard similarity over pattern tokens with a tunable threshold to decide overlap. At inference time, a lightweight retriever scores every skill against the document title and the first $2{,}000$ characters of body text and prepends the top-$k$ skills (default $k = 3$) to the core-extraction system prompt; when the library is empty or the activation flag is off, the retriever returns an empty context and the pipeline is byte-identical to the skill-free baseline, so the skill loop is a pure addition rather than a required path. Leaving our model weights untouched yields three properties that matter in deployment: cold-starting a new vertical requires no GPU run, cross-domain regression at the weight level is impossible by construction, and every skill is human-auditable and version-controlled, which is decisive in regulated domains where prompt provenance is a compliance concern rather than a nice-to-have.

\paragraph{DIY Schema via YAML.}
Some verticals require entity and relation types that fall outside the five built-in modes, for example \textsc{drug}, \textsc{enzyme}, and \textsc{adverse\_effect} for drug interaction analysis, or \textsc{party}, \textsc{obligation}, \textsc{deadline}, and \textsc{penalty} for contract review. Implementing a new extraction mode in Python for every such vertical couples schema evolution to code-deployment cycles, which is unacceptable in regulated domains. The \texttt{diy} mode therefore consumes a self-contained YAML template that declares \texttt{entity\_types}, \texttt{relation\_types}, \texttt{qualifiers}, and a few-shot example block; at runtime the mode parses the template, assembles a schema-conditioned prompt from the few-shot block, and dispatches one LLM call per chunk in the same shape as the other upgrade modes, which keeps the cost formula of Eq.~\ref{eq:gkg_cost} intact. The YAML template and the skill library compose cleanly: the template fixes the target ontology and supplies a small number of canonical examples sufficient to stand up a functional pipeline, and the skill library then accumulates domain-specific extraction patterns that operate within that ontology. The result is that the marginal cost of onboarding a new vertical becomes dominated by gold annotation time rather than by engineering time.

\paragraph{Relation to our KG.}
General-KG is a parallel track rather than a replacement. Our KG remains the canonical choice for academic corpora where the five-module schema is exactly the intended output and where citation structure and method lineage are first-class concerns. General-KG is the canonical choice for arbitrary documents where multi-view, schema-adaptive output is required and where the ontology itself evolves with the corpus. The two tracks share the OCR front-end, the semantic-anchor layer, and our model extraction backbone, and they converge at the CLI layer, which treats their outputs as interchangeable retrieval sources mediated by the tri-source adaptive fusion mechanism.

\subsection{Knowledge Organization and Storage}
\label{sec:kg_organization}

The extracted entities and relations are organized into a heterogeneous knowledge graph $\mathcal{G} = (\mathcal{V}, \mathcal{E}, \mathcal{R})$, where $\mathcal{V}$ is the set of entity nodes, $\mathcal{E}$ is the set of typed edges, and $\mathcal{R}$ is the set of relation types. We adopt a graph database (e.g., Neo4j) for flexible schema management and efficient graph traversal queries.

\textbf{Indexing for Retrieval.} To support efficient retrieval in downstream applications, we construct multiple indices: (i) a text index over entity names and descriptions using BM25, (ii) a vector index over entity embeddings for semantic search, and (iii) a graph index for fast neighborhood queries. Each entity node stores provenance information linking back to source documents, enabling evidence-grounded responses.

\textbf{Versioning and Updates.} As new papers are added to the corpus, we incrementally update the knowledge graph by detecting new entities, merging duplicates, and updating citation counts. Entity linking ensures that mentions in new papers are connected to existing canonical entities, maintaining graph coherence over time.


\subsection{Theoretical Foundations}
\label{sec:theory}

Our scientific KG of Section~\ref{subsec:knowledge_network} and the General-KG mode family of Section~\ref{sec:general_kg} use the same entity identifier space. Each node has a stable entity identifier, while canonical names and aliases are stored as attributes rather than used as join keys. This subsection explains why that design is useful for retrieval. The statements below describe the candidate stage before final ranking or truncation, and assume that entity identifiers have been assigned correctly. Errors from extraction, entity linking, and ranking are evaluated empirically in Section~\ref{sec:eval}. Proofs and constructive details are deferred to Appendix~\ref{sec:appendix_formal}.

We treat the extraction of a document $D$ as a single payload $\mathcal{P}_D=(\mathcal{V},\mathcal{E}_h)$, where $\mathcal{V}$ is a finite set of typed nodes with globally unique identifiers and $\mathcal{E}_h\subseteq\bigsqcup_{k\ge 2}\mathcal{V}^k$ is a set of typed hyperedges. Nodes and edges also carry evidence spans and chunk-level provenance, which we omit from the notation. A \textbf{view} is a deterministic map $\Phi_v\!:\mathcal{P}_D\mapsto\mathcal{P}_v$ that filters edges, qualifiers, or types but does not rename nodes. In particular, $\mathcal{V}_v=\mathcal{V}$ for every view $v$. Our scientific KG and the six modes of Table~\ref{tab:gkg_modes} are all views in this sense, and the \textit{union view} $\mathcal{P}_\cup\triangleq\bigsqcup_{v\in V}\Phi_v(\mathcal{P}_D)$ keeps the same node identifiers.

\begin{definition}[ID-Respecting Retriever]\label{def:retriever}
A candidate retriever $\Phi_R$ is \textit{ID-respecting} if it uses node identifiers as join keys and is monotone in the edge set, i.e.\ $\mathcal{G}_1\subseteq\mathcal{G}_2$ implies $\Phi_R(\mathcal{G}_1,q)\subseteq\Phi_R(\mathcal{G}_2,q)$ before final ranking and truncation. BM25 over canonical names, dense retrieval over name embeddings, and graph traversal qualify when their candidate sets are joined by stable node identifiers.
\end{definition}

The first design choice is to key every view by a globally unique node identifier rather than by canonical name. Surface-form alignment, the natural alternative, has a worse asymptotic cost and exposes the system to false merges whenever distinct entities share the same string. The following proposition makes both observations precise.
\begin{proposition}[Identifier Preserving Joins]\label{prop:id}
Let $\Phi_u,\Phi_v$ be two views and $K\subseteq\mathcal{V}$ a node-key set. If both views preserve the same stable identifiers, the cross-view join $\Phi_u(\mathcal{P})\bowtie_{\mathcal{V}}\Phi_v(\mathcal{P})$ on $K$ can be implemented as a hash join in $\mathcal{O}(|K|)$ time. If two views are produced by independent pipelines and have no shared identifiers, alignment must rely on surface form comparison. For unconstrained pairwise similarity without a blocking key or metric structure, the worst-case cost is $\Omega(|\mathcal{V}_u|\cdot|\mathcal{V}_v|)$. Let $\sigma(v)$ denote the surface form of node $v$, and let $H=\{v\in\mathcal{V}:|\sigma^{-1}(\sigma(v))|\ge 2\}$ be the homonym set. When $H$ is nonempty, matchers that merge identical surface forms can introduce false merges:
\begin{equation}
\Pr_{v\sim H,\,w\sim\mathcal{V}}\!\big[w\neq v,\ \sigma(w)=\sigma(v)\big]\;=\;\frac{\sum_{v\in H}(|\sigma^{-1}(\sigma(v))|-1)}{|H|\cdot|\mathcal{V}|}.
\label{eq:false_merge}
\end{equation}
\end{proposition}

A single binary projection (subject--predicate--object) collapses any fact that involves three or more participants, for example a temporal qualifier or a multi-argument experimental condition. We next show that the union view never reaches fewer nodes than a single view, and reaches strictly more whenever the gold reasoning path traverses a hyperedge of arity at least three.
\begin{proposition}[Cross View Reachability]\label{prop:reach}
Let $R_h(\mathcal{G},q)\subseteq\mathcal{V}$ be the set of nodes reachable in at most $h$ hops from a query-anchored seed $S(q)$ in $\mathcal{G}$, with each hyperedge of arity $k\ge 2$ counting as one hop connecting all of its endpoints. Then for every query $q$:
\begin{enumerate}
  \item \textbf{Monotonicity.} The union view contains every node reachable in any single view: $R_h(\mathcal{P}_\cup,q)\supseteq R_h(\Phi_v(\mathcal{P}_D),q)$ for every $v$.
  \item \textbf{Strict gap.} If a gold reasoning path uses a hyperedge of arity at least three, a binary projection may hide some endpoints of that edge. Let $H_{\ge3}^{(h)}(q)$ be the gold $\ge\!3$-arity hyperedges that lie on a gold $h$-hop reasoning path. For each $e\in H_{\ge3}^{(h)}(q)$, let $B_e$ be the endpoints of $e$ that are reachable from $S(q)$ only through $e$ and are not retained by the binary view. Then
  \begin{equation}
    \big|R_h(\mathcal{P}_\cup,q)\setminus R_h(\Phi_b(\mathcal{P}_D),q)\big|\;\ge\;\Big|\bigcup_{e\in H_{\ge3}^{(h)}(q)} B_e\Big|.
    \label{eq:reach_gap}
  \end{equation}
  In the common two-anchor case with disjoint hidden endpoint sets, the right-hand side equals $\sum_{e\in H_{\ge3}^{(h)}(q)}(|e|-2)$.
\end{enumerate}
Recovering the hidden higher-arity structure from the binary view alone is hard in the worst case; Appendix~\ref{sec:appendix_formal} gives the reduction.
\end{proposition}

The reachability gap above translates directly into a candidate-stage recall bound. The union view's recall is at least the best single-view recall plus the fraction of gold answers that only the joined view exposes, which is exactly the quantity stress-tested in Section~\ref{sec:eval}.
\begin{proposition}[Candidate Coverage]\label{prop:coverage}
Let $\Phi_R$ be an ID-respecting candidate retriever. Let $A_h^\star(q)\subseteq\mathcal{V}$ be a nonempty gold $h$-hop answer set, and let $\rho_h(\mathcal{G},q)=|\Phi_R(\mathcal{G},q)\cap A_h^\star(q)|/|A_h^\star(q)|$ be the candidate-stage recall. Define
\[
\Delta_h(q)=\frac{\big|A_h^\star(q)\cap\big(\Phi_R(\mathcal{P}_\cup,q)\setminus \bigcup_v \Phi_R(\Phi_v(\mathcal{P}_D),q)\big)\big|}{|A_h^\star(q)|},
\]
the fraction of gold answers retrieved only after the views are joined. Then
\begin{equation}
\rho_h(\mathcal{P}_\cup,q)\;\ge\;\max_{v}\,\rho_h(\Phi_v(\mathcal{P}_D),q)\;+\;\Delta_h(q),
\label{eq:recall_bound}
\end{equation}
where $\Delta_h(q)\ge 0$. Proposition~\ref{prop:reach} gives one condition under which this term is positive: the gold path uses evidence that is visible in the union view but hidden by a single binary projection. Thus, adding an identifier-preserving view can expand the candidate evidence available before the final ranker selects its top results.
\end{proposition}

Together, these propositions state a simple design principle. Stable identifiers make evidence from different views easy to join; multiple views expose evidence that a single binary projection may miss; and the joined candidate set can cover more gold evidence before ranking. The witness construction in Appendix~\ref{sec:appendix_formal} turns the strict gap term into a concrete stress test, and the queries used in Section~\ref{sec:eval} are constructed to exercise it.

\section{LLM Layer: Reinforcement-Learned Information Extraction Backbone}
\label{sec:ie-grpo}

The knowledge graph and scientific knowledge network described in Sections~\ref{subsec:hierarchical_kg} provide the structural scaffolding for downstream retrieval and reasoning, but their practical utility is tightly coupled to the accuracy of the information extraction module that populates them. Noisy or incomplete entity and relation extraction propagates directly into retrieval errors and degraded agent decisions. Rather than rely on a general-purpose instruction model for this step, we train a compact, domain-specialised IE backbone via reinforcement learning with a rule-based reward. The resulting model starts from \textit{Qwen3-4B-Instruct} and is optimised with GRPO. We describe the algorithm, the reward design, and the training data below.

\subsection{Training Algorithm}

At each step $t$, the current policy $\pi_{\theta_{\mathrm{old}}}$ samples
a batch of $B$ prompts from the training distribution $\mathcal{D}$.
For every prompt $q_b$, we draw $G$ independent rollouts
$\{o_{b,g}\}_{g=1}^G$ and score each rollout with a rule-based reward
$r_{b,g} = R(o_{b,g}, y_b^\star, \tau_b)$, where $y_b^\star$ is the
ground-truth extraction and $\tau_b$ is the per-sample task type
(\texttt{NER}, \texttt{RE}, or \texttt{STRUCT\_NER}). The advantage is
obtained by standardising rewards within each group:
\begin{equation}
    \hat{A}_{b,g} \;=\; \frac{r_{b,g} - \mu_b}{\sigma_b + \epsilon},
    \qquad
    \mu_b = \tfrac{1}{G}\textstyle\sum_{g} r_{b,g},
    \quad
    \sigma_b^2 = \tfrac{1}{G}\textstyle\sum_{g}(r_{b,g} - \mu_b)^2.
\end{equation}
No value network is trained; the group means serve as the baseline.
The policy is updated with the clipped PPO objective, regularised by a
low-variance KL penalty to the frozen reference $\pi_{\mathrm{ref}} =
\pi_{\theta_0}$ that is added as a loss term rather than folded
into the reward:
\begin{align}
    \mathcal{L}_{\mathrm{PG}}(\theta) &= -\,\mathbb{E}\!\left[
        \min\!\Big(
            \rho_{b,g}\,\hat{A}_{b,g},\;
            \mathrm{clip}(\rho_{b,g},\,1-\epsilon_{\mathrm{clip}},\,1+\epsilon_{\mathrm{clip}})\,\hat{A}_{b,g}
        \Big)\right], \\
    \mathcal{L}_{\mathrm{KL}}(\theta) &= \beta\cdot
        \widehat{\mathrm{KL}}_{\mathrm{lv}}\!\left(\pi_\theta\,\big\|\,\pi_{\mathrm{ref}}\right), \\
    \mathcal{L}(\theta) &= \mathcal{L}_{\mathrm{PG}}(\theta) + \mathcal{L}_{\mathrm{KL}}(\theta),
\end{align}
where
$\rho_{b,g} = \pi_\theta(o_{b,g}\!\mid\!q_b)\,/\,\pi_{\theta_{\mathrm{old}}}(o_{b,g}\!\mid\!q_b)$.
The full procedure is summarised in Algorithm~\ref{alg:ie-grpo}.

\begin{algorithm}[t]
\caption{IE-GRPO Training}
\label{alg:ie-grpo}
\begin{algorithmic}[1]
\Require base policy $\pi_{\theta_0}$, reference $\pi_{\mathrm{ref}} \leftarrow \pi_{\theta_0}$, dataset $\mathcal{D}$, reward $R$, batch size $B$, group size $G$, PPO epochs $K$, clip $\epsilon_{\mathrm{clip}}$, KL coeff.\ $\beta$, learning rate $\eta$.
\State Initialise $\pi_\theta \leftarrow \pi_{\theta_0}$, $\pi_{\mathrm{old}} \leftarrow \pi_\theta$.
\For{$t = 1, \ldots, T$}
    \State Sample prompts $\{q_b\}_{b=1}^B \sim \mathcal{D}$.
    \For{$b = 1, \ldots, B$} \Comment{rollout phase}
        \State Sample $G$ responses $\{o_{b,g}\}_{g=1}^G \sim \pi_{\mathrm{old}}(\cdot\!\mid\!q_b)$.
        \State Compute rewards $r_{b,g} = R(o_{b,g}, y_b^\star, \tau_b)$ for $g=1,\ldots,G$.
        \State Compute $\hat{A}_{b,g}$ via group-relative standardisation.
    \EndFor
    \For{ epoch $k = 1, \ldots, K$} \Comment{update phase}
        \For{each mini-batch $\mathcal{M}$ of $(q_b, o_{b,g})$}
            \State $\rho_{b,g} \leftarrow \pi_\theta(o_{b,g}\!\mid\!q_b)\,/\,\pi_{\mathrm{old}}(o_{b,g}\!\mid\!q_b)$.
            \State Form $\mathcal{L}_{\mathrm{PG}}$ and $\mathcal{L}_{\mathrm{KL}}$ as in Eqs.\,(4)--(5).
            \State $\theta \leftarrow \theta - \eta\,\nabla_\theta(\mathcal{L}_{\mathrm{PG}} + \mathcal{L}_{\mathrm{KL}})$.
        \EndFor
    \EndFor
    \State $\pi_{\mathrm{old}} \leftarrow \pi_\theta$.
\EndFor
\end{algorithmic}
\end{algorithm}

\subsection{Reward Function}

Let $\hat{y}$ denote the content parsed from the \verb|<answer>...</answer>|
span of a rollout $o$. We define a rule-based reward
$R(o, y^\star, \tau) \in [0,1]$ as the sum of three components:
\begin{align}
    R(o, y^\star, \tau) &= R_{\mathrm{fmt}}(o) \,+\, R_{\mathrm{json}}(o) \,+\, R_{\mathrm{F1}}(o, y^\star, \tau), \\[2pt]
    R_{\mathrm{fmt}}(o) &= 0.1\cdot\mathbbm{1}\!\left[\langle\texttt{think}\rangle\text{ present}\right]
                      + 0.1\cdot\mathbbm{1}\!\left[\langle\texttt{answer}\rangle\text{ present}\right], \\
    R_{\mathrm{json}}(o) &= 0.1\cdot\mathbbm{1}\!\left[\hat{y}\text{ parses as \texttt{dict}}\right]
                      + 0.05\cdot\mathbbm{1}\!\left[\hat{y}\text{ is valid JSON but not \texttt{dict}}\right], \\
    R_{\mathrm{F1}}(o, y^\star, \tau) &= 0.7 \cdot \mathrm{F1}_{\tau}(\hat{y}, y^\star).
\end{align}
The format and JSON terms provide a dense shaping signal that is easy to
optimise early in training, while the task-specific $\mathrm{F1}_\tau$
term carries the semantic signal. The F1 computation is dispatched by
the per-sample task type $\tau$:
\begin{itemize}
    \item \textbf{$\tau = \texttt{NER}$.} Each value in $\hat{y}$ and $y^\star$
    is a list of entity strings keyed by entity type. Let
    $E(y) = \{(t, \textsc{norm}(e)) : t \in \mathrm{keys}(y),\, e \in y[t]\}$,
    where $\textsc{norm}$ lower-cases and trims the string. The score is
    the set-level $\mathrm{F1}$ between $E(\hat{y})$ and $E(y^\star)$.

    \item \textbf{$\tau = \texttt{RE}$.} Each value is a list of relation
    triples with \texttt{subject}/\texttt{object} fields. Let
    $T(y) = \{(r,\, \textsc{norm}(s),\, \textsc{norm}(o))\}$. The score
    is the set-level $\mathrm{F1}$ between $T(\hat{y})$ and $T(y^\star)$.

    \item \textbf{$\tau = \texttt{STRUCT\_NER}$.} Used for long-form
    structured extraction with deeply nested schemas. For each top-level
    key $k$ in $y^\star$, we recursively flatten the subtree into a set
    of $(\text{path},\,\textsc{norm}(\text{leaf}))$ tuples, filtering
    placeholder leaves such as \texttt{N/A}, \texttt{none}, \texttt{null},
    and compute a per-key $\mathrm{F1}_k$. The final score is the
    macro-average over top-level keys,
    $\mathrm{F1}_{\texttt{STRUCT\_NER}} = \tfrac{1}{|\mathrm{keys}(y^\star)|}\sum_k \mathrm{F1}_k$,
    which prevents a single large branch from dominating the signal.
\end{itemize}
On top of this, set-level $\mathrm{F1}$ follows the standard definition
$\mathrm{F1}(A,B) = 2|A\cap B|/(|A|+|B|)$, with both-empty sets treated
as a perfect match.


\section{CLI Layer: A Knowledge-Grounded Multi-Agent Research Toolkit}
\label{sec:agent}

Sections~\ref{subsec:hierarchical_kg} through~\ref{sec:theory} build the knowledge infrastructure: a multimodal hierarchical graph, a schema-adaptive extension to arbitrary documents (General-KG), an information-extraction backbone, and a formal account of how cross-view information linking expands candidate evidence. This section turns that infrastructure into a research agent. Our CLI departs from monolithic autonomous research agents along three axes. First, a tri-source retrieval mechanism combines web search, multimodal graph anchors, and cross-document network traversal under a shared evidence schema, using stable node identifiers when a source can be linked to the graph (Definition~\ref{def:retriever}). Second, a small set of typed operators over the multimodal graph is exposed to the agent as deterministic primitives, so verifiable graph computation is separated from open-ended language reasoning. Third, a multi-role swarm execution layer composed of a coordinator, specialised workers, and an aggregator produces inspectable run artifacts rather than a single conversational answer.

The motivation is practical. A scientific task typically contains heterogeneous work that fails in different ways: retrieving recent papers, inspecting figures and tables, recovering method lineage, reading code, judging novelty, drafting a survey, and occasionally producing a prototype. Hiding all of this inside one chat loop makes the final answer hard to audit. Promoting it to a swarm of evidence-grounded roles makes intermediate objects (plans, evidence bundles, worker artifacts, manifests) first-class. The same graph can therefore support paper survey, code understanding, idea grounding, idea generation, trend prediction, and idea-to-prototype workflows under a single execution model. For graph-linked evidence, the design realises the ID-respecting retriever of Definition~\ref{def:retriever}, and so inherits the candidate coverage property of Proposition~\ref{prop:coverage} before final ranking and truncation.

\subsection{Tri-Source Knowledge Retrieval and Fusion}
\label{subsec:tri_source}

A fundamental limitation of existing autonomous research systems lies in their reliance on a single knowledge source. Web search gives broad coverage and recency but returns noisy, document-level results. Vector retrieval gives semantic recall but no structural reasoning. Pure graph traversal gives precision but no link to external recency. Our CLI uses a tri-source mechanism in which each source exposes a typed tool interface, and the agent decides, on a per-query basis, which sources to invoke and how to combine the evidence. Records that resolve to existing graph entities are keyed by the same stable node identifiers (Section~\ref{sec:theory}), so their cross-source fusion is realised as a hash join rather than a fuzzy match. Web results that cannot yet be linked to a graph node are retained as document-level evidence and are not used in ID-based joins.

\paragraph{Web Search ($\mathcal{S}_{\mathrm{web}}$).}
The web module accesses recent publications, preprints, and technical reports through academic search services such as arXiv, Semantic Scholar, and Google Scholar. Given a query $q$, it decomposes $q$ into keyword combinations and scores each retrieved document $d$ by
\begin{equation}
R_{\mathrm{web}}(d, q) \;=\; \alpha_{\mathrm{title}}\,\mathrm{sim}(d_{\mathrm{title}}, q)\;+\;\alpha_{\mathrm{abs}}\,\mathrm{sim}(d_{\mathrm{abstract}}, q),
\label{eq:r_web}
\end{equation}
where $\mathrm{sim}(\cdot,\cdot)$ is a dense-embedding cosine similarity, and $(\alpha_{\mathrm{title}},\alpha_{\mathrm{abs}})$ are normalised weights with default $(0.6, 0.4)$. The web module excels at recency: it surfaces work too new to have been ingested by the graph builder and provides the primary check for claims that fall outside the current graph snapshot.

\paragraph{Multimodal Graph Retrieval ($\mathcal{S}_{\mathrm{mmkg}}$).}
The multimodal graph of Section~\ref{subsec:hierarchical_kg} enables retrieval beyond text matching. For a query $q$, we identify relevant semantic anchors via a hybrid scheme that combines dense embedding similarity with lexical matching:
\begin{equation}
\mathcal{A}_{\mathrm{relevant}}(q) \;=\; \big\{\,a_j \,:\, \mathrm{sim}(\mathbf{e}_q,\mathbf{e}_{a_j}) > \tau_{\mathrm{anchor}} \;\lor\; \mathrm{lex}(q,\mathrm{desc}(a_j)) > \tau_{\mathrm{lex}}\,\big\},
\label{eq:a_relevant}
\end{equation}
where $\mathbf{e}_q,\mathbf{e}_{a_j}$ are query and anchor embeddings and $\mathrm{lex}(\cdot,\cdot)$ is a token-overlap score over normalised labels and descriptions. Each retrieved anchor carries provenance $\langle\mathrm{doc\_id},\mathrm{page},\mathrm{bbox}\rangle$ for figure and table content, and $\langle\mathrm{file\_path},\mathrm{symbol},\mathrm{line\_span}\rangle$ for code anchors. The MMKG is the source through which figures, tables, and equations enter the agent loop as first-class evidence rather than caption-only proxies. 

\paragraph{Knowledge Network Traversal ($\mathcal{S}_{\mathrm{kn}}$).}
The scientific knowledge network of Section~\ref{subsec:knowledge_network} supports cross-document reasoning over the heterogeneous graph of Modules~A through E linked by typed relations such as \texttt{cites}, \texttt{proposes\_method}, \texttt{uses\_method}, \texttt{evaluates\_on}, \texttt{compares\_to}, and \texttt{reports\_metric}. The traversal layer exposes deterministic graph primitives (Section~\ref{subsec:operators_swarm}) that the agent invokes as tools, providing precision and reproducibility while leaving open-ended reasoning to the LLM.

\paragraph{Fusion.}
The agent fuses the three sources into a typed bundle
\begin{equation}
\mathcal{K}_{\mathrm{fused}}(q) \;=\; \operatorname{TopK}_{e\in \mathcal{S}_{\mathrm{web}}(q) \cup \mathcal{S}_{\mathrm{mmkg}}(q) \cup \mathcal{S}_{\mathrm{kn}}(q)}
\bigl[\lambda_w\,s_w(e) + \lambda_m\,s_m(e) + \lambda_k\,s_k(e)\bigr],
\label{eq:fuse}
\end{equation}
where $s_w,s_m,s_k$ are per-source normalised scores in $[0,1]$, missing source scores are set to zero, $(\lambda_w,\lambda_m,\lambda_k)$ are non-negative weights with default $(0.30,0.40,0.30)$, and the resulting set retains source labels, evidence type, reference id, and provenance fields so that the aggregator can later show which figure, paragraph, citation path, or code symbol supported each output. A lightweight intent classifier maps $q$ to one of five categories, namely \texttt{recency}, \texttt{multimodal}, \texttt{lineage}, \texttt{comparative}, and \texttt{general}, and re-balances the weights accordingly: \texttt{recency} uses $(0.70,0.15,0.15)$; \texttt{multimodal} uses $(0.15,0.70,0.15)$; \texttt{lineage} and \texttt{comparative} use $(0.20,0.20,0.60)$; \texttt{general} keeps the default. The classifier is a few-shot prompt over a frozen LLM and is called once per query, so the routing overhead is bounded by a single LLM call regardless of how many fan-out retrievals follow.

\paragraph{Theoretical anchor.}
For entries that are linked to graph nodes, Equation~\eqref{eq:fuse} is an instance of the ID-respecting retriever family of Definition~\ref{def:retriever}. Every linked entry $e$ is keyed by a stable node identifier from the shared vertex space $\mathcal{V}$ of the universal payload, so cross-source joins are deterministic hash lookups (Proposition~\ref{prop:id}). Consequently, Proposition~\ref{prop:coverage} implies that the fused candidate set for $\mathcal{S}_{\mathrm{web}} \cup \mathcal{S}_{\mathrm{mmkg}} \cup \mathcal{S}_{\mathrm{kn}}$ is at least as large in gold-answer coverage as the best linked single source, before final ranking and truncation. It becomes strictly larger when the joined sources retrieve additional gold candidates, i.e.\ when $\Delta_h(q)>0$. Unlinked web records remain useful evidence, but they are outside this identifier-based guarantee until entity linking assigns them to $\mathcal{V}$.

\subsection{Graph Operators and Multi-Agent Coordination}
\label{subsec:operators_swarm}

The agent does not operate on the graph as an unstructured text dump. Instead, it invokes a small set of typed operators that take a typed input and emit a typed output with explicit provenance, and a swarm execution layer composes these operators into research workflows. We describe the operator set first, then the swarm runtime that orchestrates them, and finally the tool surface through which both are exposed.

\paragraph{(O1) Seed Resolution.}
The deterministic primitive maps a candidate mention string to a canonical node set $\mathcal{V}_{\mathrm{seed}}\subseteq\mathcal{V}$ via case-insensitive label matching with degree-based disambiguation. Synonym expansion and contextual disambiguation are handled by the agent, which may issue multiple deterministic resolution calls before committing to a seed set.

\paragraph{(O2) Citation Lineage Reconstruction.}
Forward and backward traversal along \texttt{cites} edges, plus shortest-path queries weighted by lineage-bearing relations such as \texttt{BUILDS\_ON}, \texttt{EXTENDS}, \texttt{DERIVES\_FROM}, and \texttt{cites}, recovers method evolution chains across papers. The output is an ordered list $\langle p_1,r_1,p_2,r_2,\ldots,p_T\rangle$ of papers and the typed relations that connect them.

\paragraph{(O3) Comparative Baseline Retrieval.}
Given a dataset $D$ and an optional metric $M$, returns all triples $(P,m,v)$ such that $P\xrightarrow{\texttt{evaluates\_on}}D$ and $P\xrightarrow{\texttt{proposes\_method}}m$. If $M$ is provided, the query keeps only records with a matching \texttt{reports\_metric} edge; otherwise $v$ is returned when any reported value is available and marked missing when it is not. This realises the request ``find all methods evaluated on $D$ under $M$'' as a single graph query, without per-paper LLM inspection.

\paragraph{(O4) Multimodal Anchor Retrieval.}
Implements Eq.~\eqref{eq:a_relevant} and returns the relevant anchor set $\mathcal{A}_{\mathrm{relevant}}(q)$ together with the original payloads (cropped figure regions, serialised tables, LaTeX equations, code symbols) and bounding-box or line-span provenance.

\paragraph{(O5) Gap Detection.}
Surfaces structural indicators of under-explored regions: (i) orphan methods proposed but not reused, (ii) singleton datasets, (iii) papers disconnected from the main component, and (iv) sparse cells in the Method-Task projection $(t,\mathcal{M}_t)$ where $|\mathcal{M}_t| < \tau_{\mathrm{cov}}$. The output is a prioritised list of gap descriptors that the IdeaWorker can target.

\paragraph{(O6) Idea Grounding and Novelty Judging.}
Given a candidate idea $I$, it retrieves the top-$k$ most related methods $\mathcal{R}_k(I)$ via O1 and O2, then asks an LLM judge $\mathcal{J}_{\mathrm{LLM}}$ to score methodological overlap on a structured rubric covering problem formulation, algorithmic mechanism, training strategy, and target domain:
\begin{equation}
\mathrm{Novelty}(I) \;=\; \mathcal{J}_{\mathrm{LLM}}\!\bigl(I \,\big|\, \mathcal{R}_k(I)\bigr) \;\in\; [0,1].
\label{eq:novelty}
\end{equation}
This formulation is intended to capture method-level novelty more directly than embedding cosine similarity, which can conflate surface form with methodological substance.

The split between deterministic operators and agent-mediated composition keeps each layer simple. Graph operations are efficient and verifiable, and the agent absorbs synonym handling, multi-step planning, and open-ended reasoning. Several capabilities therefore emerge from composition rather than from any single module: lexical seed resolution (O1) becomes semantic retrieval when the agent expands queries into synonyms; single-hop primitives (O2, O3) become multi-hop research workflows when the agent chains them with intermediate filtering; and structural retrieval (O5) becomes methodological novelty judgement when paired with an LLM rubric (O6).

\paragraph{Swarm runtime.}
We formalise a swarm run as follows. Given a graph $G$, a user task $q$, and an execution mode $m\in\{\texttt{code-wiki},\texttt{survey},\texttt{idea-loop},\texttt{all}\}$, the coordinator emits a typed plan
\begin{equation}
\mathcal{P} \;=\; \mathrm{Coord}(G, q, m) \;=\; \{\,j_i = (r_i, x_i, o_i, d_i)\,\}_{i=1}^{n},
\label{eq:plan}
\end{equation}
where $r_i$ is the worker role, $x_i$ is the payload, $o_i$ is the output contract, and $d_i$ is an optional dependency set. Each worker receives a compact evidence bundle drawn from $\mathcal{K}_{\mathrm{fused}}$ rather than the whole graph. The aggregator then writes
\begin{equation}
\mathcal{A} \;=\; \mathrm{Agg}\!\bigl(\mathcal{P},\,\{y_i\}_{i=1}^{n},\,\{E_i\}_{i=1}^{n}\bigr),
\label{eq:agg}
\end{equation}
where $y_i$ is the worker output and $E_i$ is the evidence used by that worker. The multi-role decomposition by itself is not novel, and we do not claim improved reasoning quality from the swarm structure. The three properties of Eq.~\eqref{eq:plan} and Eq.~\eqref{eq:agg} that matter for the rest of this report are concrete and narrow. First, the output contract $o_i$ in $\mathcal{P}$ is a typed schema rather than a free-form prompt, so worker outputs are checked against $o_i$ before being accepted by the aggregator, and a failed contract becomes a routable signal rather than a silent textual error. Second, the run object $\mathcal{A}$ is a manifest with on-disk artifacts, not conversation history, so a single worker can be rerun, a single evidence bundle replaced, or a single artifact audited without replaying the full session. Third, the dependency set $d_i$ makes failure isolation computable: when a worker fails, the affected sub-tree is read off $d_i$ rather than reconstructed from a chat trace, so recovery cost scales with the size of the dependent sub-graph rather than with the size of the whole plan. Worker-to-worker negotiation, debate, and majority voting are not part of the design, and the aggregator does not arbitrate disagreements between workers beyond accepting or rejecting outputs against $o_i$.

\paragraph{Roles.}
The Coordinator inspects the graph and the task and emits a plan. The CodeWikiWorker reads code communities and writes repository documentation. The SurveyWorker clusters paper nodes and writes topic sections. The IdeaWorker performs grounding (via O6), generation, critique, refinement, and novelty judging. The PrototypeWorker converts a selected idea into a method specification and a code scaffold. The Aggregator merges artifacts, writes the manifest, and surfaces failures rather than hiding them in a final summary. On worker failure, the Coordinator either retries the failing job with an enlarged evidence bundle for transient errors or replans the affected sub-tree for persistent errors. Failed jobs are recorded in $\mathcal{A}$ verbatim so that downstream consumers can inspect them.

\paragraph{Tool surface.}
The deterministic graph primitives are exposed to the LLM through three concrete interfaces: a Python API for direct programmatic use, a unified command-line interface, and a Model Context Protocol (MCP) server that registers each primitive as a typed tool. We provide integrations for Claude Code via an \texttt{/graphanything} skill and for Nano-Claude-Code, while the MCP interface is designed to be provider-agnostic for agents that support typed tool calls. Table~\ref{tab:tool_catalog} summarises the complete tool surface, organised into four functional categories.

\begin{table}[t]
\centering
\caption{Tool catalogue exposed by the CLI runtime. Each command is also available as a Python function and as an MCP tool.}
\label{tab:tool_catalog}
\small
\setlength{\tabcolsep}{4pt}
\begin{tabular}{@{}p{0.40\linewidth}p{0.56\linewidth}@{}}
\toprule
\textbf{Command} & \textbf{Function} \\
\midrule
\multicolumn{2}{l}{\textit{Ingestion}} \\
\midrule
\texttt{graphanything ingest-kg merged\_*.json} & Convert structured paper extraction JSON into a paper graph. \\
\texttt{graphanything add <url>}                & Ingest a paper or code repository from a URL. \\
\midrule
\multicolumn{2}{l}{\textit{Paper-level query (core primitives)}} \\
\midrule
\texttt{graphanything find-paper "<keyword>"}   & Search Paper nodes by keyword, DOI, arXiv ID, or author. \\
\texttt{graphanything paper-details "<ref>"}    & Return a structured card: methods, datasets, contributions, citations. \\
\texttt{graphanything citations-of "<ref>"}     & List papers cited by the given paper (outgoing). \\
\texttt{graphanything citations-to "<ref>"}     & List papers that cite the given paper (incoming). \\
\texttt{graphanything lineage "<A>" "<B>"}      & Shortest research-evolution path between two concepts. \\
\texttt{graphanything baselines "<dataset>"}    & Methods evaluated on a dataset, optionally under a metric. \\
\texttt{graphanything tri-search "<q>"}         & Route a query over web, multimodal anchors, and the knowledge network. \\
\midrule
\multicolumn{2}{l}{\textit{Academic-metric analysis}} \\
\midrule
\texttt{graphanything influential-methods}      & Rank methods by structural influence in the graph. \\
\texttt{graphanything bridging-papers}          & Identify papers that bridge distinct research communities. \\
\texttt{graphanything lineage-depth}            & Compute the longest extension chain rooted at each method. \\
\midrule
\multicolumn{2}{l}{\textit{Graph evolution}} \\
\midrule
\texttt{graphanything evolve absorb}            & Fold answered questions back into the graph as Question nodes. \\
\texttt{graphanything evolve compress}          & Surface near-duplicate nodes as merge candidates. \\
\texttt{graphanything evolve gaps}              & Detect orphan methods, singleton datasets, missing citations, detached papers. \\
\bottomrule
\end{tabular}
\end{table}

This split, with deterministic primitives below and agent-mediated composition above, keeps each layer simple: graph operations remain efficient and verifiable, while the agent layer absorbs synonym handling, multi-step planning, and open-ended reasoning. As a result, several capabilities of our framework emerge from composition rather than being implemented in any single module, which we treat as an explicit design choice rather than as an emergent property of prompt engineering.

\subsection{Idea-to-Experiment Pipeline}
\label{subsec:idea_pipeline}

Our CLI transforms research questions into method specifications and code prototypes through an iterative pipeline that couples knowledge retrieval, idea generation, novelty assessment, and code synthesis. Given a task description, the agent first retrieves background knowledge through the tri-source mechanism of Section~\ref{subsec:tri_source}. Retrieval proceeds from a broad survey stage to a focused deep-analysis stage, so that figures, tables, equations, and implementation details enter the loop together with prose.

\paragraph{Iterative idea refinement.}
The IdeaWorker generates diverse candidates and refines them through iterative critique:
\begin{equation}
I^{(t+1)} \;=\; G_{\mathrm{LLM}}\!\bigl(I^{(t)},\,C^{(t)},\,\mathcal{K}_{\mathrm{new}}\bigr),
\label{eq:idea_iter}
\end{equation}
where $C^{(t)}$ is the structured critique at iteration $t$ and $\mathcal{K}_{\mathrm{new}}$ is the freshly retrieved evidence. Initial iterations sample with high diversity; later iterations emphasise specificity and feasibility by incorporating targeted feedback from $C^{(t)}$ and from new evidence retrieved in response to gaps surfaced at iteration $t$. Each candidate is assessed along four dimensions: coherence, credibility, feasibility, and novelty (Eq.~\ref{eq:novelty}). Novelty is grounded against the knowledge network rather than judged from the idea text alone, which helps steer the loop toward ideas that are specific, feasible, and easier to audit against prior work.

\paragraph{From idea to method specification and code.}
A promising idea is translated by the PrototypeWorker into a structured method specification covering problem formulation, algorithm design, and training strategy. The specification is iteratively refined for clarity and completeness against the same novelty rubric. The methodology is then converted into code, using existing repositories retrieved through the paper-to-code bridge of Section~\ref{subsec:knowledge_network} as scaffolding. For complex methods, the system decomposes the design into modular components and validates them incrementally,
\begin{equation}
\mathcal{M} \;=\; \{m_1,\ldots,m_k\} \;\Longrightarrow\; \mathcal{C} \;=\; \{C_1,\ldots,C_k\},
\label{eq:modular_synth}
\end{equation}
with errors resolved through an exception-guided debugging loop. The PrototypeWorker emits a prototype repository together with a smoke test that exercises the main entry point. Full reproduction of reported numerical results is not in scope of the smoke test and is treated as a downstream evaluation step rather than as a closed-loop signal at idea time. Accepted ideas, generated specifications, and prototypes are written back into the graph as Hypothesis, MethodSpec, and Implementation nodes through the explicit \texttt{graphanything evolve} command, so the graph records both literature and agent activity, but the gating decision is taken by the user rather than by the agent itself.

\subsection{Downstream Tasks and Examples}
\label{subsec:downstream_examples}

The system is best understood through downstream tasks. Each example below is rendered as an output card produced by a CLI run and saved by the aggregator. The structured field labels reflect the actual artifact schema rather than free prose, and the examples are intended as illustrative outputs rather than as benchmark numbers; quantitative comparisons are reported in Section~\ref{sec:eval}.

\begin{table}[t]
\centering
\caption{Representative downstream tasks supported by our CLI.}
\label{tab:downstream_tasks}
\small
\setlength{\tabcolsep}{4pt}
\begin{tabular}{@{}p{0.26\linewidth}p{0.34\linewidth}p{0.34\linewidth}@{}}
\toprule
\textbf{Task} & \textbf{Active workers} & \textbf{Output artifact} \\
\midrule
Idea grounding and evaluation   & Coordinator, IdeaWorker, SurveyWorker            & Claims, evidence paragraphs, similar / different points, novelty risks. \\
Idea generation                 & IdeaWorker, SurveyWorker, Aggregator             & Candidate ideas with novelty, significance, risks, and key evidence. \\
Research trend prediction       & SurveyWorker, IdeaWorker, Aggregator             & Stage summaries, unresolved bottlenecks, future directions. \\
Paper-code bridge analysis      & CodeWikiWorker, SurveyWorker, Aggregator         & Mapping between method descriptions and implementation symbols. \\
Swarm execution audit           & Coordinator, all selected workers, Aggregator    & Plan, worker status, artifact paths, evidence ids, error messages. \\
\bottomrule
\end{tabular}
\end{table}

\begin{tcolorbox}[colback=blue!4, colframe=blue!50!black, title=Example: Idea Grounding and Evaluation, fonttitle=\bfseries\small, fontupper=\small, breakable]
\textbf{Task:} Decide whether a target idea has genuine novelty after grounding it in related work.\\[2pt]
\textbf{Target idea:} Evidence-Certified Scientific Idea Generation with a Coordinator-Worker-Aggregator Swarm.\\[2pt]
\textbf{Target claim from the idea:} A research agent should not only output a new idea; it should also separate supported claims, new combinations, feasibility risks, and implementation anchors, so that novelty can be audited.\\[2pt]
\textbf{Retrieved evidence:} Knowledge-graph-based scientific search retrieves papers and paragraphs for an idea, then compares the target claim with prior evidence across motivation, method, and experimental design.\\[2pt]
\textbf{Matching aspect:} Knowledge-grounded idea evaluation.\\[2pt]
\textbf{Similar point:} Both the target idea and the evidence treat idea evaluation as a grounding problem rather than as a pure language-model judgment, and both require retrieved evidence before judging novelty.\\[2pt]
\textbf{Different point:} The evidence focuses on paper-and-paragraph grounding; the target idea adds a swarm execution layer, code anchors, manifest records, and separate worker roles for grounding, critique, and aggregation.\\[2pt]
\textbf{Evaluation result:} The idea is not novel if stated only as retrieve-then-judge. Its stronger novelty comes from combining multimodal paper anchors, code-graph bridges, and an auditable swarm manifest into one research workflow.
\end{tcolorbox}

\begin{tcolorbox}[colback=green!4, colframe=green!50!black, title=Example: Idea Generation, fonttitle=\bfseries\small, fontupper=\small, breakable]
\textbf{Command:} \texttt{graphanything swarm run --mode all --task "evaluate evidence-certified idea generation"}.\\[2pt]
\textbf{Coordinator plan:} Launch one survey worker for the topic cluster, two code-wiki workers for implementation communities, and one idea worker for generation, critique, and novelty judging.\\[2pt]
\textbf{Worker outputs:} The survey worker writes a topic section with citation lineage; the code workers write module notes and identify implementation anchors; the idea worker writes three candidate ideas and a novelty comparison.\\[2pt]
\textbf{Aggregator output:} A manifest with job ids, status, artifact paths, evidence ids, and failure messages, plus a readable report summarising the survey, implementation map, idea cards, and novelty risks.\\[2pt]
\textbf{Audit value:} The user can rerun only the failed worker, replace one evidence bundle, or inspect the exact graph nodes that supported an idea, without re-issuing the entire prompt.
\end{tcolorbox}

\section{Experiments}
\label{sec:eval}

This section evaluates Agents-K1 along four axes that collectively cover its three components and the empirical claim attached to its theoretical foundation. We first characterise the corpus that Agents-K1 is built on, reporting domain coverage across six scientific disciplines (Section~\ref{subsec:domain_dist}) and the LLM-as-Judge evaluation protocol used throughout (Section~\ref{subsec:eval_metrics}). We then quantify our model extraction quality through the cross-domain breakdown of Section~\ref{subsec:cross_domain} and the head-to-head comparison against open-source extraction models at multiple parameter scales in Section~\ref{sec:ie-grpo-eval}. We turn next to the agent layer: Section~\ref{subsec:knowledgeable_research} validates our CLI on knowledge-grounded scientific question answering, including geoscience research questions and the \textsc{FrontierScience-Research} benchmark, and Section~\ref{subsec:open_source_bench} measures the same agent against nine graph-augmented retrieval baselines on HotpotQA, 2WikiMultiHopQA, and MuSiQue. 

\subsection{Domain Distribution and Coverage}
\label{subsec:domain_dist}
\begin{figure*}[htbp] 
    \centering
    \includegraphics[width=1\textwidth]{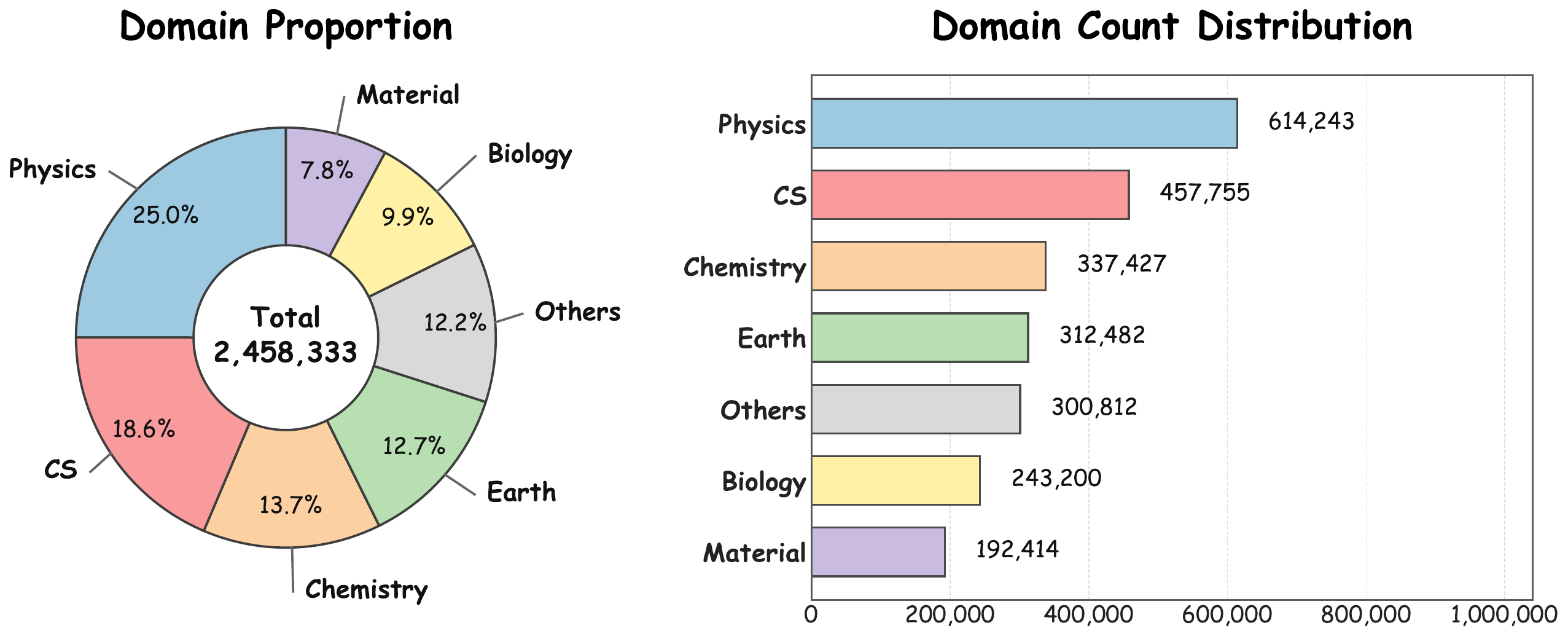}
    \caption{Domain Proportion and Count Distribution, where CS represents computer science.}
    \label{fig:domain}
\end{figure*}

To evaluate the scalability and domain adaptability of our framework, we constructed a large-scale heterogeneous scientific knowledge repository that ensures comprehensive coverage of major scientific disciplines. As illustrated in Figure ~\ref{fig:domain}, the dataset spans a diverse range of fields, predominantly encompassing Physics, Chemistry, Computer Science, Earth Science, Materials Science, Biology, and others, complemented by a significant collection of interdisciplinary literature. This broad domain distribution is critical for verifying the model's robustness, ensuring it can effectively handle the varied terminologies, reasoning patterns, and data modalities inherent to different scientific contexts.

\subsection{Evaluation Metrics}\label{subsec:eval_metrics}

To comprehensively assess the quality of the constructed knowledge graph, we employ a rigorous \textbf{LLM-as-a-Judge} evaluation protocol. Unlike traditional exact-matching metrics, our framework utilizes a state-of-the-art large language model (DeepSeek-V3~\cite{deepseekai2025deepseekv3technicalreport}) acting as an ``Expert Scientific Editor'' to evaluate the extraction results based on semantic correctness and scientific essence.

\subsubsection{Metric Formulations}
We quantify performance using three standard metrics: \textbf{Precision ($P$)}, \textbf{Recall ($R$)}, and the \textbf{F1-score ($F1$)}.
Let $N_{ext}$ denote the total number of items extracted by the model. Through the LLM evaluation process, we identify $N_{err}$ as the count of incorrect extractions (hallucinations or factual errors) and $N_{miss}$ as the count of critical information omitted from the source text.

Accordingly, the number of correctly extracted items (True Positives) is defined as $N_{correct} = N_{ext} - N_{err}$. The estimated ground truth size is derived as the sum of correct items and missed items. The metrics are formulated as follows:

\begin{equation}
    P = \frac{N_{ext} - N_{err}}{N_{ext}}, \quad
    R = \frac{N_{ext} - N_{err}}{ (N_{ext} - N_{err}) + N_{miss}}, \quad
    F1 = \frac{2 \cdot P \cdot R}{P + R}
\end{equation}

\begin{itemize}[leftmargin=*, align=left]
    \item \textbf{Precision ($P$)}: Reflects the \textit{trustworthiness} of the system by penalizing hallucinations and factual contradictions.
    \item \textbf{Recall ($R$)}: Reflects the \textit{completeness} of the system by penalizing the omission of core scientific concepts.
    \item \textbf{F1-score}: Serves as the harmonic mean to provide a holistic performance index.
\end{itemize}

\subsubsection{Semantic-Aware Evaluation Criteria}

To address the inherent complexity and heterogeneity of scientific literature, we developed a multi-dimensional evaluation protocol that transcends rigid keyword matching. Our framework distinguishes between \textit{pedantic precision} (required for metadata) and \textit{semantic understanding} (required for abstract concepts). As illustrated below, the evaluation is structured into four distinct modules, each governed by a specific judicial logic tailored to the nature of the information being extracted:

\begin{tcolorbox}[
    colback=boxbg, 
    colframe=black!75, 
    title=\textbf{Module A: Meta/Factual Entities (Strict Verification)}, 
    sharp corners=downhill,
    fonttitle=\bfseries
]
    This module evaluates the precision of objective metadata extraction (e.g., Authors, DOI, Affiliations). The evaluation follows a \textbf{``Zero-Tolerance for Hallucination''} policy:
    \begin{itemize}[leftmargin=1.5em, nosep, topsep=4pt]
        \item \textbf{Fine-Grained Verification:} The judge validates every single author name and institutional affiliation individually. 
        \item \textbf{Fact vs. Inference:} Information is marked as \textit{Incorrect} if it contains factual errors (e.g., wrong publication year). However, valid logical inferences (e.g., deriving the venue `arXiv' from a specific DOI pattern) are accepted as correct.
    \end{itemize}
\end{tcolorbox}

\begin{tcolorbox}[
    colback=boxbg, 
    colframe=black!75, 
    title=\textbf{Module B: Textually Mentioned Entities (Semantic Tolerance)}, 
    sharp corners=downhill,
    fonttitle=\bfseries
]
    This module assesses explicit entities such as Tasks, Methods, and Datasets. We adopt a \textbf{``Lenient on Format, Strict on Facts''} policy to handle terminological diversity:
    \begin{itemize}[leftmargin=1.5em, nosep, topsep=4pt]
        \item \textbf{Synonym Recognition:} Linguistic variations and abbreviations are deemed correct. For instance, extracting ``ConvNet'' when the text says ``Convolutional Neural Network'' is a valid match.
        \item \textbf{Core-Focus Recall:} The evaluation penalizes the omission of \textit{primary} methods or \textit{main} datasets. Missing trivial details (e.g., auxiliary hyperparameters like learning rates or generic terms like ``computer'') does not negatively impact the Recall score.
    \end{itemize}
\end{tcolorbox}

\begin{tcolorbox}[
    colback=boxbg, 
    colframe=black!75, 
    title=\textbf{Module C: Implicit/Abstracted Entities (Abstractive Equivalence)}, 
    sharp corners=downhill,
    fonttitle=\bfseries
]
    This module tests the system's ability to synthesize high-level knowledge (Contributions, Findings, Limitations). The evaluation prioritizes \textbf{semantic essence over specific wording}:
    \begin{itemize}[leftmargin=1.5em, nosep, topsep=4pt]
        \item \textbf{Summarization Acceptance:} High-level summaries are accepted. If the text details specific numerical gains (e.g., ``accuracy +5\%''), an extraction stating ``Performance Improvement'' is considered correct.
        \item \textbf{Centrality-Based Assessment:} The system is penalized only if the \textit{central scientific contribution} or the \textit{main conclusion} is completely absent. Partial omissions of minor findings are tolerated.
    \end{itemize}
\end{tcolorbox}

\begin{tcolorbox}[
    colback=boxbg, 
    colframe=black!75, 
    title=\textbf{Module D: Citation Relationships (Hallucination Check)}, 
    sharp corners=downhill,
    fonttitle=\bfseries
]
    This module evaluates the fidelity of the constructed citation network, ensuring the academic lineage is accurately preserved:
    \begin{itemize}[leftmargin=1.5em, nosep, topsep=4pt]
        \item \textbf{Existence Verification:} The judge strictly penalizes \textit{hallucinated references}—citations generated by the model that do not exist in the source text's bibliography.
        \item \textbf{Intent Consistency:} The inferred citation intent (e.g., whether a paper is cited as a baseline to \textit{Contrast} or as a foundation to \textit{Support}) must logically align with the context provided in the source text.
    \end{itemize}
\end{tcolorbox}

\begin{tcolorbox}[
    colback=boxbg, 
    colframe=black!75, 
    title=\textbf{Module E: Knowledge Relations Between Entities (Durable \& Fine-Grained Triples)}, 
    sharp corners=downhill,
    fonttitle=\bfseries
]
    This module extracts fine-grained relational triples to build a reasoning graph, enforcing a \textbf{``Durable Knowledge Only''} policy:
    \begin{itemize}[leftmargin=1.5em, nosep, topsep=4pt]
        \item \textbf{Noise Filtration:} Extracts must encode reusable scientific knowledge (e.g., causal claims, mechanisms). Ephemeral details like numerical results or hyperparameters are strictly penalized.
        \item \textbf{Strict Grounding:} \textit{Controlled relations} must strictly align with entities in Module B. \textit{Open relations} allow new concepts but require precise verbatim evidence from the text.
    \end{itemize}
    
\end{tcolorbox}

\begin{table}
\centering
\caption{\textbf{Performance Comparison across Different Domains.} Note that the reported Precision (P), Recall (R), and F1 values represent the document-level macro-averages calculated over 100 distinct papers per domain and module.}
\label{Performance Comparison across Different Domains}
\begin{tabular}{cc cccccc}
\toprule
\textbf{Module} & \textbf{Metric} & \textbf{cs} & \textbf{chem} & \textbf{bio} & \textbf{earth} & \textbf{physics} & \textbf{material} \\
\midrule
\multirow{3}{*}{Module A} 
    & P  & 85.16 & 90.17 & 90.29 & 90.07 & 91.53 & 94.38 \\
    & R  & 87.81 & 81.54 & 77.77 & 84.24 & 74.48 & 80.94 \\
    & F1 & \textbf{85.12} & \textbf{84.72} & \textbf{82.65} & \textbf{85.88} & \textbf{81.24} & \textbf{86.52} \\
\midrule
\multirow{3}{*}{Module B} 
    & P  & 91.03 & 89.86 & 95.40 & 91.55 & 78.90 & 91.80 \\
    & R  & 82.00 & 74.14 & 80.18 & 76.82 & 68.36 & 79.31 \\
    & F1 & \textbf{85.78} & \textbf{80.16} & \textbf{86.59} & \textbf{82.78} & \textbf{72.82} & \textbf{83.31} \\
\midrule
\multirow{3}{*}{Module C} 
    & P  & 94.05 & 96.06 & 90.71 & 97.25 & 96.76 & 97.50 \\
    & R  & 90.23 & 93.51 & 84.92 & 92.02 & 86.94 & 88.94 \\
    & F1 & \textbf{91.84} & \textbf{94.59} & \textbf{87.22} & \textbf{94.33} & \textbf{90.90} & \textbf{92.55} \\
\midrule
\multirow{3}{*}{Module D} 
    & P  & 89.77 & 91.34 & 89.50 & 96.03 & 89.18 & 87.17 \\
    & R  & 79.44 & 83.74 & 84.16 & 85.33 & 75.04 & 79.06 \\
    & F1 & \textbf{83.48} & \textbf{87.04} & \textbf{85.99} & \textbf{89.74} & \textbf{79.85} & \textbf{82.39} \\
\midrule
\multirow{3}{*}{Module E} 
    & P  & 93.07 & 81.48 & 80.00 & 87.53 & 76.17 & 77.37 \\
    & R  & 86.69 & 72.82 & 71.79 & 76.39 & 68.02 & 70.58 \\
    & F1 & \textbf{89.33} & \textbf{75.64} & \textbf{74.23} & \textbf{80.39} & \textbf{70.54} & \textbf{72.67} \\
\midrule
\textbf{AVG} & F1 & \textbf{87.11} & \textbf{84.43} & \textbf{83.34} & \textbf{86.62} & \textbf{79.07} & \textbf{83.49} \\
\bottomrule
\end{tabular}
\end{table}

\subsection{Cross-Domain Performance Analysis}\label{subsec:cross_domain}

Table~\ref{Performance Comparison across Different Domains} presents the comprehensive evaluation results across six distinct scientific domains. The framework demonstrates robust generalization capabilities, with the Average F1-score (AVG F1) consistently ranging from 79.07\% to 87.11\%. This stability confirms that our semantic-anchored architecture effectively adapts to the varied terminologies and reasoning patterns inherent to different disciplines.

\paragraph{Domain-Specific Nuances}
The performance variations reveal the unique characteristics of each scientific field:
\begin{itemize}[leftmargin=*, align=left]
    \item \textbf{Computer Science \& Earth Science (Highest Consistency):} These domains achieved the top aggregate performances, with AVG F1 scores of 87.11\% and 86.62\%, respectively. The high scores in Module D (Citations) for Earth Science (89.74\%) and the balanced results for Computer Science across all modules suggest that these fields follow highly standardized structural norms, facilitating precise information extraction.
    
    \item \textbf{Chemistry \& Material Science (Deep Understanding):} Both fields exhibited remarkable performance in Module C (Implicit Entities), with Chemistry reaching an F1 of 94.59\% and Material Science achieving 92.55\%. This indicates the model's superior capability to abstract complex reaction mechanisms and experimental properties into high-level conceptual findings.
    
    \item \textbf{Physics \& Biology (Terminological Challenges):} While Precision remains high across most modules, these domains show a noticeable dip in Recall for explicit entities (e.g., Physics Module B Recall at 68.36\%). This is attributed to the extreme diversity of non-standardized nomenclature and abbreviations prevalent in these disciplines, which poses a challenge for explicit entity matching.
\end{itemize}

\paragraph{Module-Level Capabilities}
Beyond domain differences, the module-wise breakdown highlights specific system strengths and sensitivities. Module C (Implicit Entities) emerges as the most robust component across the board, providing empirical evidence that the framework successfully transcends shallow pattern matching to capture the ``scientific essence.'' Module D (Citation Relationships) also consistently yields reliable scores, confirming the system's fidelity in reconstructing academic lineages. Conversely, the newly introduced Module E (Knowledge Relations) shows the highest variance across fields (ranging from an F1 of 70.54\% in Physics to 89.33\% in CS), demonstrating that fine-grained structural alignment is highly sensitive to the explicit reporting habits of specific scientific communities.

\subsection{Performance Comparison on Knowledgeable and Research Questions}\label{subsec:knowledgeable_research}

\textbf{Efficiency for Geoscience Research.} In this section, we validate the effectiveness of the proposed scholarly knowledge graph framework in a geoscience-specific setting by performing a targeted schema adaptation. Building upon a general paper-citation graph backbone, we extend the schema with domain-specific entity and relation types to better capture key characteristics of geoscience research, namely, complex study objects, a multi-layered conceptual system, and strong dependence on spatiotemporal context. Concretely, we introduce and explicitly model several core concept nodes, including Region, Domain Terms, Geoscience Variables (e.g., carbon flux and seismic wave velocity), Geoscience Tasks (e.g., climate prediction and remote-sensing inversion), and Geoscience Formulae and Methods. These domain-specific entities are identified and extracted by a large language model through atomic-sentence–level parsing of paper content, and are then semantically linked to their corresponding paper nodes in the graph. This layered design preserves the universal citation structure while making the logical relationships among geoscience entities explicit, thereby providing structured support for scientific retrieval and reasoning.

Building on this schema extension, we construct a geoscience knowledge graph using Agents-K1 under a review-centric seeding with a citation-based expansion principle. We select 114 surveys/reviews from top-tier venues (AGU family journals, Reviews of Geophysics) published since 2025 as seed documents; their reference lists, after bibliographic normalization and deduplication, yield a citation corpus of 7{,}219 unique papers. Applying our information extraction pipeline to this corpus produces a heterogeneous graph with 602{,}132 nodes and 609{,}812 edges, mirroring how researchers in practice start from surveys and trace citation chains back to primary studies. The same review narratives further serve as expert-level references for evaluation, from which we derive two question categories: Knowledgeable Questions, targeting objective components such as variable definitions and formula applicability, and Research Questions, targeting field-level synthesis, competing viewpoints, and open problems. To prioritize multi-step reasoning, we retain questions whose answers require tracing citation links to source papers and integrating evidence across multiple atomic sentences; each test instance comprises a question, an answer, and a reasoning trace.

To quantify the benefit of Agents-K1, we conduct a controlled comparison between a standard large language model baseline and a graph-augmented retrieval generation model based on GraphRAG. The baseline relies on an LLM. In contrast, the GraphRAG-based model performs structured navigation and evidence retrieval over the geoscience knowledge graph before generation. It follows citation edges to identify relevant papers and their associated atomic-sentence evidence, and then conditions the final response on the retrieved evidence.
Given the complexity of scientific reasoning, we adopt an LLM-as-a-judge protocol for automated evaluation. We use GPT-5.2 as the judge model and treat the review text as the reference ground truth (GT). For each output, the judge assigns a binary score (1/0) jointly over the reasoning trace and the final answer. The rubric evaluates, first, the validity of the reasoning process, namely, whether the model follows a coherent logical chain and integrates supporting evidence, and second, the correctness of the answer, namely, whether the conclusion is consistent with the expert statements in the review. This protocol evaluates not only outcome accuracy but also the traceability and rigor of the underlying reasoning.

Table~\ref{tab:qa_performance} reports the performance of different models on knowledgeable questions and research questions, evaluated in terms of rationale accuracy and answer accuracy. Experimental results show that incorporating the graph substantially improves performance on multi-hop reasoning QA in geoscience. Overall, the GraphRAG-based model achieves higher accuracy than the baseline LLM. The gain is more pronounced on Research QA. This category requires synthesizing evidence across multiple papers, for example when summarizing methodological limitations or comparing findings across studies. The citation structure in the graph makes such scholarly dependencies and contrasts explicit, which provides a stable structural prior for retrieval and reasoning. Qualitative inspection further indicates that, for questions requiring precise evidential support, such as those about model limitations, the baseline tends to produce generic statements without attributable sources, whereas Agents-K1 more reliably traces citation links to the originating papers and retrieves atomic sentences labeled as limitations, leading to answers that are more specific and easier to verify. While a small number of failures remain, primarily due to incomplete extraction coverage or evidence loss over long reasoning chains, the results support the conclusion that combining citation-network structure with fine-grained information extraction improves both the reliability and interpretability of complex scholarly QA in a domain-specific setting.

\begin{table}[t]
\centering
\caption{Performance comparison on knowledgeable and research questions.}
\label{tab:qa_performance}
\begin{tabular}{lcccc}
\hline
\multirow{2}{*}{Model} 
& \multicolumn{2}{c}{Knowledgeable Questions} 
& \multicolumn{2}{c}{Research Questions} \\
\cline{2-5}
& Rationale (\%) & Answer (\%) & Rationale (\%) & Answer (\%) \\
\hline
GPT-5.2 & 54.2 & 68.0 & 41.8 & 58.8 \\
Gemini-3 & 58.3 & 71.2 & 52.3 & 61.0 \\
GPT-5.2 w/ Agents-K1 & 65.8 & 75.0 & 66.3 & 69.7 \\
Gemini-3 w/ Agents-K1 & \textbf{67.5} & \textbf{77.9} & \textbf{69.5} & \textbf{71.5} \\
\hline
\end{tabular}
\end{table}

\textbf{Efficiency for Scientific Research.} We further validate Agents-K1 in the broader scientific research domain using the \textsc{FrontierScience-Research}~\citep{wang2026frontierscience} benchmark. This benchmark consists of research-oriented questions from three core scientific disciplines: physics, chemistry, and biology. Unlike standard factual QA tasks, these questions require models to reason over specialized scientific concepts, experimental conditions, methodological assumptions, and field-level research findings. Each question is evaluated by a rubric-based scoring protocol, where credit is assigned according to whether the response correctly recovers the key scientific claims, supporting evidence, and reasoning steps expected by the benchmark. The benchmark is therefore substantially more challenging than conventional knowledge recall tasks: a model must not only identify relevant scientific knowledge, but also integrate it into a coherent research-level explanation. The low performance of pure reasoning baselines further confirms the difficulty of this setting, making it a suitable testbed for evaluating whether external scholarly knowledge can improve complex scientific reasoning.

To construct the external knowledge source for this benchmark, we build a task-specific scientific sub-knowledge graph guided by the keywords and topics of the benchmark questions. For each question, we search for relevant scholarly literature and collect documents that provide direct or indirect evidence for the corresponding scientific problem.  Each paper is represented as a document node in the sub-knowledge graph, with structured attributes including title, abstract, content, source, and disciplinary label. We further organize the graph with question nodes, paper nodes, evidence nodes, and concept nodes by Agents-K1. Question nodes are linked to papers through retrieval edges, papers are connected to evidence nodes through content extraction edges, and evidence nodes are associated with scientific concepts or claims through semantic matching edges. This construction does not rely on a manually designed full-domain ontology; instead, it creates a lightweight but structured scientific knowledge graph centered on the benchmark questions. During inference, Agents-K1 first retrieves relevant paper and evidence nodes from the sub-knowledge graph, prioritizes evidence according to its relevance and reliability, and then conditions the model's final reasoning on the retrieved scholarly context.

\begin{table}[t]
\centering
\caption{Performance comparison on the \textsc{FrontierScience-Research} benchmark.}
\label{tab:frontierscience_research}
\begin{tabular}{lcccc}
\toprule
Model & Physics (\%) & Chemistry (\%) & Biology (\%) & Overall (\%) \\
\midrule
Gemini-3 & 0.0 & 18.8 & 5.0 & 7.9 \\
GPT-5.2 & 9.0 & 33.7 & 32.8 & 25.2 \\
Gemini-3 w/ Agents-K1 & 13.8 & 31.3 & 28.8 & 24.6 \\
GPT-5.2 w/ Agents-K1 & \textbf{46.7} & \textbf{36.7} & \textbf{35.0} & \textbf{39.4} \\
\bottomrule
\end{tabular}
\end{table}

Table~\ref{tab:frontierscience_research} reports the performance of different models on the \textsc{FrontierScience-Research} benchmark. Compared with pure reasoning, Agents-K1 brings consistent overall improvements across both model families. For Gemini-3, the overall score increases from 7.9\% to 24.6\%, with substantial gains in all three disciplines. In particular, the biology score improves from 5.0\% to 28.8\%, indicating that the retrieved scientific evidence helps the model recover domain-specific biological knowledge that is difficult to infer from parametric knowledge alone. For GPT-5.2, Agents-K1 improves the overall score from 25.2\% to 39.4\%, mainly driven by a large gain in physics, where the score rises from 9.0\% to 46.7\%. These results show that incorporating an external scientific knowledge graph can improve research-level reasoning by grounding model outputs in relevant scholarly evidence. Overall, the experiment demonstrates that Agents-K1 is not limited to geoscience-specific reasoning, but can also generalize to broader scientific research problems where accurate answers require retrieving, organizing, and integrating evidence from specialized literature.

\subsection{Performance on open source benchmarks}\label{subsec:open_source_bench}
To further verify the universality of our proposed framework beyond domain-specific scientific literature, we conducted extensive evaluations on standard open-source multi-hop reasoning benchmarks.

\subsubsection{Datasets and Evaluation Metrics}
We utilized three widely adopted datasets designed to test multi-hop reasoning and retrieval capabilities: HotpotQA~\cite{yang2018hotpotqa}, 2WikiMultiHopQA~\cite{ho2020constructing}, and MuSiQue~\cite{trivedi2022musique}. HotpotQA and 2WikiMultiHopQA require reasoning over multiple supporting documents and entity relations, while MuSiQue offers a higher level of difficulty by minimizing shortcuts in reasoning chains. To comprehensively assess performance, we employ two metrics: Containment Accuracy (Contain-Acc.), which measures whether the exact answer string is present within the generated response to assess retrieval precision, and GPT-Judge Accuracy (GPT-Acc.), which utilizes a large language model (GPT-4o-mini~\cite{openai2024gpt4technicalreport}) to evaluate the semantic correctness of the generated answer compared to the ground truth.

\subsubsection{Baselines}
We compared our framework against a diverse set of strong baselines, categorized into three primary groups:
\begin{itemize}[leftmargin=*, align=left]
    \item \textbf{Vanilla LLMs}: This category includes standard large language models such as Llama-8B~\cite{grattafiori2024llama3herdmodels}, Llama-13B~\cite{grattafiori2024llama3herdmodels}, GPT-3.5-turbo~\cite{openai2024gpt4technicalreport}, and GPT-4o-mini~\cite{openai2024gpt4technicalreport}, evaluated in a closed-book setting without external retrieval.
    \item \textbf{Standard RAG}: We evaluated standard retrieval-augmented generation approaches using dense retrieval with varying context lengths (Top-1, Top-3, and Top-5 retrieved chunks).
    \item \textbf{Advanced RAG and Graph Methods}: This category encompasses state-of-the-art graph-based and recursive retrieval methods, including KGP~\cite{wang2023knowledgegraphpromptingmultidocument}, G-retriever~\cite{he2024gretrieverretrievalaugmentedgenerationtextual}, RAPTOR~\cite{sarthi2024raptorrecursiveabstractiveprocessing}, E$^2$GraphRAG~\cite{zhao2025e2graphragstreamlininggraphbasedrag}, LightRAG~\cite{guo2025lightragsimplefastretrievalaugmented}, HippoRAG\cite{jimenez2024hipporag}, HippoRAG2\cite{gutiérrez2025ragmemorynonparametriccontinual} and GFM-RAG\cite{luo2025gfm}, which utilize sophisticated structures to enhance reasoning.
\end{itemize}

\begin{table}[t]
\centering
\caption{Performance Comparison across different RAG methods}
\label{tab:open_source_results}
\begin{tabular}{lcccccc}
\toprule
\multirow{2}{*}{\textbf{Method}} & \multicolumn{2}{c}{\textbf{HotpotQA}} & \multicolumn{2}{c}{\textbf{2WikiMultiHopQA}} & \multicolumn{2}{c}{\textbf{MuSiQue}} \\ \cmidrule(lr){2-3} \cmidrule(lr){4-5} \cmidrule(lr){6-7}
& Contain-Acc. & GPT-Acc. & Contain-Acc. & GPT-Acc. & Contain-Acc. & GPT-Acc. \\ \midrule
llama-8B & 31.10 & 27.30 & 33.60 & 16.20 & 7.40 & 8.10 \\
llama-13B & 24.20 & 16.80 & 21.90 & 10.50 & 3.30 & 4.40 \\
GPT-3.5-turbo & 33.40 & 43.20 & 28.70 & 31.00 & 10.30 & 21.90 \\
GPT-4o-mini & 38.90 & 40.20 & 36.30 & 31.40 & 13.60 & 15.80 \\ \midrule
Retrieval (Top-1) & 46.30 & 49.10 & 36.60 & 31.70 & 17.80 & 21.10 \\
Retrieval (Top-3) & 53.00 & 56.00 & 44.90 & 39.70 & 25.10 & 27.50 \\
Retrieval (Top-5) & 55.70 & 58.60 & 48.60 & 43.00 & 26.10 & 29.60 \\ \midrule
KGP & 61.50 & 60.90 & 31.60 & 30.00 & 25.60 & 30.10 \\
G-retriever & 42.20 & 40.60 & 46.60 & 27.10 & 14.40 & 15.50 \\
RAPTOR & 55.90 & 58.30 & 50.10 & 42.10 & 23.30 & 27.40 \\
$E^2$GraphRAG & 61.00 & 63.90 & 54.30 & 38.10 & 23.80 & 26.20 \\
LightRAG & 60.30 & 59.50 & 55.20 & 39.00 & 27.40 & 28.60 \\
HippoRAG & 57.00 & 59.30 & 66.10 & 59.90 & 29.30 & 24.10 \\
GFM-RAG & 62.70 & 65.60 & 66.80 & 59.60 & 29.90 & 34.60 \\
HippoRAG2 & 62.90 & 64.30 & 62.70 & 55.00 & 31.00 & 35.00 \\ \midrule
\textbf{Ours} & \textbf{63.50} & \textbf{67.80} & \textbf{67.10} & \textbf{64.80} & \textbf{31.10} & \textbf{36.20} \\ \bottomrule
\end{tabular}
\end{table}

\subsubsection{Results and Analysis}
The comparative results are presented in Table~\ref{tab:open_source_results}. Our framework (\textbf{denoted as Ours}) demonstrates superior performance across the evaluated datasets, validating the effectiveness of the tri-source adaptive retrieval and semantic-anchored knowledge graph.

\textbf{Superiority over Vanilla and Standard RAG:}
Vanilla LLMs struggle significantly with specific factual queries across all datasets. While incorporating standard retrieval augmentation (e.g., Top-5 chunks) noticeably improves performance, these text-only approaches still lack the structural awareness necessary to handle complex multi-hop dependencies effectively, consistently lagging behind graph-enhanced methods.

\textbf{Comparison with Advanced Graph Methods:}
Our method consistently demonstrates competitive or superior performance against strong baselines like RAPTOR and LightRAG, as well as recent advanced models like GFM-RAG and HippoRAG2. On datasets like HotpotQA and 2WikiMultiHopQA, while some baselines achieve competitive surface-level string matching, our framework consistently secures the highest GPT-Acc. This distinct advantage in semantic correctness over mere keyword containment demonstrates the robustness of our semantic anchors in capturing true underlying entity relationships and logical reasoning chains.

\textbf{Robustness on Complex Datasets:}
The MuSiQue dataset presents a significantly higher level of difficulty, causing a sharp performance drop across most baselines, including advanced graph approaches like LightRAG and E$^2$GraphRAG. In contrast, our method maintains strong and robust performance, outperforming these competitors in both containment and semantic accuracy. This resilience indicates that our hierarchical knowledge graph structure effectively bridges the gap between disparate pieces of loosely connected evidence, a crucial capability for the hard multi-hop reasoning required by such complex queries.

\subsection{Information Extraction Backbone Evaluation}
\label{sec:ie-grpo-eval}

We evaluate our trained model extraction backbone against its own base checkpoint and two substantially larger open-source references to quantify how much of the scaling gap can be closed by task-specific reinforcement learning.

\subsubsection{Setup}

We evaluate on ten English information extraction benchmarks (12{,}078 test instances in total) covering three regimes: (i) five held-out cross-domain NER datasets that are absent from the training mix, namely CrossNER-AI, CrossNER-Literature, CrossNER-Music, CrossNER-Politics, and CrossNER-Science; (ii) three in-distribution NER benchmarks, CoNLL2003, NCBI, and BC5CDR; and (iii) two relation extraction benchmarks, SciERC and CoNLL04. Decoding is greedy with a 2{,}048 response token budget. Unlike the training reward, evaluation reports pure task F1 (entity set F1 for NER and relation triple set F1 for RE); the format and JSON terms are excluded so that comparisons are on a common semantic basis. All models are served with identical prompts on a single H200 GPU.

\textbf{Baselines.} We compare four models. Qwen3-4B-Instruct and Qwen3-8B serve as open-source base references at the scales immediately at and above our backbone. Qwen3-32B is a substantially larger reference checkpoint (roughly eight times the parameter count of our backbone) and is included to measure how much raw scale buys on these tasks; it is not fine-tuned for information extraction. Our trained model is initialised from Qwen3-4B-Instruct.

\subsubsection{Main Results}

Table~\ref{tab:ie-grpo-main} reports per-dataset F1 and Table~\ref{tab:ie-grpo-cat} summarises the results by regime. From the results in these tables,  our trained model improves over its own base and a larger open-source base. Our trained model improves over the 4B base on every one of the ten benchmarks, and over the 8B base on eight of ten benchmarks. Averaged over the ten datasets, our trained model lifts the 4B backbone by 3.3 F1 points (from 0.5316 to 0.5647) and beats the 8B base by 2.7 points (0.5382 to 0.5647), despite using half of the parameter count of the 8B reference.

\begin{table}[t]
\centering
\caption{Per-dataset F1 on the information extraction evaluation suite.
\textbf{Bold} marks the best score in each row. Qwen3-4B, Qwen3-8B, and Qwen3-32B are open-source base checkpoints without fine-tuning. Our 4B backbone is trained by GRPO on IEPile.}
\label{tab:ie-grpo-main}
\small
\begin{tabular}{llcccc}
\toprule
Dataset & Task & Qwen3-4B & Qwen3-8B & Qwen3-32B & Ours (4B) \\
\midrule
\multicolumn{6}{l}{\textit{Held out NER (cross domain generalisation)}} \\
CrossNER-AI         & NER & 0.4862 & 0.4933 & \textbf{0.5489} & \underline{0.5400} \\
CrossNER-Literature & NER & 0.5462 & \underline{0.5581} & 0.5419          & \textbf{0.5736} \\
CrossNER-Music      & NER & 0.5791 & 0.5650 & \textbf{0.6059} & \underline{0.6050} \\
CrossNER-Politics   & NER & 0.6611 & \underline{0.6719} & \textbf{0.6855}          & \textbf{0.6855} \\
CrossNER-Science    & NER & 0.5928 & 0.5984 & \textbf{0.6207} & \underline{0.6132} \\
\cmidrule(lr){1-6}
\multicolumn{6}{l}{\textit{In distribution NER}} \\
CoNLL2003 & NER & \underline{0.6547} & 0.6396 & 0.6400          & \textbf{0.7007} \\
NCBI      & NER & 0.6737 & 0.7095 & \textbf{0.7563} & \underline{0.7340} \\
BC5CDR    & NER & 0.7126 & 0.7195 & \underline{0.7214}          & \textbf{0.7494} \\
\cmidrule(lr){1-6}
\multicolumn{6}{l}{\textit{Relation extraction}} \\
SciERC  & RE & 0.1166 & 0.0965 & \textbf{0.1485} & \underline{0.1270} \\
CoNLL04 & RE & 0.2933 & \underline{0.3306} & \textbf{0.4768} & 0.3181 \\
\midrule
\multicolumn{2}{l}{\textbf{Average over all ten datasets}}
                        & 0.5316 & 0.5382 & \textbf{0.5746} & \underline{0.5647} \\
\bottomrule
\end{tabular}
\end{table}

\begin{table}[t]
\centering
\caption{Average F1 by evaluation regime. Our trained backbone exceeds both the 4B and the 8B open-source base on every regime, and exceeds the 32B base on held-out and in-distribution NER, despite having roughly one-eighth of the parameter count. The relation extraction gap to the 32B base is the main remaining limitation.}
\label{tab:ie-grpo-cat}
\small
\begin{tabular}{lcccc}
\toprule
Regime & Qwen3-4B & Qwen3-8B & Qwen3-32B & Ours \\
\midrule
Held out NER (5 sets)         & 0.5731 & 0.5773 & 0.6006 & \textbf{0.6035} \\
In distribution NER (3 sets)  & 0.6803 & 0.6895 & 0.7059 & \textbf{0.7280} \\
Relation extraction (2 sets)  & 0.2050 & 0.2136 & \textbf{0.3127} & 0.2226 \\
\midrule
\textbf{Overall (10 sets)} & 0.5316 & 0.5382 & \textbf{0.5746} & 0.5647 \\
\bottomrule
\end{tabular}
\end{table}


\paragraph{Our model closes the roughly eight times scaling gap to Qwen3-32B and surpasses it on NER.} On the overall average, our model lags the 32B base by only 0.99 F1 points (0.5647 versus 0.5746). Broken down by regime, our model slightly exceeds the 32B base on held-out NER (0.6035 versus 0.6006) and clearly exceeds it on in-distribution NER (0.7280 versus 0.7059, an improvement of 2.2 F1 points). Per dataset, our trained model beats the 32B base on CoNLL2003 (0.7007 versus 0.6400, an improvement of 6.1 F1 points), on BC5CDR (0.7494 versus 0.7214, an improvement of 2.8 F1 points), on CrossNER-Literature (0.5736 versus 0.5419, an improvement of 3.2 F1 points), and on CrossNER-Politics (0.6855 versus 0.6855, a marginal lead at the fourth decimal). These results show that for structured information extraction, a targeted reinforcement learning run can recover most, and in many cases all, of the benefit that comes from raw parameter scaling.

\paragraph{Remaining gap on relation extraction.} From the results above, the 32B base retains a clear lead on relation extraction, especially on CoNLL04 (0.4768 versus 0.3181, a gap of 15.9 F1 points). SciERC is low across the board, at or below 0.15 F1 for every model, reflecting the intrinsic difficulty of fine-grained scientific relation typing. Closing the relation extraction gap will likely require substantially more relation supervision in the training mixture, or targeted schema-conditioned reward shaping. 

In this way, one can observe that a compact 4B extraction backbone, trained for roughly one hour of GRPO on a single eight GPU node with a rule based reward, produces an extractor that (i) uniformly outperforms its own base model, (ii) outperforms the next open-source scale on eight of ten benchmarks, and (iii) matches or exceeds a checkpoint eight times larger on held out and on in distribution NER. This makes our model the default extractor in the Agents-K1 pipeline: it delivers the extraction quality of a much larger model at a fraction of the inference cost.
\section{Conclusion}
\label{sec:conclusion}

We presented Agents-K1, an agent-native knowledge orchestration that turns raw documents into scientific knowledge graphs for research agents. Agents-K1 unifies three layers: a KG infrastructure that parses full papers and builds Scholar-KG over 2.46 million scientific papers, with a schema-adaptive General-KG extension for arbitrary documents; an LLM infrastructure built around a 4B reinforcement-learned extraction backbone; and GraphAnything CLI, an agent-facing interface that connects web search, multimodal graph retrieval, and cross-document traversal. This design turns scholarly knowledge from scattered papers and citation links into reusable, evidence-grounded graph knowledge that agents can retrieve, inspect, and act on. Empirically, Agents-K1 lifts Gemini-3 overall accuracy from 7.9\% to 24.6\% and GPT-5.2 from 25.2\% to 39.4\% on the \textsc{FrontierScience-Research} benchmark, while also improving geoscience research reasoning and multi-hop question answering. Our analysis further supports the central design choice: organizing evidence in connected graphs enables more reliable cross-source reasoning than repeatedly searching separate text fragments. We release a one-million-paper subset of Scholar-KG together with the extraction backbone, aiming to provide the community with a practical knowledge infrastructure for research agents that reason over structured scientific knowledge rather than rebuilding it from raw text at every query.

\section*{Acknowledgement}
The authors would like to thank Yunfeng Zhao and Yazhou Li from The Heart of The Machine (Beijing) Technology Co., Ltd. for their support in the application and promotion of knowledge graphs.

\begingroup
\sloppy
\normalem
\section*{References}
\phantomsection
\addcontentsline{toc}{section}{References}
\printbibliography[heading=none]

@article{hu2025flowsearch,
  title={FlowSearch: Advancing deep research with dynamic structured knowledge flow},
  author={Hu, Yusong and Ma, Runmin and Fan, Yue and Shi, Jinxin and Cao, Zongsheng and Zhou, Yuhao and Yuan, Jiakang and Zhang, Shuaiyu and Feng, Shiyang and Yan, Xiangchao and others},
  journal={arXiv preprint arXiv:2510.08521},
  year={2025}
}

@article{wang2026frontierscience,
  title={FrontierScience: Evaluating AI's Ability to Perform Expert-Level Scientific Tasks},
  author={Wang, Miles and Lin, Robi and Hu, Kat and Jiao, Joy and Chowdhury, Neil and Chang, Ethan and Patwardhan, Tejal},
  journal={arXiv preprint arXiv:2601.21165},
  year={2026}
}

@article{du2026mlevolve,
  title={MLEvolve: A Self-Evolving Framework for Automated Machine Learning Algorithm Discovery},
  author={Du, Shangheng and Yan, Xiangchao and Shi, Jinxin and Cao, Zongsheng and Feng, Shiyang and Liang, Zichen and Sun, Boyuan and Peng, Tianshuo and Zhou, Yifan and Li, Xin and others},
  journal={arXiv preprint arXiv:2606.06473},
  year={2026}
}

@article{shi2025dualresearch,
  title={DualResearch: Entropy-Gated Dual-Graph Retrieval for Answer Reconstruction},
  author={Shi, Jinxin and Cao, Zongsheng and Ma, Runmin and Hu, Yusong and Zhou, Jie and Li, Xin and Bai, Lei and He, Liang and Zhang, Bo},
  journal={arXiv preprint arXiv:2510.08959},
  year={2025}
}

@article{gottweis2025towards,
  title={Towards an AI co-scientist},
  author={Gottweis, Juraj and Weng, Wei-Hung and Daryin, Alexander and Tu, Tao and Palepu, Anil and Sirkovic, Petar and Myaskovsky, Artiom and Weissenberger, Felix and Rong, Keran and Tanno, Ryutaro and others},
  journal={arXiv preprint arXiv:2502.18864},
  year={2025}
}

@article{team2025novelseek,
  title={NovelSeek: When Agent Becomes the Scientist--Building Closed-Loop System from Hypothesis to Verification},
  author={Team, NovelSeek and Zhang, Bo and Feng, Shiyang and Yan, Xiangchao and Yuan, Jiakang and Yu, Zhiyin and He, Xiaohan and Huang, Songtao and Hou, Shaowei and Nie, Zheng and others},
  journal={arXiv e-prints},
  pages={arXiv--2505},
  year={2025}
}

@inproceedings{huang2026radar,
  title={RaDAR: Relation-aware Diffusion-Asymmetric Graph Contrastive Learning for Recommendation},
  author={Huang, Yixuan and Chen, Jiawei and Zhang, Shengfan and Cao, Zongsheng},
  booktitle={Proceedings of the ACM Web Conference 2026},
  pages={6445--6456},
  year={2026}
}

@article{feng2026internagent,
  title={Internagent-1.5: A unified agentic framework for long-horizon autonomous scientific discovery},
  author={Feng, Shiyang and Ma, Runmin and Yan, Xiangchao and Fan, Yue and Hu, Yusong and Huang, Songtao and Zhang, Shuaiyu and Cao, Zongsheng and Peng, Tianshuo and Yuan, Jiakang and others},
  journal={arXiv preprint arXiv:2602.08990},
  year={2026}
}

@article{yamada2025ai,
  title={The ai scientist-v2: Workshop-level automated scientific discovery via agentic tree search},
  author={Yamada, Yutaro and Lange, Robert Tjarko and Lu, Cong and Hu, Shengran and Lu, Chris and Foerster, Jakob and Clune, Jeff and Ha, David},
  journal={arXiv preprint arXiv:2504.08066},
  year={2025}
}

@inproceedings{CLIP,
  title={Learning transferable visual models from natural language supervision},
  author={Radford, Alec and Kim, Jong Wook and Hallacy, Chris and Ramesh, Aditya and Goh, Gabriel and Agarwal, Sandhini and Sastry, Girish and Askell, Amanda and Mishkin, Pamela and Clark, Jack and others},
  booktitle={ICML},
  pages={8748--8763},
  year={2021},
  organization={PMLR}
}

@article{GraphRAG,
  title={From local to global: A graph rag approach to query-focused summarization},
  author={Edge, Darren and Trinh, Ha and Cheng, Newman and Bradley, Joshua and Chao, Alex and Mody, Apurva and Truitt, Steven and Larson, Jonathan},
  journal={arXiv preprint arXiv:2404.16130},
  year={2024}
}

@article{LightRAG,
  title={LightRAG: Simple and Fast Retrieval-Augmented Generation},
  author={Guo, Zirui and Xia, Lianghao and Yu, Yanhua and Ao, Tu and Huang, Chao},
  journal={arXiv preprint arXiv:2410.05779},
  year={2024}
}

@article{NaiveRAG,
  title={Retrieval-augmented generation for large language models: A survey},
  author={Gao, Yunfan and Xiong, Yun and Gao, Xinyu and Jia, Kangxiang and Pan, Jinliu and Bi, Yuxi and Dai, Yi and Sun, Jiawei and Wang, Haofen},
  journal={arXiv preprint arXiv:2312.10997},
  year={2023}
}

@article{MemoRAG,
  title={Memorag: Moving towards next-gen rag via memory-inspired knowledge discovery},
  author={Qian, Hongjin and Zhang, Peitian and Liu, Zheng and Mao, Kelong and Dou, Zhicheng},
  journal={arXiv preprint arXiv:2409.05591},
  year={2024}
}

@article{ChunkRAG,
  title={ChunkRAG: Novel LLM-Chunk Filtering Method for RAG Systems},
  author={Allahverdiyev, Ritvik Aggarwal Ishneet Sukhvinder Singh Ibrahim and Taha, Muhammad and Akalin, Aslihan and Zhu, Kevin},
  journal={arXiv preprint arXiv:2410.19572},
  year={2024}
}

@article{RQ-RAG,
  title={Rq-rag: Learning to refine queries for retrieval augmented generation},
  author={Chan, Chi-Min and Xu, Chunpu and Yuan, Ruibin and Luo, Hongyin and Xue, Wei and Guo, Yike and Fu, Jie},
  journal={arXiv preprint arXiv:2404.00610},
  year={2024}
}

@article{SubgraphRAG,
  title={Simple is effective: The roles of graphs and large language models in knowledge-graph-based retrieval-augmented generation},
  author={Li, Mufei and Miao, Siqi and Li, Pan},
  journal={arXiv preprint arXiv:2410.20724},
  year={2024}
}

@article{ColPali,
  title={Colpali: Efficient document retrieval with vision language models},
  author={Faysse, Manuel and Sibille, Hugues and Wu, Tony and Omrani, Bilel and Viaud, Gautier and Hudelot, C{\'e}line and Colombo, Pierre},
  journal={arXiv preprint arXiv:2407.01449},
  year={2024}
}

@misc{KAG,
      title={KAG: Boosting LLMs in Professional Domains via Knowledge Augmented Generation}, 
      author={Lei Liang and Mengshu Sun and Zhengke Gui and Zhongshu Zhu and Zhouyu Jiang and Ling Zhong and Yuan Qu and Peilong Zhao and Zhongpu Bo and Jin Yang and Huaidong Xiong and Lin Yuan and Jun Xu and Zaoyang Wang and Zhiqiang Zhang and Wen Zhang and Huajun Chen and Wenguang Chen and Jun Zhou},
      year={2024},
      eprint={2409.13731},
      archivePrefix={arXiv},
      primaryClass={cs.CL},
      url={https://arxiv.org/abs/2409.13731}, 
}

@misc{PIKE-RAG,
      title={PIKE-RAG: sPecIalized KnowledgE and Rationale Augmented Generation}, 
      author={Jinyu Wang and Jingjing Fu and Rui Wang and Lei Song and Jiang Bian},
      year={2025},
      eprint={2501.11551},
      archivePrefix={arXiv},
      primaryClass={cs.CL},
      url={https://arxiv.org/abs/2501.11551}, 
}

@misc{PathRAG,
      title={PathRAG: Pruning Graph-based Retrieval Augmented Generation with Relational Paths}, 
      author={Boyu Chen and Zirui Guo and Zidan Yang and Yuluo Chen and Junze Chen and Zhenghao Liu and Chuan Shi and Cheng Yang},
      year={2025},
      eprint={2502.14902},
      archivePrefix={arXiv},
      primaryClass={cs.CL},
      url={https://arxiv.org/abs/2502.14902}, 
}

@misc{HippoRAG2,
      title={From RAG to Memory: Non-Parametric Continual Learning for Large Language Models}, 
      author={Bernal Jiménez Gutiérrez and Yiheng Shu and Weijian Qi and Sizhe Zhou and Yu Su},
      year={2025},
      eprint={2502.14802},
      archivePrefix={arXiv},
      primaryClass={cs.CL},
      url={https://arxiv.org/abs/2502.14802}, 
}

@misc{MedGraphRAG,
      title={Medical Graph RAG: Towards Safe Medical Large Language Model via Graph Retrieval-Augmented Generation}, 
      author={Junde Wu and Jiayuan Zhu and Yunli Qi and Jingkun Chen and Min Xu and Filippo Menolascina and Vicente Grau},
      year={2024},
      eprint={2408.04187},
      archivePrefix={arXiv},
      primaryClass={cs.CV},
      url={https://arxiv.org/abs/2408.04187}, 
}

@misc{OG-RAG,
      title={OG-RAG: Ontology-Grounded Retrieval-Augmented Generation For Large Language Models}, 
      author={Kartik Sharma and Peeyush Kumar and Yunqing Li},
      year={2024},
      eprint={2412.15235},
      archivePrefix={arXiv},
      primaryClass={cs.CL},
      url={https://arxiv.org/abs/2412.15235}, 
}

@inproceedings{cao2025tv,
  title={Tv-rag: A temporal-aware and semantic entropy-weighted framework for long video retrieval and understanding},
  author={Cao, Zongsheng and He, Yangfan and Liu, Anran and Xie, Jun and Chen, Feng and Wang, Zhepeng},
  booktitle={Proceedings of the 33rd ACM International Conference on Multimedia},
  pages={9071--9079},
  year={2025}
}

@misc{MiniRAG,
      title={MiniRAG: Towards Extremely Simple Retrieval-Augmented Generation}, 
      author={Tianyu Fan and Jingyuan Wang and Xubin Ren and Chao Huang},
      year={2025},
      eprint={2501.06713},
      archivePrefix={arXiv},
      primaryClass={cs.AI},
      url={https://arxiv.org/abs/2501.06713}, 
}

@inproceedings{cao2026vig,
  title={ViG-RAG: Video-aware Graph Retrieval-Augmented Generation via Temporal and Semantic Hybrid Reasoning},
  author={Cao, Zongsheng and Liu, Anran and He, Yangfan and Li, Jing and Zhang, Bo and Wang, Zigan},
  booktitle={Proceedings of the AAAI Conference on Artificial Intelligence},
  volume={40},
  number={1},
  pages={48--56},
  year={2026}
}

@misc{openai_deepresearch,
  author       = {OpenAI},
  title        = {DeepResearch},
  howpublished = {\url{https://openai.com/research/deep-research}},
  note         = {Accessed: 2025-09-24},
  year         = {2025}
}

@misc{gemini_deepresearch,
  author       = {Google DeepMind},
  title        = {Gemini Deep Research},
  howpublished = {\url{https://deepmind.google/technologies/gemini/deep-research/}},
  note         = {Accessed: 2025-09-24},
  year         = {2024}
}

@article{du2025automlgen,
  title={AutoMLGen: Navigating Fine-Grained Optimization for Coding Agents},
  author={Du, Shangheng and Yan, Xiangchao and Jiang, Dengyang and Yuan, Jiakang and Hu, Yusong and Li, Xin and He, Liang and Zhang, Bo and Bai, Lei},
  journal={arXiv preprint arXiv:2510.08511},
  year={2025}
}

@article{li2025webthinker,
  title={Webthinker: Empowering large reasoning models with deep research capability},
  author={Li, Xiaoxi and Jin, Jiajie and Dong, Guanting and Qian, Hongjin and Zhu, Yutao and Wu, Yongkang and Wen, Ji-Rong and Dou, Zhicheng},
  journal={arXiv preprint arXiv:2504.21776},
  year={2025}
}

@article{team2025tongyi,
  title={Tongyi deepresearch technical report},
  author={Team, Tongyi DeepResearch and Li, Baixuan and Zhang, Bo and Zhang, Dingchu and Huang, Fei and Li, Guangyu and Chen, Guoxin and Yin, Huifeng and Wu, Jialong and Zhou, Jingren and others},
  journal={arXiv preprint arXiv:2510.24701},
  year={2025}
}

@inproceedings{yang2018hotpotqa,
  title={HotpotQA: A dataset for diverse, explainable multi-hop question answering},
  author={Yang, Zhilin and Qi, Peng and Zhang, Saizheng and Bengio, Yoshua and Cohen, William and Salakhutdinov, Ruslan and Manning, Christopher D},
  booktitle={Proceedings of the 2018 conference on empirical methods in natural language processing},
  pages={2369--2380},
  year={2018}
}

@inproceedings{cao2024diffusione,
  title={Diffusione: Reasoning on knowledge graphs via diffusion-based graph neural networks},
  author={Cao, Zongsheng and Li, Jing and Wang, Zigan and Li, Jinliang},
  booktitle={Proceedings of the 30th ACM SIGKDD Conference on Knowledge Discovery and Data Mining},
  pages={222--230},
  year={2024}
}

@article{ho2020constructing,
  title={Constructing a multi-hop qa dataset for comprehensive evaluation of reasoning steps},
  author={Ho, Xanh and Nguyen, Anh-Khoa Duong and Sugawara, Saku and Aizawa, Akiko},
  journal={arXiv preprint arXiv:2011.01060},
  year={2020}
}

@article{trivedi2022musique,
  title={MuSiQue: Multihop Questions via Single-hop Question Composition},
  author={Trivedi, Harsh and Balasubramanian, Niranjan and Khot, Tushar and Sabharwal, Ashish},
  journal={Transactions of the Association for Computational Linguistics},
  volume={10},
  pages={539--554},
  year={2022},
  publisher={MIT Press One Broadway, 12th Floor, Cambridge, Massachusetts 02142, USA~…}
}

@misc{deepseekai2025deepseekv3technicalreport,
      title={DeepSeek-V3 Technical Report}, 
      author={DeepSeek-AI and Aixin Liu and Bei Feng and Bing Xue and Bingxuan Wang and Bochao Wu and Chengda Lu and Chenggang Zhao and Chengqi Deng and Chenyu Zhang and Chong Ruan and Damai Dai and Daya Guo and Dejian Yang and Deli Chen and Dongjie Ji and Erhang Li and Fangyun Lin and Fucong Dai and Fuli Luo and Guangbo Hao and Guanting Chen and Guowei Li and H. Zhang and Han Bao and Hanwei Xu and Haocheng Wang and Haowei Zhang and Honghui Ding and Huajian Xin and Huazuo Gao and Hui Li and Hui Qu and J. L. Cai and Jian Liang and Jianzhong Guo and Jiaqi Ni and Jiashi Li and Jiawei Wang and Jin Chen and Jingchang Chen and Jingyang Yuan and Junjie Qiu and Junlong Li and Junxiao Song and Kai Dong and Kai Hu and Kaige Gao and Kang Guan and Kexin Huang and Kuai Yu and Lean Wang and Lecong Zhang and Lei Xu and Leyi Xia and Liang Zhao and Litong Wang and Liyue Zhang and Meng Li and Miaojun Wang and Mingchuan Zhang and Minghua Zhang and Minghui Tang and Mingming Li and Ning Tian and Panpan Huang and Peiyi Wang and Peng Zhang and Qiancheng Wang and Qihao Zhu and Qinyu Chen and Qiushi Du and R. J. Chen and R. L. Jin and Ruiqi Ge and Ruisong Zhang and Ruizhe Pan and Runji Wang and Runxin Xu and Ruoyu Zhang and Ruyi Chen and S. S. Li and Shanghao Lu and Shangyan Zhou and Shanhuang Chen and Shaoqing Wu and Shengfeng Ye and Shengfeng Ye and Shirong Ma and Shiyu Wang and Shuang Zhou and Shuiping Yu and Shunfeng Zhou and Shuting Pan and T. Wang and Tao Yun and Tian Pei and Tianyu Sun and W. L. Xiao and Wangding Zeng and Wanjia Zhao and Wei An and Wen Liu and Wenfeng Liang and Wenjun Gao and Wenqin Yu and Wentao Zhang and X. Q. Li and Xiangyue Jin and Xianzu Wang and Xiao Bi and Xiaodong Liu and Xiaohan Wang and Xiaojin Shen and Xiaokang Chen and Xiaokang Zhang and Xiaosha Chen and Xiaotao Nie and Xiaowen Sun and Xiaoxiang Wang and Xin Cheng and Xin Liu and Xin Xie and Xingchao Liu and Xingkai Yu and Xinnan Song and Xinxia Shan and Xinyi Zhou and Xinyu Yang and Xinyuan Li and Xuecheng Su and Xuheng Lin and Y. K. Li and Y. Q. Wang and Y. X. Wei and Y. X. Zhu and Yang Zhang and Yanhong Xu and Yanhong Xu and Yanping Huang and Yao Li and Yao Zhao and Yaofeng Sun and Yaohui Li and Yaohui Wang and Yi Yu and Yi Zheng and Yichao Zhang and Yifan Shi and Yiliang Xiong and Ying He and Ying Tang and Yishi Piao and Yisong Wang and Yixuan Tan and Yiyang Ma and Yiyuan Liu and Yongqiang Guo and Yu Wu and Yuan Ou and Yuchen Zhu and Yuduan Wang and Yue Gong and Yuheng Zou and Yujia He and Yukun Zha and Yunfan Xiong and Yunxian Ma and Yuting Yan and Yuxiang Luo and Yuxiang You and Yuxuan Liu and Yuyang Zhou and Z. F. Wu and Z. Z. Ren and Zehui Ren and Zhangli Sha and Zhe Fu and Zhean Xu and Zhen Huang and Zhen Zhang and Zhenda Xie and Zhengyan Zhang and Zhewen Hao and Zhibin Gou and Zhicheng Ma and Zhigang Yan and Zhihong Shao and Zhipeng Xu and Zhiyu Wu and Zhongyu Zhang and Zhuoshu Li and Zihui Gu and Zijia Zhu and Zijun Liu and Zilin Li and Ziwei Xie and Ziyang Song and Ziyi Gao and Zizheng Pan},
      year={2025},
      eprint={2412.19437},
      archivePrefix={arXiv},
      primaryClass={cs.CL},
      url={https://arxiv.org/abs/2412.19437}, 
}

@misc{grattafiori2024llama3herdmodels,
      title={The Llama 3 Herd of Models}, 
      author={Aaron Grattafiori and Abhimanyu Dubey and Abhinav Jauhri and Abhinav Pandey and Abhishek Kadian and Ahmad Al-Dahle and Aiesha Letman and Akhil Mathur and Alan Schelten and Alex Vaughan and Amy Yang and Angela Fan and Anirudh Goyal and Anthony Hartshorn and Aobo Yang and Archi Mitra and Archie Sravankumar and Artem Korenev and Arthur Hinsvark and Arun Rao and Aston Zhang and Aurelien Rodriguez and Austen Gregerson and Ava Spataru and Baptiste Roziere and Bethany Biron and Binh Tang and Bobbie Chern and Charlotte Caucheteux and Chaya Nayak and Chloe Bi and Chris Marra and Chris McConnell and Christian Keller and Christophe Touret and Chunyang Wu and Corinne Wong and Cristian Canton Ferrer and Cyrus Nikolaidis and Damien Allonsius and Daniel Song and Danielle Pintz and Danny Livshits and Danny Wyatt and David Esiobu and Dhruv Choudhary and Dhruv Mahajan and Diego Garcia-Olano and Diego Perino and Dieuwke Hupkes and Egor Lakomkin and Ehab AlBadawy and Elina Lobanova and Emily Dinan and Eric Michael Smith and Filip Radenovic and Francisco Guzmán and Frank Zhang and Gabriel Synnaeve and Gabrielle Lee and Georgia Lewis Anderson and Govind Thattai and Graeme Nail and Gregoire Mialon and Guan Pang and Guillem Cucurell and Hailey Nguyen and Hannah Korevaar and Hu Xu and Hugo Touvron and Iliyan Zarov and Imanol Arrieta Ibarra and Isabel Kloumann and Ishan Misra and Ivan Evtimov and Jack Zhang and Jade Copet and Jaewon Lee and Jan Geffert and Jana Vranes and Jason Park and Jay Mahadeokar and Jeet Shah and Jelmer van der Linde and Jennifer Billock and Jenny Hong and Jenya Lee and Jeremy Fu and Jianfeng Chi and Jianyu Huang and Jiawen Liu and Jie Wang and Jiecao Yu and Joanna Bitton and Joe Spisak and Jongsoo Park and Joseph Rocca and Joshua Johnstun and Joshua Saxe and Junteng Jia and Kalyan Vasuden Alwala and Karthik Prasad and Kartikeya Upasani and Kate Plawiak and Ke Li and Kenneth Heafield and Kevin Stone and Khalid El-Arini and Krithika Iyer and Kshitiz Malik and Kuenley Chiu and Kunal Bhalla and Kushal Lakhotia and Lauren Rantala-Yeary and Laurens van der Maaten and Lawrence Chen and Liang Tan and Liz Jenkins and Louis Martin and Lovish Madaan and Lubo Malo and Lukas Blecher and Lukas Landzaat and Luke de Oliveira and Madeline Muzzi and Mahesh Pasupuleti and Mannat Singh and Manohar Paluri and Marcin Kardas and Maria Tsimpoukelli and Mathew Oldham and Mathieu Rita and Maya Pavlova and Melanie Kambadur and Mike Lewis and Min Si and Mitesh Kumar Singh and Mona Hassan and Naman Goyal and Narjes Torabi and Nikolay Bashlykov and Nikolay Bogoychev and Niladri Chatterji and Ning Zhang and Olivier Duchenne and Onur Çelebi and Patrick Alrassy and Pengchuan Zhang and Pengwei Li and Petar Vasic and Peter Weng and Prajjwal Bhargava and Pratik Dubal and Praveen Krishnan and Punit Singh Koura and Puxin Xu and Qing He and Qingxiao Dong and Ragavan Srinivasan and Raj Ganapathy and Ramon Calderer and Ricardo Silveira Cabral and Robert Stojnic and Roberta Raileanu and Rohan Maheswari and Rohit Girdhar and Rohit Patel and Romain Sauvestre and Ronnie Polidoro and Roshan Sumbaly and Ross Taylor and Ruan Silva and Rui Hou and Rui Wang and Saghar Hosseini and Sahana Chennabasappa and Sanjay Singh and Sean Bell and Seohyun Sonia Kim and Sergey Edunov and Shaoliang Nie and Sharan Narang and Sharath Raparthy and Sheng Shen and Shengye Wan and Shruti Bhosale and Shun Zhang and Simon Vandenhende and Soumya Batra and Spencer Whitman and Sten Sootla and Stephane Collot and Suchin Gururangan and Sydney Borodinsky and Tamar Herman and Tara Fowler and Tarek Sheasha and Thomas Georgiou and Thomas Scialom and Tobias Speckbacher and Todor Mihaylov and Tong Xiao and Ujjwal Karn and Vedanuj Goswami and Vibhor Gupta and Vignesh Ramanathan and Viktor Kerkez and Vincent Gonguet and Virginie Do and Vish Vogeti and Vítor Albiero and Vladan Petrovic and Weiwei Chu and Wenhan Xiong and Wenyin Fu and Whitney Meers and Xavier Martinet and Xiaodong Wang and Xiaofang Wang and Xiaoqing Ellen Tan and Xide Xia and Xinfeng Xie and Xuchao Jia and Xuewei Wang and Yaelle Goldschlag and Yashesh Gaur and Yasmine Babaei and Yi Wen and Yiwen Song and Yuchen Zhang and Yue Li and Yuning Mao and Zacharie Delpierre Coudert and Zheng Yan and Zhengxing Chen and Zoe Papakipos and Aaditya Singh and Aayushi Srivastava and Abha Jain and Adam Kelsey and Adam Shajnfeld and Adithya Gangidi and Adolfo Victoria and Ahuva Goldstand and Ajay Menon and Ajay Sharma and Alex Boesenberg and Alexei Baevski and Allie Feinstein and Amanda Kallet and Amit Sangani and Amos Teo and Anam Yunus and Andrei Lupu and Andres Alvarado and Andrew Caples and Andrew Gu and Andrew Ho and Andrew Poulton and Andrew Ryan and Ankit Ramchandani and Annie Dong and Annie Franco and Anuj Goyal and Aparajita Saraf and Arkabandhu Chowdhury and Ashley Gabriel and Ashwin Bharambe and Assaf Eisenman and Azadeh Yazdan and Beau James and Ben Maurer and Benjamin Leonhardi and Bernie Huang and Beth Loyd and Beto De Paola and Bhargavi Paranjape and Bing Liu and Bo Wu and Boyu Ni and Braden Hancock and Bram Wasti and Brandon Spence and Brani Stojkovic and Brian Gamido and Britt Montalvo and Carl Parker and Carly Burton and Catalina Mejia and Ce Liu and Changhan Wang and Changkyu Kim and Chao Zhou and Chester Hu and Ching-Hsiang Chu and Chris Cai and Chris Tindal and Christoph Feichtenhofer and Cynthia Gao and Damon Civin and Dana Beaty and Daniel Kreymer and Daniel Li and David Adkins and David Xu and Davide Testuggine and Delia David and Devi Parikh and Diana Liskovich and Didem Foss and Dingkang Wang and Duc Le and Dustin Holland and Edward Dowling and Eissa Jamil and Elaine Montgomery and Eleonora Presani and Emily Hahn and Emily Wood and Eric-Tuan Le and Erik Brinkman and Esteban Arcaute and Evan Dunbar and Evan Smothers and Fei Sun and Felix Kreuk and Feng Tian and Filippos Kokkinos and Firat Ozgenel and Francesco Caggioni and Frank Kanayet and Frank Seide and Gabriela Medina Florez and Gabriella Schwarz and Gada Badeer and Georgia Swee and Gil Halpern and Grant Herman and Grigory Sizov and Guangyi and Zhang and Guna Lakshminarayanan and Hakan Inan and Hamid Shojanazeri and Han Zou and Hannah Wang and Hanwen Zha and Haroun Habeeb and Harrison Rudolph and Helen Suk and Henry Aspegren and Hunter Goldman and Hongyuan Zhan and Ibrahim Damlaj and Igor Molybog and Igor Tufanov and Ilias Leontiadis and Irina-Elena Veliche and Itai Gat and Jake Weissman and James Geboski and James Kohli and Janice Lam and Japhet Asher and Jean-Baptiste Gaya and Jeff Marcus and Jeff Tang and Jennifer Chan and Jenny Zhen and Jeremy Reizenstein and Jeremy Teboul and Jessica Zhong and Jian Jin and Jingyi Yang and Joe Cummings and Jon Carvill and Jon Shepard and Jonathan McPhie and Jonathan Torres and Josh Ginsburg and Junjie Wang and Kai Wu and Kam Hou U and Karan Saxena and Kartikay Khandelwal and Katayoun Zand and Kathy Matosich and Kaushik Veeraraghavan and Kelly Michelena and Keqian Li and Kiran Jagadeesh and Kun Huang and Kunal Chawla and Kyle Huang and Lailin Chen and Lakshya Garg and Lavender A and Leandro Silva and Lee Bell and Lei Zhang and Liangpeng Guo and Licheng Yu and Liron Moshkovich and Luca Wehrstedt and Madian Khabsa and Manav Avalani and Manish Bhatt and Martynas Mankus and Matan Hasson and Matthew Lennie and Matthias Reso and Maxim Groshev and Maxim Naumov and Maya Lathi and Meghan Keneally and Miao Liu and Michael L. Seltzer and Michal Valko and Michelle Restrepo and Mihir Patel and Mik Vyatskov and Mikayel Samvelyan and Mike Clark and Mike Macey and Mike Wang and Miquel Jubert Hermoso and Mo Metanat and Mohammad Rastegari and Munish Bansal and Nandhini Santhanam and Natascha Parks and Natasha White and Navyata Bawa and Nayan Singhal and Nick Egebo and Nicolas Usunier and Nikhil Mehta and Nikolay Pavlovich Laptev and Ning Dong and Norman Cheng and Oleg Chernoguz and Olivia Hart and Omkar Salpekar and Ozlem Kalinli and Parkin Kent and Parth Parekh and Paul Saab and Pavan Balaji and Pedro Rittner and Philip Bontrager and Pierre Roux and Piotr Dollar and Polina Zvyagina and Prashant Ratanchandani and Pritish Yuvraj and Qian Liang and Rachad Alao and Rachel Rodriguez and Rafi Ayub and Raghotham Murthy and Raghu Nayani and Rahul Mitra and Rangaprabhu Parthasarathy and Raymond Li and Rebekkah Hogan and Robin Battey and Rocky Wang and Russ Howes and Ruty Rinott and Sachin Mehta and Sachin Siby and Sai Jayesh Bondu and Samyak Datta and Sara Chugh and Sara Hunt and Sargun Dhillon and Sasha Sidorov and Satadru Pan and Saurabh Mahajan and Saurabh Verma and Seiji Yamamoto and Sharadh Ramaswamy and Shaun Lindsay and Shaun Lindsay and Sheng Feng and Shenghao Lin and Shengxin Cindy Zha and Shishir Patil and Shiva Shankar and Shuqiang Zhang and Shuqiang Zhang and Sinong Wang and Sneha Agarwal and Soji Sajuyigbe and Soumith Chintala and Stephanie Max and Stephen Chen and Steve Kehoe and Steve Satterfield and Sudarshan Govindaprasad and Sumit Gupta and Summer Deng and Sungmin Cho and Sunny Virk and Suraj Subramanian and Sy Choudhury and Sydney Goldman and Tal Remez and Tamar Glaser and Tamara Best and Thilo Koehler and Thomas Robinson and Tianhe Li and Tianjun Zhang and Tim Matthews and Timothy Chou and Tzook Shaked and Varun Vontimitta and Victoria Ajayi and Victoria Montanez and Vijai Mohan and Vinay Satish Kumar and Vishal Mangla and Vlad Ionescu and Vlad Poenaru and Vlad Tiberiu Mihailescu and Vladimir Ivanov and Wei Li and Wenchen Wang and Wenwen Jiang and Wes Bouaziz and Will Constable and Xiaocheng Tang and Xiaojian Wu and Xiaolan Wang and Xilun Wu and Xinbo Gao and Yaniv Kleinman and Yanjun Chen and Ye Hu and Ye Jia and Ye Qi and Yenda Li and Yilin Zhang and Ying Zhang and Yossi Adi and Youngjin Nam and Yu and Wang and Yu Zhao and Yuchen Hao and Yundi Qian and Yunlu Li and Yuzi He and Zach Rait and Zachary DeVito and Zef Rosnbrick and Zhaoduo Wen and Zhenyu Yang and Zhiwei Zhao and Zhiyu Ma},
      year={2024},
      eprint={2407.21783},
      archivePrefix={arXiv},
      primaryClass={cs.AI},
      url={https://arxiv.org/abs/2407.21783}, 
}

@misc{openai2024gpt4technicalreport,
      title={GPT-4 Technical Report}, 
      author={OpenAI and Josh Achiam and Steven Adler and Sandhini Agarwal and Lama Ahmad and Ilge Akkaya and Florencia Leoni Aleman and Diogo Almeida and Janko Altenschmidt and Sam Altman and Shyamal Anadkat and Red Avila and Igor Babuschkin and Suchir Balaji and Valerie Balcom and Paul Baltescu and Haiming Bao and Mohammad Bavarian and Jeff Belgum and Irwan Bello and Jake Berdine and Gabriel Bernadett-Shapiro and Christopher Berner and Lenny Bogdonoff and Oleg Boiko and Madelaine Boyd and Anna-Luisa Brakman and Greg Brockman and Tim Brooks and Miles Brundage and Kevin Button and Trevor Cai and Rosie Campbell and Andrew Cann and Brittany Carey and Chelsea Carlson and Rory Carmichael and Brooke Chan and Che Chang and Fotis Chantzis and Derek Chen and Sully Chen and Ruby Chen and Jason Chen and Mark Chen and Ben Chess and Chester Cho and Casey Chu and Hyung Won Chung and Dave Cummings and Jeremiah Currier and Yunxing Dai and Cory Decareaux and Thomas Degry and Noah Deutsch and Damien Deville and Arka Dhar and David Dohan and Steve Dowling and Sheila Dunning and Adrien Ecoffet and Atty Eleti and Tyna Eloundou and David Farhi and Liam Fedus and Niko Felix and Simón Posada Fishman and Juston Forte and Isabella Fulford and Leo Gao and Elie Georges and Christian Gibson and Vik Goel and Tarun Gogineni and Gabriel Goh and Rapha Gontijo-Lopes and Jonathan Gordon and Morgan Grafstein and Scott Gray and Ryan Greene and Joshua Gross and Shixiang Shane Gu and Yufei Guo and Chris Hallacy and Jesse Han and Jeff Harris and Yuchen He and Mike Heaton and Johannes Heidecke and Chris Hesse and Alan Hickey and Wade Hickey and Peter Hoeschele and Brandon Houghton and Kenny Hsu and Shengli Hu and Xin Hu and Joost Huizinga and Shantanu Jain and Shawn Jain and Joanne Jang and Angela Jiang and Roger Jiang and Haozhun Jin and Denny Jin and Shino Jomoto and Billie Jonn and Heewoo Jun and Tomer Kaftan and Łukasz Kaiser and Ali Kamali and Ingmar Kanitscheider and Nitish Shirish Keskar and Tabarak Khan and Logan Kilpatrick and Jong Wook Kim and Christina Kim and Yongjik Kim and Jan Hendrik Kirchner and Jamie Kiros and Matt Knight and Daniel Kokotajlo and Łukasz Kondraciuk and Andrew Kondrich and Aris Konstantinidis and Kyle Kosic and Gretchen Krueger and Vishal Kuo and Michael Lampe and Ikai Lan and Teddy Lee and Jan Leike and Jade Leung and Daniel Levy and Chak Ming Li and Rachel Lim and Molly Lin and Stephanie Lin and Mateusz Litwin and Theresa Lopez and Ryan Lowe and Patricia Lue and Anna Makanju and Kim Malfacini and Sam Manning and Todor Markov and Yaniv Markovski and Bianca Martin and Katie Mayer and Andrew Mayne and Bob McGrew and Scott Mayer McKinney and Christine McLeavey and Paul McMillan and Jake McNeil and David Medina and Aalok Mehta and Jacob Menick and Luke Metz and Andrey Mishchenko and Pamela Mishkin and Vinnie Monaco and Evan Morikawa and Daniel Mossing and Tong Mu and Mira Murati and Oleg Murk and David Mély and Ashvin Nair and Reiichiro Nakano and Rajeev Nayak and Arvind Neelakantan and Richard Ngo and Hyeonwoo Noh and Long Ouyang and Cullen O'Keefe and Jakub Pachocki and Alex Paino and Joe Palermo and Ashley Pantuliano and Giambattista Parascandolo and Joel Parish and Emy Parparita and Alex Passos and Mikhail Pavlov and Andrew Peng and Adam Perelman and Filipe de Avila Belbute Peres and Michael Petrov and Henrique Ponde de Oliveira Pinto and Michael and Pokorny and Michelle Pokrass and Vitchyr H. Pong and Tolly Powell and Alethea Power and Boris Power and Elizabeth Proehl and Raul Puri and Alec Radford and Jack Rae and Aditya Ramesh and Cameron Raymond and Francis Real and Kendra Rimbach and Carl Ross and Bob Rotsted and Henri Roussez and Nick Ryder and Mario Saltarelli and Ted Sanders and Shibani Santurkar and Girish Sastry and Heather Schmidt and David Schnurr and John Schulman and Daniel Selsam and Kyla Sheppard and Toki Sherbakov and Jessica Shieh and Sarah Shoker and Pranav Shyam and Szymon Sidor and Eric Sigler and Maddie Simens and Jordan Sitkin and Katarina Slama and Ian Sohl and Benjamin Sokolowsky and Yang Song and Natalie Staudacher and Felipe Petroski Such and Natalie Summers and Ilya Sutskever and Jie Tang and Nikolas Tezak and Madeleine B. Thompson and Phil Tillet and Amin Tootoonchian and Elizabeth Tseng and Preston Tuggle and Nick Turley and Jerry Tworek and Juan Felipe Cerón Uribe and Andrea Vallone and Arun Vijayvergiya and Chelsea Voss and Carroll Wainwright and Justin Jay Wang and Alvin Wang and Ben Wang and Jonathan Ward and Jason Wei and CJ Weinmann and Akila Welihinda and Peter Welinder and Jiayi Weng and Lilian Weng and Matt Wiethoff and Dave Willner and Clemens Winter and Samuel Wolrich and Hannah Wong and Lauren Workman and Sherwin Wu and Jeff Wu and Michael Wu and Kai Xiao and Tao Xu and Sarah Yoo and Kevin Yu and Qiming Yuan and Wojciech Zaremba and Rowan Zellers and Chong Zhang and Marvin Zhang and Shengjia Zhao and Tianhao Zheng and Juntang Zhuang and William Zhuk and Barret Zoph},
      year={2024},
      eprint={2303.08774},
      archivePrefix={arXiv},
      primaryClass={cs.CL},
      url={https://arxiv.org/abs/2303.08774}, 
}

@misc{wang2023knowledgegraphpromptingmultidocument,
      title={Knowledge Graph Prompting for Multi-Document Question Answering}, 
      author={Yu Wang and Nedim Lipka and Ryan A. Rossi and Alexa Siu and Ruiyi Zhang and Tyler Derr},
      year={2023},
      eprint={2308.11730},
      archivePrefix={arXiv},
      primaryClass={cs.CL},
      url={https://arxiv.org/abs/2308.11730}, 
}

@misc{he2024gretrieverretrievalaugmentedgenerationtextual,
      title={G-Retriever: Retrieval-Augmented Generation for Textual Graph Understanding and Question Answering}, 
      author={Xiaoxin He and Yijun Tian and Yifei Sun and Nitesh V. Chawla and Thomas Laurent and Yann LeCun and Xavier Bresson and Bryan Hooi},
      year={2024},
      eprint={2402.07630},
      archivePrefix={arXiv},
      primaryClass={cs.LG},
      url={https://arxiv.org/abs/2402.07630}, 
}

@misc{sarthi2024raptorrecursiveabstractiveprocessing,
      title={RAPTOR: Recursive Abstractive Processing for Tree-Organized Retrieval}, 
      author={Parth Sarthi and Salman Abdullah and Aditi Tuli and Shubh Khanna and Anna Goldie and Christopher D. Manning},
      year={2024},
      eprint={2401.18059},
      archivePrefix={arXiv},
      primaryClass={cs.CL},
      url={https://arxiv.org/abs/2401.18059}, 
}

@misc{zhao2025e2graphragstreamlininggraphbasedrag,
      title={E\textsuperscript{2}GraphRAG: Streamlining Graph-based RAG for High Efficiency and Effectiveness},
      author={Yibo Zhao and Jiapeng Zhu and Ye Guo and Kangkang He and Xiang Li},
      year={2025},
      eprint={2505.24226},
      archivePrefix={arXiv},
      primaryClass={cs.AI},
      url={https://arxiv.org/abs/2505.24226}, 
}

@misc{guo2025lightragsimplefastretrievalaugmented,
      title={LightRAG: Simple and Fast Retrieval-Augmented Generation}, 
      author={Zirui Guo and Lianghao Xia and Yanhua Yu and Tu Ao and Chao Huang},
      year={2025},
      eprint={2410.05779},
      archivePrefix={arXiv},
      primaryClass={cs.IR},
      url={https://arxiv.org/abs/2410.05779}, 
}

@article{jimenez2024hipporag,
  title={Hipporag: Neurobiologically inspired long-term memory for large language models},
  author={Jimenez Gutierrez, Bernal and Shu, Yiheng and Gu, Yu and Yasunaga, Michihiro and Su, Yu},
  journal={Advances in Neural Information Processing Systems},
  volume={37},
  pages={59532--59569},
  year={2024}
}

@misc{gutiérrez2025ragmemorynonparametriccontinual,
      title={From RAG to Memory: Non-Parametric Continual Learning for Large Language Models}, 
      author={Bernal Jiménez Gutiérrez and Yiheng Shu and Weijian Qi and Sizhe Zhou and Yu Su},
      year={2025},
      eprint={2502.14802},
      archivePrefix={arXiv},
      primaryClass={cs.CL},
      url={https://arxiv.org/abs/2502.14802}, 
}

@article{luo2025gfm,
  title={GFM-RAG: graph foundation model for retrieval augmented generation},
  author={Luo, Linhao and Zhao, Zicheng and Haffari, Gholamreza and Phung, Dinh and Gong, Chen and Pan, Shirui},
  journal={arXiv preprint arXiv:2502.01113},
  year={2025}
}
\endgroup

\clearpage
\appendix
\section*{Appendix}
\addcontentsline{toc}{section}{Appendix}
\setcounter{subsection}{0}
\renewcommand{\thesubsection}{\Alph{subsection}}
\renewcommand{\thesubsubsection}{\Alph{subsection}.\arabic{subsubsection}}
\label{sec:appendix}

\subsection{Citation Context Classification Schema}
\label{subsec:citation_schema}

To precisely quantify the intellectual lineage and contextual dependency of the referenced literature, we formulate the citation analysis as a fine-grained classification task. Each citation mention extracted from the full text is categorized into one of five hierarchical levels based on its contextual significance and direct impact on the proposed research. The scoring rubric is defined as follows:

\begin{description}
    \item[Level 5: Foundational Citation.] 
    
    The current research is inextricably built upon this prior work. Without this reference, the core theory, model, or methodology of the paper would lack its foundational basis. Typical contextual markers include phrases such as \textit{``We build upon...''} or \textit{``Following the methodology of...''}.
    
    \item[Level 4: Strong Citation.] 
    
    The referenced work serves as a primary pillar of support or a critical benchmark. It directly influences the study's experimental design, validation, or conclusions. Contextual markers often include \textit{``As demonstrated by...''} or \textit{``We compare our framework against...''}.
    
    \item[Level 3: Moderate Citation.] 
    
    The citation provides supporting background, empirical evidence, or conceptual inspiration, yet it does not constitute the absolute core foundation. It is frequently utilized to justify the research motivation or establish theoretical associations (e.g., \textit{``Inspired by...''} or \textit{``Consistent with findings in...''}).
    
    \item[Level 2: Contextual Citation.] 
    
    The reference is utilized primarily to map the broader research landscape or delineate related work. The direct methodological link to the current study is relatively weak, commonly appearing in comprehensive literature reviews (e.g., \textit{``Several recent studies have explored...''}).
    
    \item[Level 1: Peripheral Citation.] 
    
    The work is mentioned strictly for academic breadth, historical context, or general overview, exerting no substantive impact on the paper's core logic (e.g., \textit{``For a general overview, see...''}).
\end{description}

\paragraph{Aggregation Strategy}
Given that a single reference may be cited multiple times across different sections of a manuscript, we apply a max-pooling aggregation strategy. Specifically, the definitive classification for a given reference is determined by the maximum score it achieves across all its mentions within the text. Furthermore, references that are listed in the bibliography but lack explicit in-text mentions are conservatively assigned a default score of Level 1 (Peripheral Citation) to prevent logical anomalies in the final reference graph. For quantitative aggregation, each level is explicitly assigned a corresponding integer score ranging from 5 (Foundational) to 1 (Peripheral).

\subsection{Proofs and Constructive Details for Section~\ref{sec:theory}}
\label{sec:appendix_formal}

This appendix gives the projection rules used in the proofs, the proofs of Propositions~\ref{prop:id}, \ref{prop:reach}, and~\ref{prop:coverage}, and the witness construction that certifies the strict gap term of \eqref{eq:recall_bound}. We keep the notation of Section~\ref{sec:theory}: a payload is $\mathcal{P}_D=(\mathcal{V},\mathcal{E}_h)$, and every view is identifier preserving on $\mathcal{V}$.

\subsubsection{Projection Rules}
\label{sec:appendix_view_specs}

For a hyperedge $e=(v_1,\ldots,v_k)\in\mathcal{E}_h$, write $|e|=k$ and $\mathrm{set}(e)=\{v_1,\ldots,v_k\}$. The two views invoked in the proofs below are:
\begin{itemize}
  \item \textbf{Binary view $\Phi_b$.} Keep only arity-$2$ edges: $\Phi_b(\mathcal{P}_D)=\big(\mathcal{V},\,\{e\in\mathcal{E}_h:|e|=2\}\big)$.
  \item \textbf{$N$-ary view $\Phi_n$.} Keep $\{e\in\mathcal{E}_h:|e|\ge 3\}$ and add hyperedges produced by the upgrade operator $U_n$ from the binary skeleton, where $U_n$ promotes any clique whose pairwise edges share a chunk to a single hyperedge.
\end{itemize}
The temporal, person, event, DIY, and scientific-KG views are described in Section~\ref{sec:general_kg} and Section~\ref{subsec:knowledge_network}; each is a deterministic projection on the same $\mathcal{V}$, so the arguments below apply to them without modification.

\subsubsection{Proof of Proposition~\ref{prop:id} (Identifier Preserving Joins)}

\paragraph{Upper bound.}
$\Phi_u(\mathcal{P}_D)$ and $\Phi_v(\mathcal{P}_D)$ live over the same $\mathcal{V}$ with identical identifiers. Build a hash table on $\Phi_u\!\restriction\!K$ keyed by node identifier in $\mathcal{O}(|K|)$ time, then for each $v\in K\cap\Phi_v$ look up $H_u[v]$ in expected $\mathcal{O}(1)$. The total cost is $\mathcal{O}(|K|)$, and matches are exact because identifiers compare by equality.

\paragraph{Lower bound.}
If the two pipelines produce $\mathcal{V}_u\cap\mathcal{V}_v=\varnothing$, no shared key is available and alignment must rely on a per-pair surface-form similarity test $\mathrm{sim}(u,v)$. Without a blocking key or metric structure, all $|\mathcal{V}_u|\cdot|\mathcal{V}_v|$ pairs must be considered in the worst case, giving $\Omega(|\mathcal{V}_u|\cdot|\mathcal{V}_v|)$.

\paragraph{False merge probability.}
Let $\sigma(v)$ denote the surface form of node $v$, and let $H=\{v\in\mathcal{V}:|\sigma^{-1}(\sigma(v))|\ge 2\}$ be the homonym set. If $H$ is empty, there is no surface-form homonym to merge. Otherwise, sample $v\sim\mathrm{Unif}(H)$ and a uniform target $w\sim\mathrm{Unif}(\mathcal{V})$. For each $v\in H$, there are $|\sigma^{-1}(\sigma(v))|-1$ wrong targets with the same surface form. Averaging over $H$ gives \eqref{eq:false_merge}. A matcher that merges identical surface forms therefore has this probability of selecting a spurious merge target under the same sampling model. Shared identifiers do not remove homonyms from the text; they avoid this error because joins compare identifiers rather than $\sigma(v)$. Thus the surface-form false-merge probability is zero for ID-based joins. \hfill$\square$

\subsubsection{Proof of Proposition~\ref{prop:reach} (Cross View Reachability)}

\paragraph{(1) Monotonicity.}
Every view shares $\mathcal{V}$ and contributes its edges to $\mathcal{E}_\cup$, so any $h$-hop walk in $\Phi_v(\mathcal{P}_D)$ is also an $h$-hop walk in $\mathcal{P}_\cup$.

\paragraph{(2) Strict gap.}
Pick a query $q$ and a gold hyperedge $e=(v_1,\ldots,v_k)\in H_{\ge3}^{(h)}(q)$ with $k\ge 3$. The binary view $\Phi_b$ discards $e$ entirely; the $n$-ary view $\Phi_n$ exposes all $k$ endpoints in one hop. For each such $e$, let $B_e$ be the endpoints of $e$ that are reachable from $S(q)$ only through $e$ and are not retained by the binary view. Every node in $\bigcup_e B_e$ is reachable in $\mathcal{P}_\cup$ and not reachable in $\Phi_b(\mathcal{P}_D)$, which gives \eqref{eq:reach_gap}. In the two-anchor witness used below, $B_e=\mathrm{set}(e)\setminus\{v_a,v_b\}$; if these hidden endpoint sets are disjoint across gold hyperedges, the bound reduces to $\sum_e(|e|-2)$.

\paragraph{(3) Hardness.}
The main text only uses the monotonicity and strict-gap parts of Proposition~\ref{prop:reach}. For completeness, we also record why a compact hyperedge explanation cannot in general be recovered cheaply from the binary projection alone. We reduce from \textsc{Edge-Clique-Cover} (ECC), which asks whether the edges of a graph $G$ admit a cover by at most $k$ cliques. Define $\textsc{Hyper-Recover}(G_b,k)$ as the decision problem that asks whether a binary graph $G_b$ is the clique expansion of at most $k$ hyperedges of arity at least two. Given an ECC instance $(G,k)$, set $G_b\!:=\!G$ and ask \textsc{Hyper-Recover}$(G_b,k)$. A YES instance is exactly an edge clique cover of $G$ of size $\le k$, because each hyperedge induces one clique and each clique in the cover can be represented as one hyperedge. The reduction is polynomial, so exact minimum recovery under this projection model is NP-hard. \hfill$\square$

\subsubsection{Proof of Proposition~\ref{prop:coverage} (Candidate Coverage)}

For any single view $v$, edge monotonicity of $\Phi_R$ gives $\Phi_R(\mathcal{P}_\cup,q)\supseteq\Phi_R(\Phi_v(\mathcal{P}_D),q)$ before final ranking and truncation. Therefore the gold candidates retrieved by the best single view are also retrieved by the union view. The extra gold candidates counted by $\Delta_h(q)$ are, by definition, outside every single-view candidate set, so the two contributions are disjoint:
\[
\bigl|\Phi_R(\mathcal{P}_\cup,q)\cap A_h^\star(q)\bigr|
\;\ge\;
\max_v\bigl|\Phi_R(\Phi_v(\mathcal{P}_D),q)\cap A_h^\star(q)\bigr|
\;+\;
\big|A_h^\star(q)\cap\big(\Phi_R(\mathcal{P}_\cup,q)\setminus \bigcup_v \Phi_R(\Phi_v(\mathcal{P}_D),q)\big)\big|.
\]
Dividing by $|A_h^\star(q)|$ yields \eqref{eq:recall_bound}. The Scholar-KG case follows because its view is also identifier preserving on the same $\mathcal{V}$, so it contributes another summand to the union without renaming nodes. \hfill$\square$

\subsubsection{Witness Construction for the Strict-Gap Term}
\label{sec:appendix_witness}

The following procedure instantiates a query for which $\Delta_h(q)$ in \eqref{eq:recall_bound} is strictly positive: (i) pick a hyperedge $e\in\mathcal{E}_h$ with $|e|\ge 3$ that carries a nonempty qualifier $\kappa(e)$; (ii) choose two endpoints $v_a,v_b\in\mathrm{set}(e)$ as anchors; (iii) form the query ``which entities appear with $v_a$ and $v_b$ under qualifier $\kappa(e)$?''. The gold answer set is $\mathrm{set}(e)\setminus\{v_a,v_b\}$. The binary view drops the bundled edge $e$; the temporal bundled view $\Phi_t\circ\Phi_n$ exposes $\mathrm{set}(e)$ in a single hop. Therefore, for this query family, the union view retrieves gold candidates that the binary view does not expose. Iterating over $\mathcal{E}_h^{\ge 3}$ generates the witness corpus used in Section~\ref{sec:eval}.

\subsection{Knowledge Graph Visualization}
\label{sec:kg_visualization}

\begin{figure}[htbp]
    \centering
    \captionsetup[subfigure]{justification=centering, font=small}
    
    \begin{subfigure}[t]{0.48\textwidth}
        \centering
        \includegraphics[width=\textwidth, height=3.8cm]{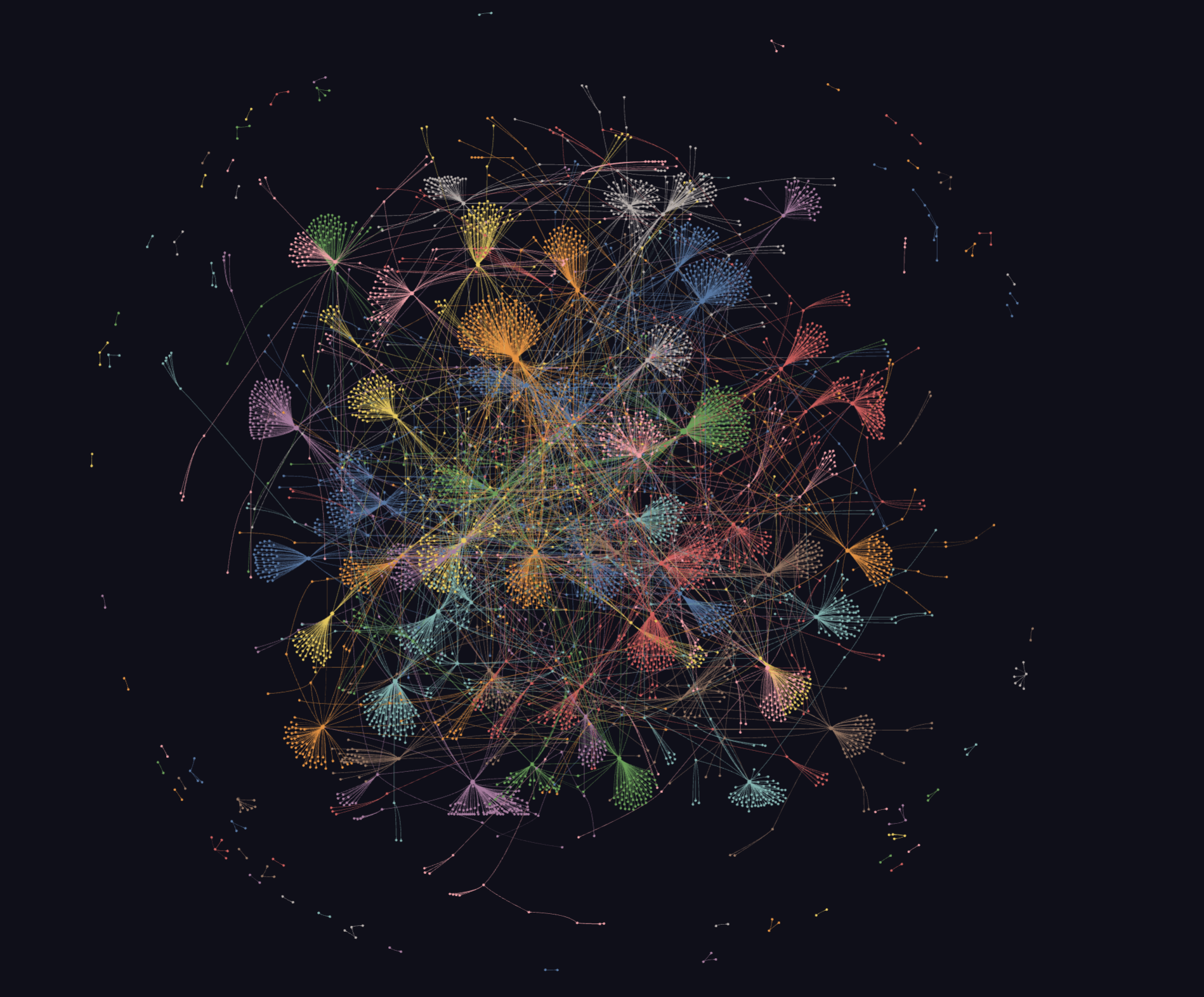}
        \caption{Macroscopic view of 30 seminal papers}
        \label{fig:grid4x2_1}
    \end{subfigure}
    \hfill
    \begin{subfigure}[t]{0.48\textwidth}
        \centering
        \includegraphics[width=\textwidth, height=3.8cm]{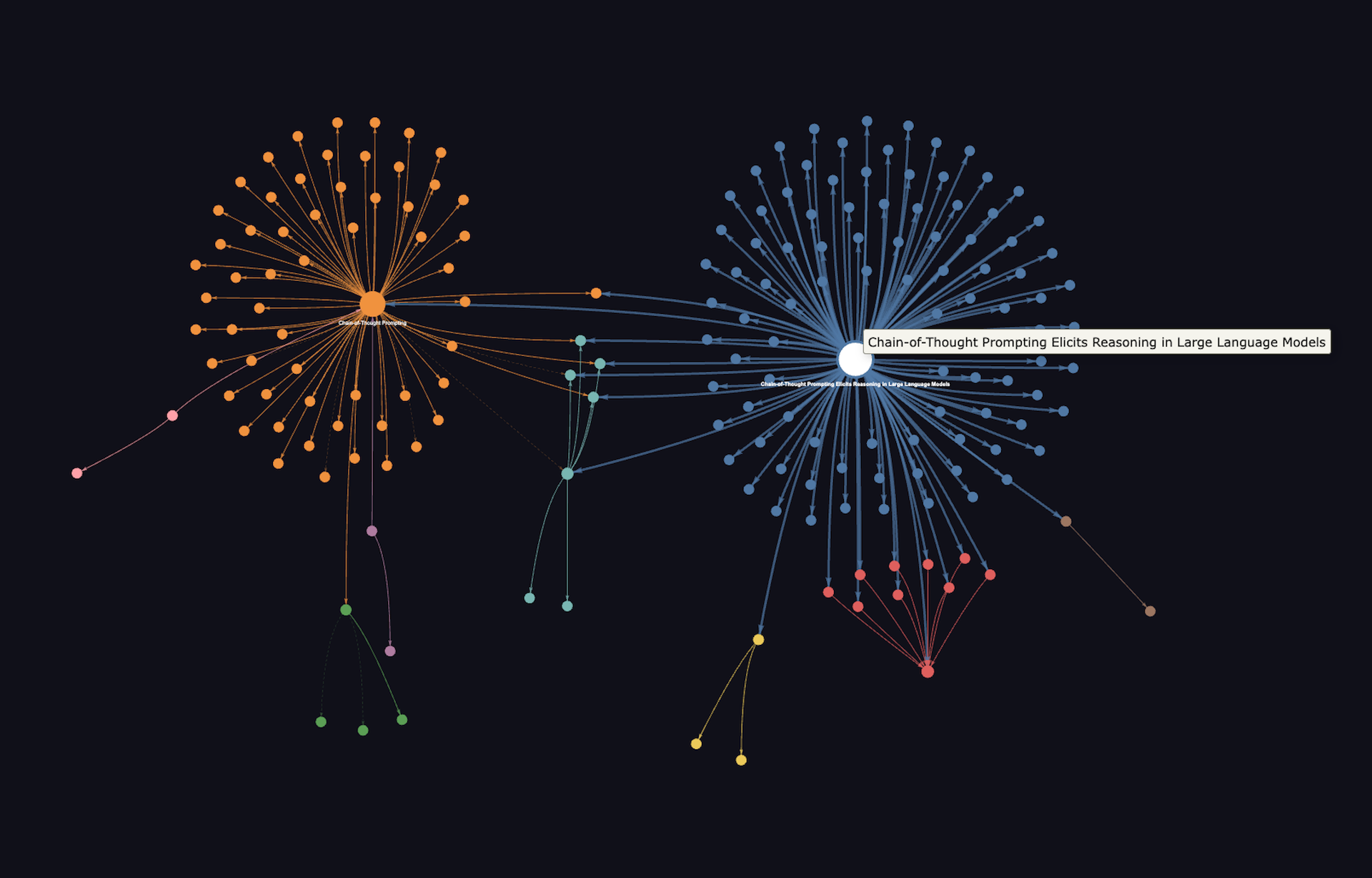}
        \caption{Focal Paper Node: ``Chain-of-Thought Prompting Elicits Reasoning in Large Language Models''(CoT)}
        \label{fig:grid4x2_2}
    \end{subfigure}

    \vspace{1.5ex} 

    \begin{subfigure}[t]{0.48\textwidth}
        \centering
        \includegraphics[width=\textwidth, height=3.8cm]{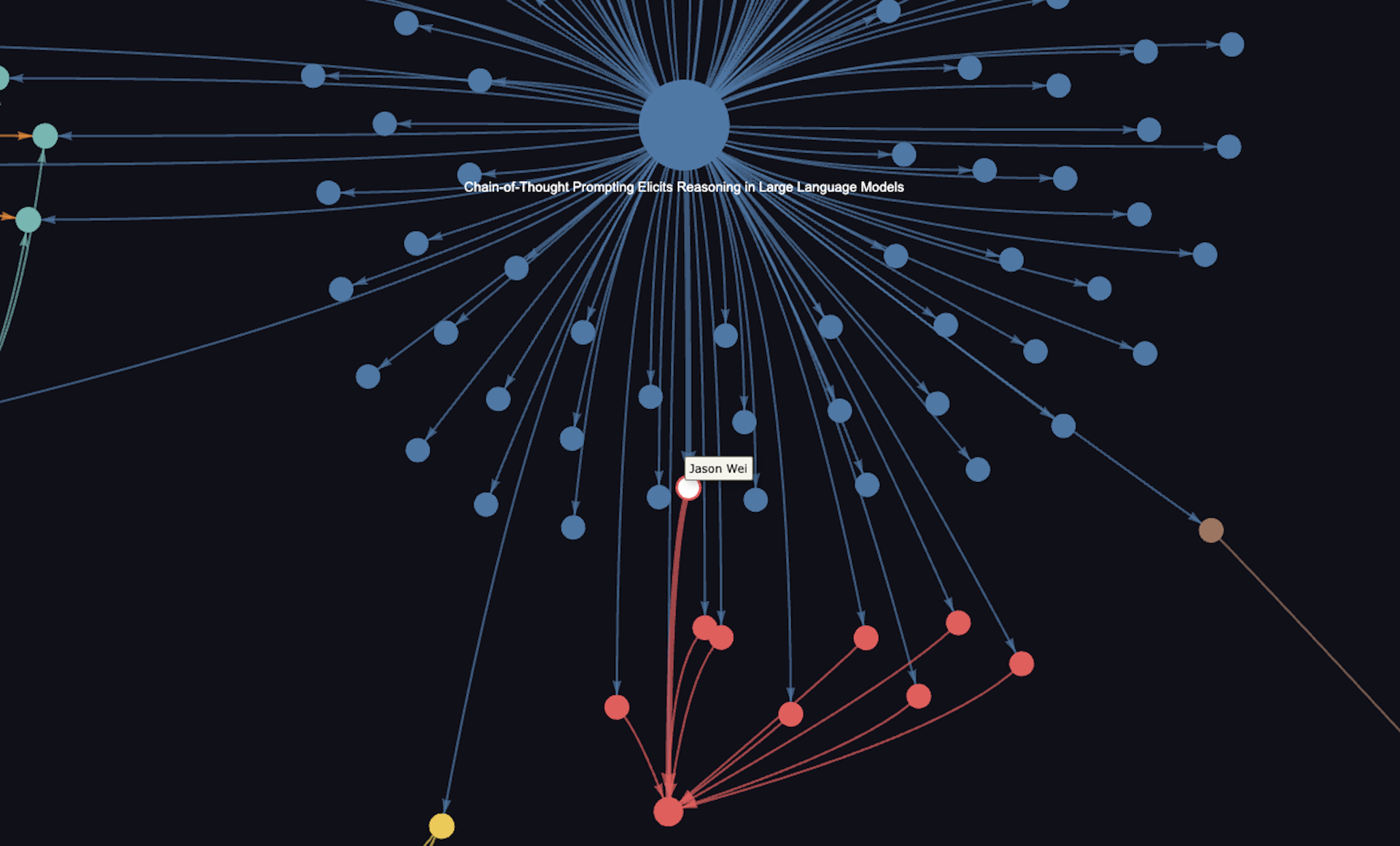} 
        \caption{Paper $\to$ Content:  CoT $\xrightarrow{\textit{authored\_by}}$ Jason Wei \\(Module A: Meta/Factual Entities)}
        \label{fig:grid4x2_3}
    \end{subfigure}
    \hfill
    \begin{subfigure}[t]{0.48\textwidth}
        \centering
        \includegraphics[width=\textwidth, height=3.8cm]{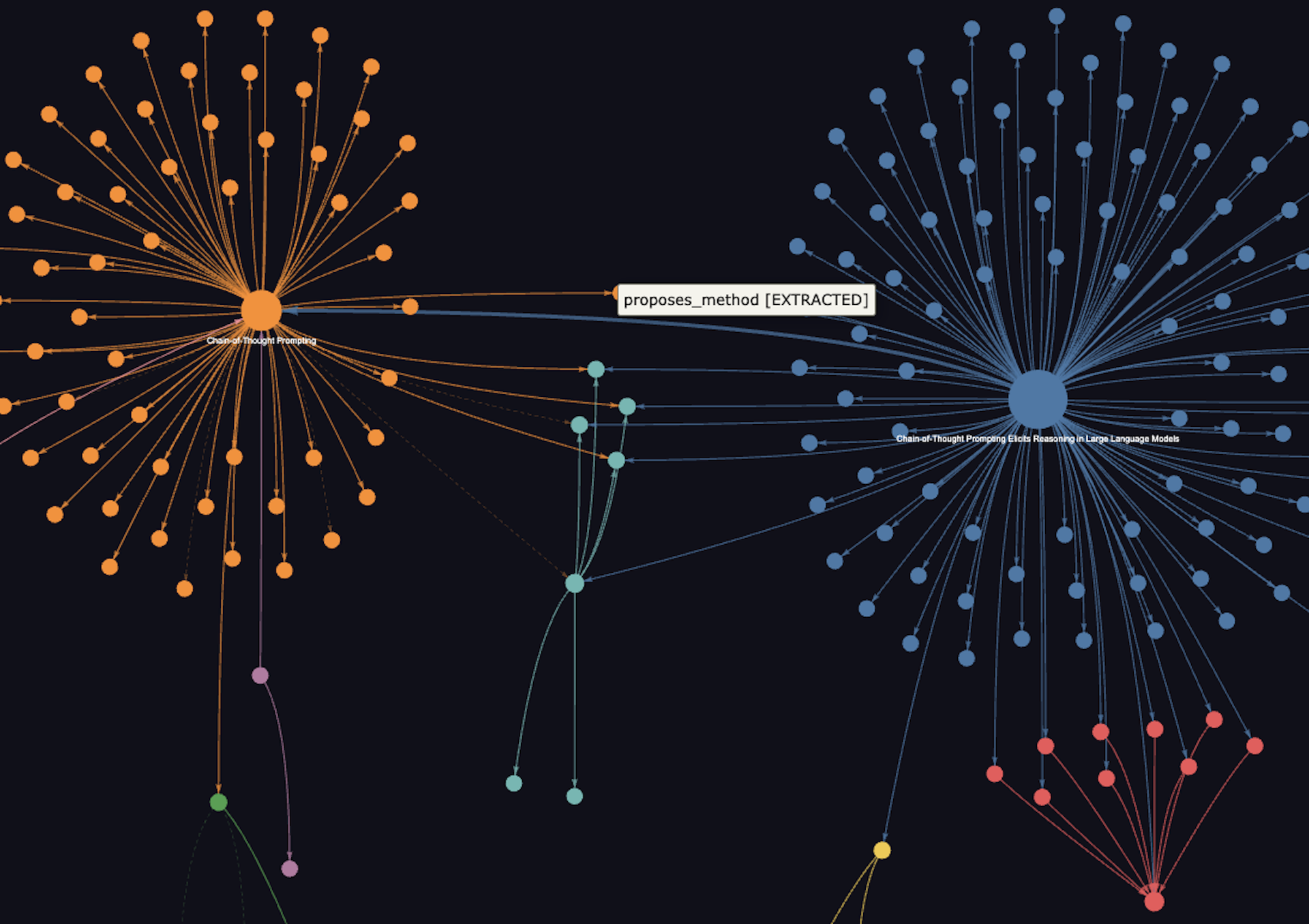}
        \caption{Paper $\to$ Content: CoT $\xrightarrow{\textit{proposes}}$ CoT Prompting \\(Module B: Textually Mentioned Entities)}
        \label{fig:grid4x2_4}
    \end{subfigure}

    \vspace{1.5ex} 

    \begin{subfigure}[t]{0.48\textwidth}
        \centering
        \includegraphics[width=\textwidth, height=3.8cm]{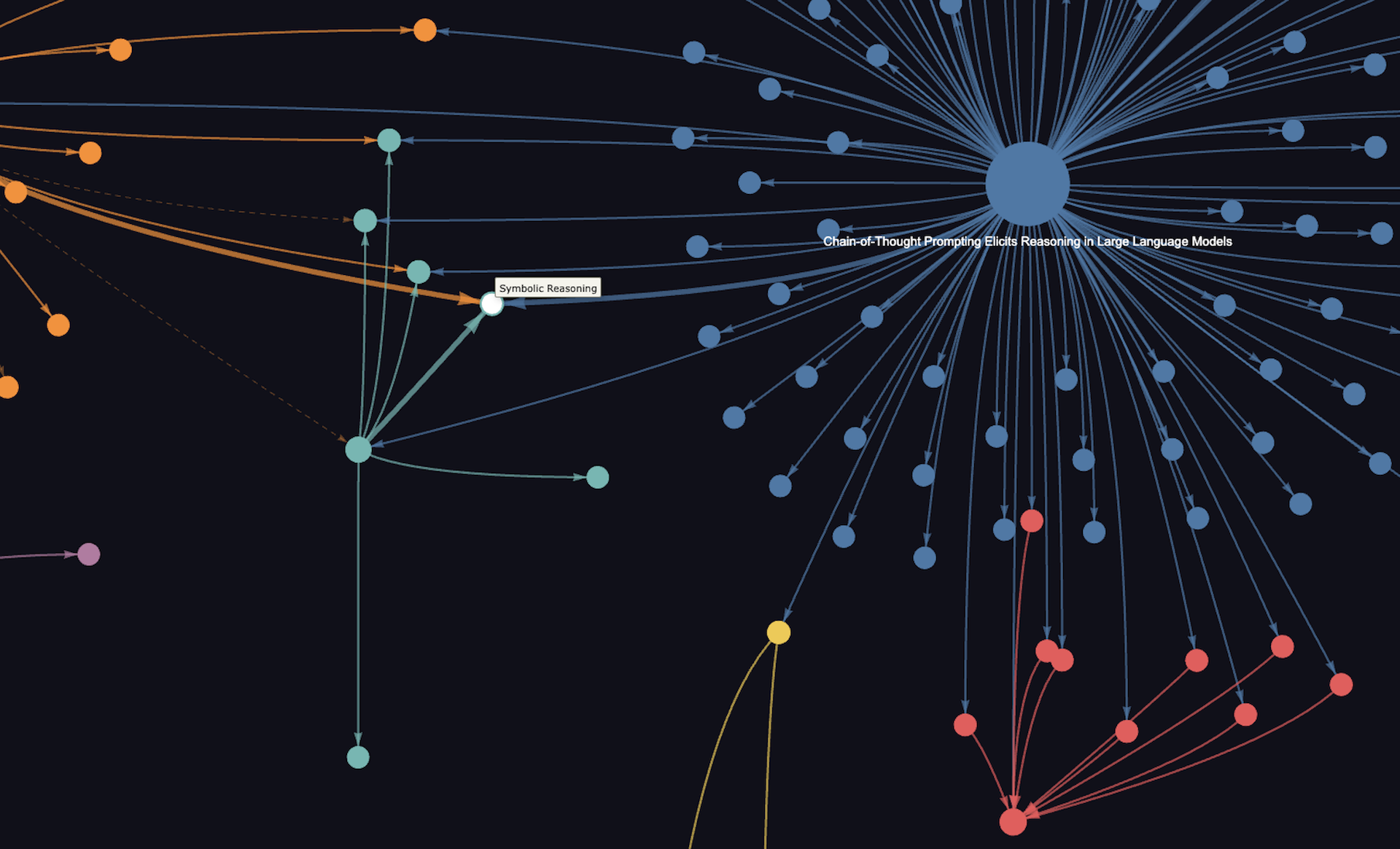}
        \caption{Paper $\to$ Content: \\ CoT $\xrightarrow{\textit{involved\_task}}$  Symbolic Reasoning\\(Module C: Implicit/Abstracted Entities)}
        \label{fig:grid4x2_5}
    \end{subfigure}
    \hfill
    \begin{subfigure}[t]{0.48\textwidth}
        \centering
        \includegraphics[width=\textwidth, height=3.8cm]{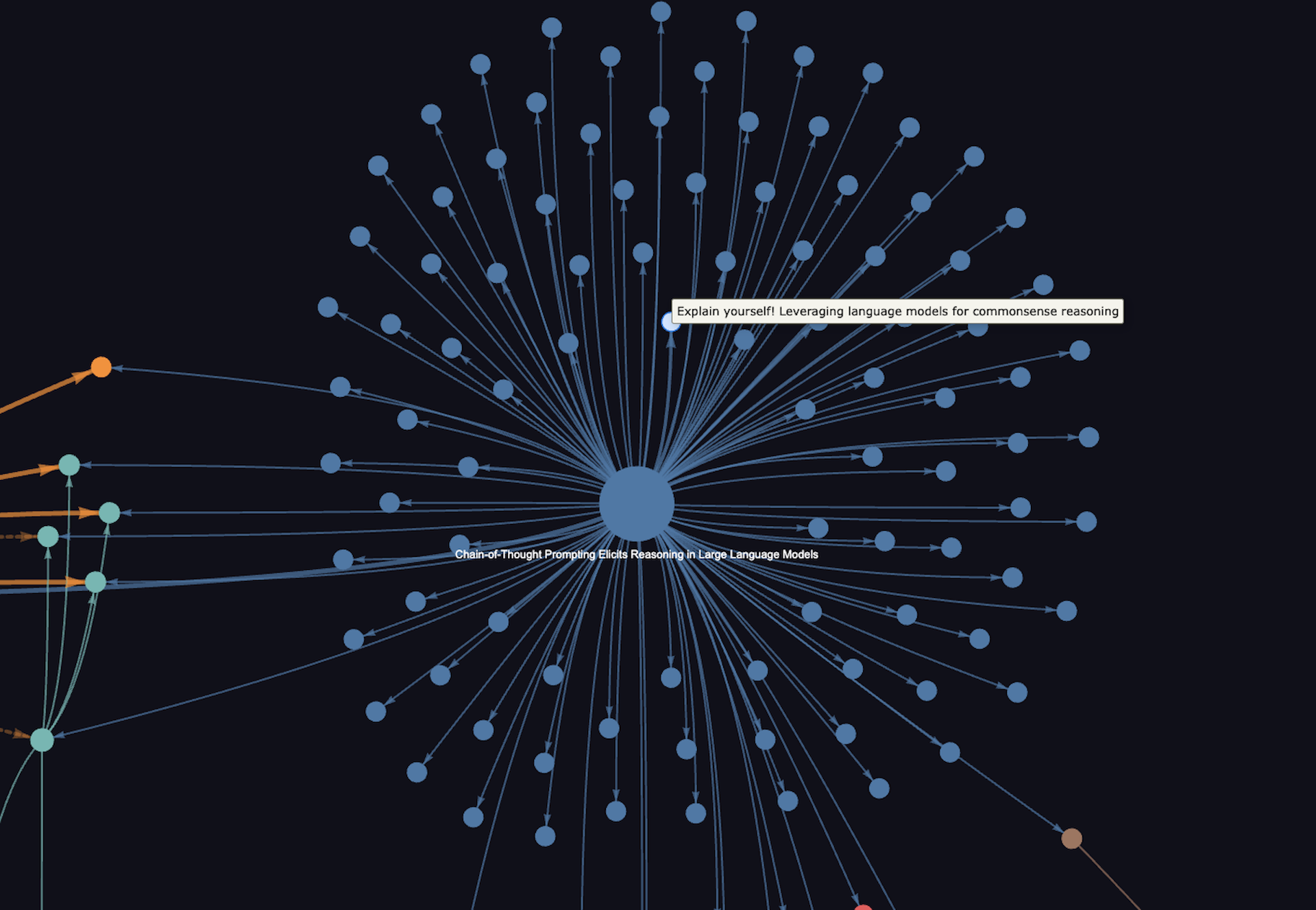}
        \caption{Paper $\to$ Paper:\\ CoT $\xrightarrow{\textit{cites}}$ ``Explain yourself! Leveraging language models for commonsense reasoning''\\(Module D: Citation Relationships)}
        \label{fig:grid4x2_6}
    \end{subfigure}

    \vspace{1.5ex} 

    \begin{subfigure}[t]{0.48\textwidth}
        \centering
        \includegraphics[width=\textwidth, height=3.8cm]{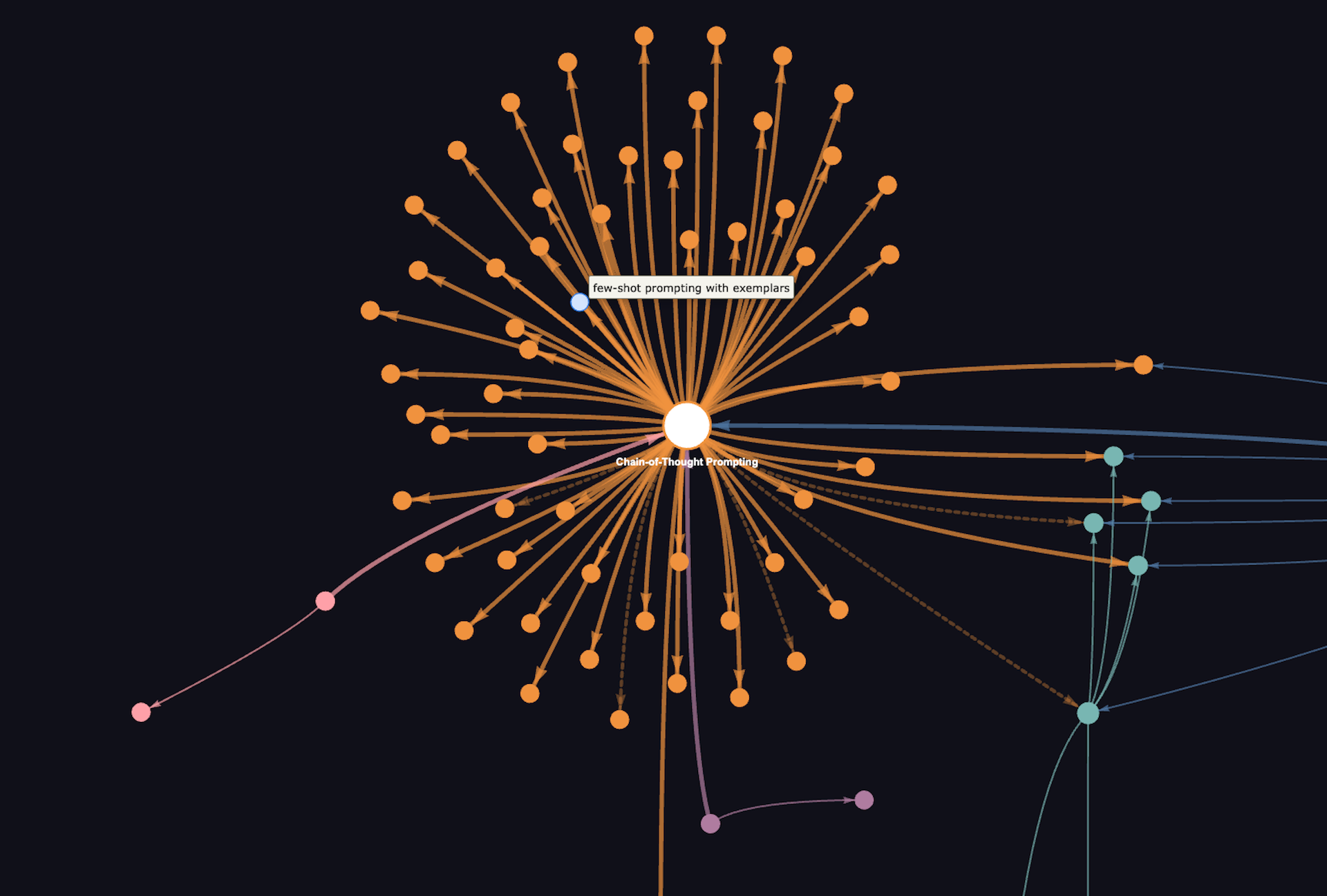}
        \caption{Content $\to$ Content: \\CoT $\xrightarrow{\textit{implements}}$ few-shot prompting}
        \label{fig:grid4x2_7}
    \end{subfigure}
    \hfill
    \begin{subfigure}[t]{0.48\textwidth}
        \centering
        \includegraphics[width=\textwidth, height=3.8cm]{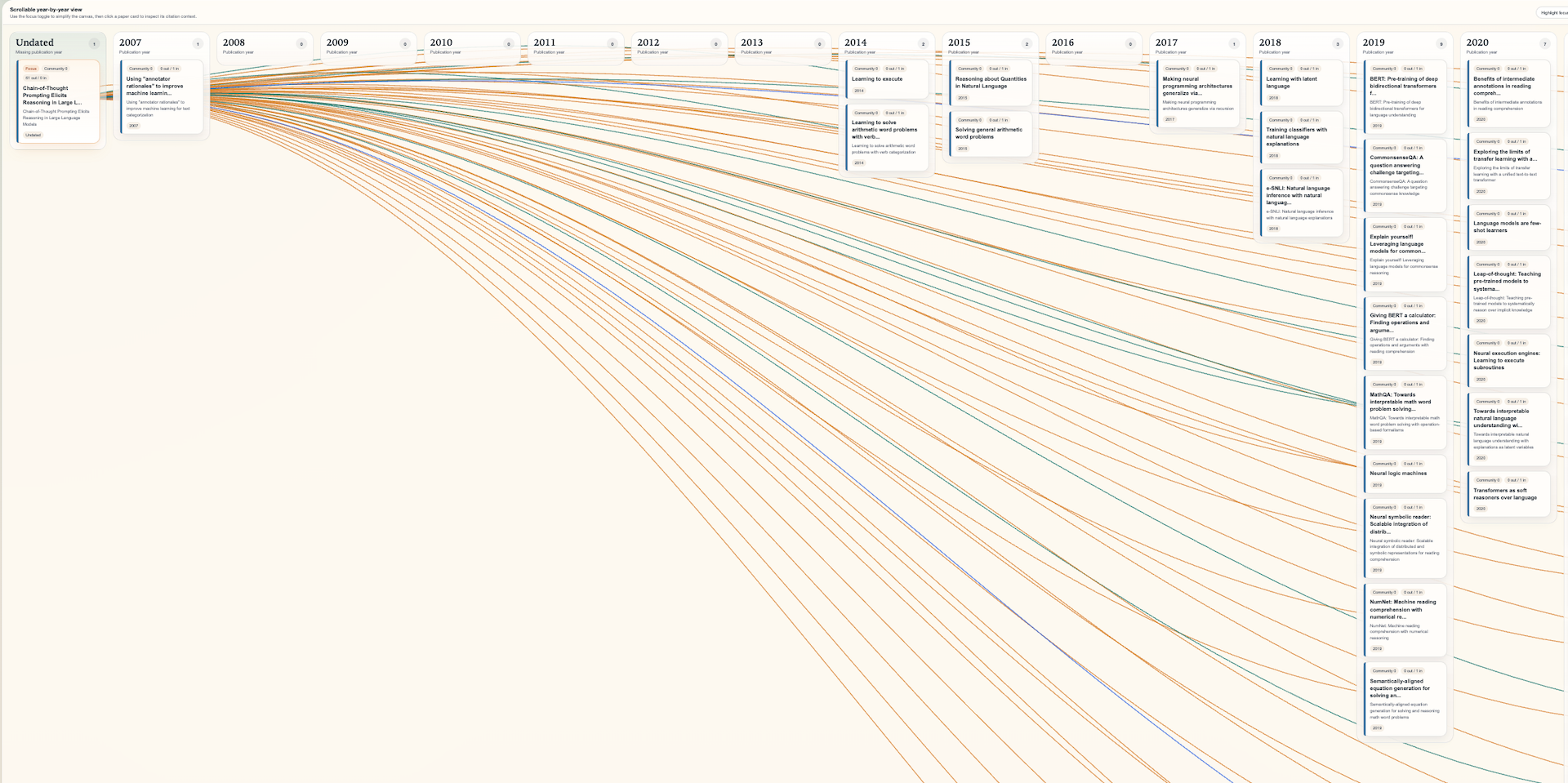}
        \caption{Temporal Evolution: Longitudinal timeline of CoT citations}
        \label{fig:grid4x2_8}
    \end{subfigure}

    \caption{\small Comprehensive, multi-dimensional visualization of the knowledge network. Subfigures (a)--(b) present the global topology and focal nodes. Subfigures (c)--(g) explicitly detail the rich variety of extracted semantic triplets, covering authorship, methodological claims, involved tasks, document citations, and technical implementations. Finally, subfigure (h) maps the focal paper's citation network onto a longitudinal temporal timeline.}
    \label{fig:kg_grid_4x2}
\end{figure}

To provide comprehensive insight into the constructed knowledge network, Figure~\ref{fig:kg_grid_4x2} illustrates a hierarchical breakdown of the extracted subgraphs. Progressively zooming from a macroscopic community structure down to influential focal nodes and specific relational triplets, the visualization highlights our framework's fine-grained parsing capabilities. The densely populated edges denote heterogeneous relationships that integrate the network across multiple granularities: document citations (Paper $\to$ Paper), methodological groundings (Paper $\to$ Content), and semantic dependencies (Content $\to$ Content). Together, these links weave isolated scientific literature into a unified reasoning space.

To further elucidate the extraction capabilities demonstrated in Figure~\ref{fig:kg_grid_4x2}, we provide a detailed examination of the focal node representing the seminal ``Chain-of-Thought Prompting'' (CoT) paper. While subfigures (a) and (b) contextualize the document within the macroscopic topology, subfigures (c) through (f) systematically break down the extracted relational triplets according to our proposed modular taxonomy. 

Specifically, \textbf{Module A} (Figure~\ref{fig:kg_grid_4x2}c) handles explicit meta-factual attributes, accurately anchoring the document to its real-world entities (e.g., the author, Jason Wei). Moving into the semantic content of the PDF, \textbf{Module B} (Figure~\ref{fig:kg_grid_4x2}d) captures textually explicit entities, correctly identifying that the document \textit{proposes} the ``CoT Prompting'' methodology. Beyond surface-level extraction, \textbf{Module C} (Figure~\ref{fig:kg_grid_4x2}e) highlights the framework's advanced natural language understanding by extracting implicit or abstracted entities. It deduces that the paper's \textit{involved\_task} is ``Symbolic Reasoning''—an abstraction that requires synthesizing the paper's core contributions rather than merely executing exact keyword matching.

Furthermore, while \textbf{Module D} (Figure~\ref{fig:kg_grid_4x2}f) preserves traditional document-level structural integrity by mapping explicit citation relationships (Paper $\to$ Paper), our framework fundamentally transcends these document boundaries. As shown in Figure~\ref{fig:kg_grid_4x2}(g), it reveals deep technical dependencies within the semantic space (Content $\to$ Content), recognizing that the CoT concept explicitly \textit{implements} ``few-shot prompting.'' 

Finally, because scientific knowledge is inherently dynamic, Figure~\ref{fig:kg_grid_4x2}(h) projects the focal paper's citation network onto a longitudinal temporal axis. This temporal evolution graph illustrates the developmental trajectory and subsequent academic impact of the CoT methodology. Collectively, this multi-level breakdown validates that our framework does not merely parse text into isolated nodes; rather, it fully reconstructs the underlying scientific reasoning process into a structured, navigable, and temporally-aware knowledge infrastructure.

\subsection{Disaggregated Knowledge Graph Schema}
\label{sec:appendix_json}
To ensure full transparency and provide a comprehensive view of the extraction schema, the complete raw JSON representation of the generated knowledge graph is presented below. This structured data encompasses meta-factual entities, textually mentioned entities, implicitly abstracted variables, and complex relational edges.
\begin{lstlisting}[language=json, title={Knowledge Graph Entity and Relation Extraction (JSON)}]
{
  "A_Meta_Factual_Entities": {
    "Paper": {
      "title": "Adam: A Method for Stochastic Optimization",
      "pub_year": "2014",
      "type": "research paper",
      "language": "English"
    },
    "Authors": [
      {
        "name": "Jimmy Ba",
        "ordering": 0,
        "corresponding_flag": false
      },
      {
        "name": "Diederik P. Kingma",
        "ordering": 1,
        "corresponding_flag": false
      }
    ]
  },
  "B_Textually_Mentioned_Entities": {
    "Tasks": [
      {
        "name": "Stochastic Optimization",
        "type": "optimization",
        "input_modality": "stochastic objective function gradients",
        "output_modality": "optimized parameters",
        "constraints": "high-dimensional parameter spaces, noisy/sparse gradients",
        "aliases": [
          "stochastic gradient descent",
          "adaptive optimization"
        ]
      }
    ],
    "Methods": [
      {
        "name": "Adam",
        "proposed_or_cited": "proposed",
        "components": [
          "adaptive moment estimation",
          "bias correction",
          "exponential moving averages"
        ],
        "training_objectives": [
          "minimize expected value of stochastic objective function"
        ],
        "inference_strategies": [
          "parameter updates using bias-corrected moment estimates"
        ],
        "aliases": [
          "adaptive moment estimation",
          "Adam optimizer"
        ]
      },
      {
        "name": "AdaGrad",
        "proposed_or_cited": "cited",
        "components": [
          "adaptive learning rates for sparse gradients"
        ],
        "training_objectives": [
          "optimize sparse gradient problems"
        ],
        "inference_strategies": [
          "diagonal rescaling of gradients"
        ],
        "aliases": [
          "Adaptive Gradient Algorithm"
        ]
      },
      {
        "name": "RMSProp",
        "proposed_or_cited": "cited",
        "components": [
          "moving average of squared gradients"
        ],
        "training_objectives": [
          "handle non-stationary objectives"
        ],
        "inference_strategies": [
          "divide gradient by running average of magnitudes"
        ],
        "aliases": [
          "Root Mean Square Propagation"
        ]
      },
      {
        "name": "AdaMax",
        "proposed_or_cited": "proposed",
        "components": [
          "infinity norm variant of Adam"
        ],
        "training_objectives": [
          "simplify Adam's bias correction"
        ],
        "inference_strategies": [
          "use L-infinity norm for moment estimation"
        ],
        "aliases": [
          "Adaptive Moment Estimation with Infinity Norm"
        ]
      }
    ],
    "Metrics": [
      {
        "full_name": "Regret Bound",
        "abbreviation": "Regret",
        "formula": "sum_{t=1}^T [f_t(theta_t) - f_t(theta^*)]"
      }
    ],
    "Baselines": [
      {
        "name": "AdaGrad",
        "strong_baseline": true
      },
      {
        "name": "RMSProp",
        "strong_baseline": true
      }
    ],
    "Implementation": {
      "learning_rate": "0.001"
    },
    "Theorems": [
      {
        "number": "4.1",
        "statement": "Adam achieves O(sqrt(T)) regret bound under bounded gradients and parameter constraints",
        "premises": "Bounded gradients, bounded parameter distances, beta_1/sqrt(beta_2) < 1",
        "conclusion": "Regret bound comparable to best known results in online convex optimization"
      }
    ],
    "Figures_Tables": [
      {
        "type": "algorithm",
        "number": "1",
        "caption": "Adam: Proposed algorithm for stochastic optimization",
        "content_summary": "Pseudocode for Adam optimizer with bias-corrected moment estimates"
      }
    ],
    "Domain_Terms": [
      {
        "term": "Signal-to-Noise Ratio (SNR)",
        "standard_name": "SNR",
        "definition": "Ratio of estimated first moment to square root of estimated second moment (m_hat_t / sqrt(v_hat_t))"
      },
      {
        "term": "Bias-Corrected Moment Estimates",
        "standard_name": "Bias Correction",
        "definition": "Adjustment of initial moment estimates to counteract zero initialization bias (m_hat_t = m_t / (1-beta_1^t), v_hat_t = v_t / (1-beta_2^t))"
      }
    ]
  },
  "C_Implicit_Abstracted_Entities": {
    "Problem_Definition": {
      "input_space": "Stochastic objective function f(theta) with parameters theta",
      "output_space": "Optimized parameter vector theta_t minimizing E[f(theta)]",
      "constraints": [
        "High-dimensional parameter spaces",
        "Noisy/sparse gradients",
        "Non-stationary objectives"
      ],
      "assumptions": [
        "Differentiable objective function",
        "Bounded gradients",
        "Exponentially decaying moment estimates"
      ]
    },
    "Motivation": {
      "existing_limitations": [
        "AdaGrad's poor performance on non-stationary objectives",
        "RMSProp's lack of bias correction for initial moments",
        "SGD's sensitivity to learning rate scheduling"
      ],
      "gap_categories": [
        "Need for adaptive learning rates with bias correction",
        "Combining strengths of AdaGrad and RMSProp",
        "Handling sparse gradients in non-stationary environments"
      ]
    },
    "Contributions": {
      "main_contributions": [
        "Proposed Adam optimizer combining AdaGrad and RMSProp advantages",
        "Theoretical analysis of O(sqrt(T)) regret bound",
        "Bias correction technique for moment estimates",
        "AdaMax variant using infinity norm"
      ],
      "component_alignment": [
        "Adaptive learning rates -> AdaGrad/RMSProp combination",
        "Bias correction -> Theoretical convergence guarantees",
        "Regret analysis -> Online convex optimization framework"
      ]
    },
    "Hypotheses": [
      {
        "hypothesis": "Adam's adaptive learning rates with bias correction will outperform existing methods on non-stationary objectives",
        "testable": true,
        "scope": "High-dimensional stochastic optimization problems"
      }
    ],
    "Findings": {
      "quantitative": [
        "Empirical results show Adam consistently outperforms other methods",
        "Regret bound matches best known O(sqrt(T)) results"
      ],
      "qualitative": [
        "Automatic step size annealing through SNR mechanism",
        "Invariance to gradient rescaling improves robustness"
      ]
    },
    "Explanations": [
      {
        "phenomenon": "Adam's convergence guarantees",
        "explanation": "Theoretical analysis shows regret bound depends on gradient norms and parameter distances, with adaptive scaling reducing effective step sizes when SNR is low",
        "type": "theoretical"
      }
    ],
    "Limitations": {
      "generalizability": [
        "Assumes bounded gradients and parameter distances",
        "Theoretical analysis limited to convex objectives"
      ],
      "computational_cost": [
        "Requires storage of first and second moment vectors"
      ]
    },
    "Design_Rationales": [
      "Bias correction addresses initial moment estimates being zero-biased",
      "Exponential moving averages balance recent and historical gradient information",
      "Signal-to-noise ratio mechanism provides automatic step size annealing"
    ],
    "Future_Work": [
      "Extending to non-convex optimization settings",
      "Developing variants with improved memory efficiency",
      "Analyzing performance on specific application domains"
    ],
    "Error_Analysis": {
      "failure_types": [
        "Large initial steps when beta_2 is small without bias correction",
        "Suboptimal convergence when SNR is persistently low"
      ]
    }
  },
  "D_Reference": {
    "Natural gradient works efficiently in learning": {
      "cite_type": "Moderate Cite",
      "related": "The work provides supporting background, as it is referenced to explain the concept of natural gradient descent in relation to Adam.",
      "evidence": [
        3
      ],
      "pub_time": 1998,
      "authors": [
        "Amari, Shun-Ichi"
      ],
      "source": "Neural computation"
    },
    "Recent advances in deep learning for speech research at microsoft": {
      "cite_type": "Moderate Cite",
      "related": "Provides supporting background or inspiration, used to illustrate the impact of SGD in deep learning success stories.",
      "evidence": [
        3
      ],
      "pub_time": 2013,
      "authors": [
        "Deng, Li",
        "Li, Jinyu",
        "Huang, Jui-Ting",
        "Yao, Kaisheng",
        "Yu, Dong",
        "Seide, Frank",
        "Selter, Michael",
        "Zweig, Geoff",
        "He, Xiaodong",
        "Williams, Jason"
      ],
      "source": "ICASSP 2013"
    },
    "Adaptive subgradient methods for online learning and stochastic optimization": {
      "cite_type": "Strong Cite",
      "related": "The work is a main support as the method combines its advantages for sparse gradients.",
      "evidence": [
        4,
        4,
        3,
        4,
        4,
        3,
        3
      ],
      "pub_time": 2011,
      "authors": [
        "Duchi, John",
        "Hazan, Elad",
        "Singer, Yoram"
      ],
      "source": "The Journal of Machine Learning Research"
    },
    "Generating sequences with recurrent neural networks": {
      "cite_type": "Moderate Cite",
      "related": "Provides supporting background on related optimization methods.",
      "evidence": [
        3,
        3
      ],
      "pub_time": 2013,
      "authors": [
        "Graves, Alex"
      ],
      "source": "arXiv preprint arXiv:1308.0850"
    },
    "Speech recognition with deep recurrent neural networks": {
      "cite_type": "Moderate Cite",
      "related": "Provides supporting background or inspiration, used to illustrate the impact of SGD in deep learning success stories.",
      "evidence": [
        3
      ],
      "pub_time": 2013,
      "authors": [
        "Graves, Alex",
        "Mohamed, Abdel-rahman",
        "Hinton, Geoffrey"
      ],
      "source": "In Acoustics, Speech and Signal Processing (ICASSP), 2013 IEEE International Conference on"
    },
    "Reducing the dimensionality of data with neural networks": {
      "cite_type": "Moderate Cite",
      "related": "Provides supporting background or inspiration, used to illustrate the impact of SGD in deep learning success stories.",
      "evidence": [
        3
      ],
      "pub_time": 2006,
      "authors": [
        "Hinton, G.E.",
        "Salakhutdinov, R.R."
      ],
      "source": "Science"
    },
    "Deep neural networks for acoustic modeling in speech recognition: The shared views of four research groups": {
      "cite_type": "Moderate Cite",
      "related": "Provides supporting background or inspiration, used to illustrate the impact of SGD in deep learning success stories.",
      "evidence": [
        3
      ],
      "pub_time": 2012,
      "authors": [
        "Hinton, Geoffrey",
        "Deng, Li",
        "Yu, Dong",
        "Dahl, George E",
        "Mohamed, Abdel-rahman",
        "Jaitly, Navdeep",
        "Senior, Andrew",
        "Vanhoucke, Vincent",
        "Nguyen, Patrick",
        "Sainath, Tara N"
      ],
      "source": "Signal Processing Magazine, IEEE"
    },
    "Improving neural networks by preventing co-adaptation of feature detectors": {
      "cite_type": "Moderate Cite",
      "related": "Provides supporting background or inspiration, used to illustrate additional sources of noise in objectives.",
      "evidence": [
        3
      ],
      "pub_time": 2012,
      "authors": [
        "Hinton, Geoffrey E",
        "Srivastava, Nitish",
        "Krizhevsky, Alex",
        "Sutskever, Ilya",
        "Salakhutdinov, Ruslan R"
      ],
      "source": "arXiv preprint arXiv:1207.0580"
    },
    "Auto-Encoding Variational Bayes": {
      "cite_type": "Strong Cite",
      "related": "The work is a main support or benchmark, as the paper uses the same VAE architecture described in this work for training.",
      "evidence": [
        4
      ],
      "pub_time": 2013,
      "authors": [
        "Kingma, Diederik P",
        "Welling, Max"
      ],
      "source": "In The 2nd International Conference on Learning Representations (ICLR)"
    },
    "Imagenet classification with deep convolutional neural networks": {
      "cite_type": "Moderate Cite",
      "related": "Provides supporting background or inspiration, used to illustrate the impact of SGD in deep learning success stories.",
      "evidence": [
        3
      ],
      "pub_time": 2012,
      "authors": [
        "Krizhevsky, Alex",
        "Sutskever, Ilya",
        "Hinton, Geoffrey E"
      ],
      "source": "In Advances in neural information processing systems"
    },
    "Learning word vectors for sentiment analysis": {
      "cite_type": "Strong Cite",
      "related": "The dataset from this work is directly used in the experiments, making it a key benchmark for the study.",
      "evidence": [
        4
      ],
      "pub_time": 2011,
      "authors": [
        "Maas, Andrew L",
        "Daly, Raymond E",
        "Pham, Peter T",
        "Huang, Dan",
        "Ng, Andrew Y",
        "Potts, Christopher"
      ],
      "source": "In Proceedings of the 49th Annual Meeting of the Association for Computational Linguistics: Human Language Technologies-Volume 1"
    },
    "Non-asymptotic analysis of stochastic approximation algorithms for machine learning": {
      "cite_type": "Strong Cite",
      "related": "The work is a main support, as it demonstrates the improvement of convergence using Polyak-Ruppert averaging.",
      "evidence": [
        4
      ],
      "pub_time": 2011,
      "authors": [
        "Moulines, Eric",
        "Bach, Francis R"
      ],
      "source": "In Advances in Neural Information Processing Systems"
    },
    "Revisiting natural gradient for deep networks": {
      "cite_type": "Moderate Cite",
      "related": "The work provides supporting background, as it is referenced to explain the Fisher information matrix approximation used in Adam.",
      "evidence": [
        3
      ],
      "pub_time": 2013,
      "authors": [
        "Pascanu, Razvan",
        "Bengio, Yoshua"
      ],
      "source": "arXiv preprint arXiv:1301.3584"
    },
    "Acceleration of stochastic approximation by averaging": {
      "cite_type": "Strong Cite",
      "related": "The work is a main support, as it introduces the Polyak-Ruppert averaging method discussed in the text.",
      "evidence": [
        4
      ],
      "pub_time": 1992,
      "authors": [
        "Polyak, Boris T",
        "Juditsky, Anatoli B"
      ],
      "source": "SIAM Journal on Control and Optimization"
    },
    "A fast natural newton method": {
      "cite_type": "Moderate Cite",
      "related": "Provides supporting background on related optimization methods.",
      "evidence": [
        3
      ],
      "pub_time": 2010,
      "authors": [
        "Roux, Nicolas L",
        "Fitzgibbon, Andrew W"
      ],
      "source": "In Proceedings of the 27th International Conference on Machine Learning (ICML-10)"
    },
    "Efficient estimations from a slowly convergent robbins-monro process": {
      "cite_type": "Strong Cite",
      "related": "The work is a main support, as it introduces the Polyak-Ruppert averaging method discussed in the text.",
      "evidence": [
        4
      ],
      "pub_time": 1988,
      "authors": [
        "Ruppert, David"
      ],
      "source": "Technical report, Cornell University Operations Research and Industrial Engineering"
    },
    "No more pesky learning rates": {
      "cite_type": "Moderate Cite",
      "related": "Provides supporting background on related optimization methods.",
      "evidence": [
        3
      ],
      "pub_time": 2012,
      "authors": [
        "Schaul, Tom",
        "Zhang, Sixin",
        "LeCun, Yann"
      ],
      "source": "arXiv preprint arXiv:1206.1106"
    },
    "Fast large-scale optimization by unifying stochastic gradient and quasi-newton methods": {
      "cite_type": "Strong Cite",
      "related": "The work is a main support or benchmark, as the paper compares Adam's memory requirements to those of SFO.",
      "evidence": [
        4,
        4
      ],
      "pub_time": 2014,
      "authors": [
        "Sohl-Dickstein, Jascha",
        "Poole, Ben",
        "Ganguli, Surya"
      ],
      "source": "In Proceedings of the 31st International Conference on Machine Learning (ICML-14)"
    },
    "On the importance of initialization and momentum in deep learning": {
      "cite_type": "Moderate Cite",
      "related": "The work provides supporting evidence for the importance of decaying the momentum coefficient, but is not central to the paper's core theory.",
      "evidence": [
        3
      ],
      "pub_time": 2013,
      "authors": [
        "Sutskever, Ilya",
        "Martens, James",
        "Dahl, George",
        "Hinton, Geoffrey"
      ],
      "source": "In Proceedings of the 30th International Conference on Machine Learning (ICML-13)"
    },
    "Lecture 6.5 - RMSProp": {
      "cite_type": "Strong Cite",
      "related": "The work is a main support as the method combines its advantages for on-line and non-stationary settings.",
      "evidence": [
        4,
        3,
        4,
        4,
        3,
        3
      ],
      "pub_time": 2012,
      "authors": [
        "Tieleman, T.",
        "Hinton, G."
      ],
      "source": "COURSERA: Neural Networks for Machine Learning"
    },
    "Fast dropout training": {
      "cite_type": "Moderate Cite",
      "related": "The paper uses a suggestion from this work as background for applying dropout noise during training.",
      "evidence": [
        3
      ],
      "pub_time": 2013,
      "authors": [
        "Wang, Sida",
        "Manning, Christopher"
      ],
      "source": "In Proceedings of the 30th International Conference on Machine Learning (ICML-13)"
    },
    "Adadelta: An adaptive learning rate method": {
      "cite_type": "Moderate Cite",
      "related": "Provides supporting background on related optimization methods.",
      "evidence": [
        3
      ],
      "pub_time": 2012,
      "authors": [
        "Zeiler, Matthew D"
      ],
      "source": "arXiv preprint arXiv:1212.5701"
    },
    "Online convex programming and generalized infinitesimal gradient ascent": {
      "cite_type": "Strong Cite",
      "related": "The work is a main support for the study's design, as the online learning framework proposed in this paper is directly used for the convergence analysis.",
      "evidence": [
        4
      ],
      "pub_time": 2003,
      "authors": [
        "Zinkevich, Martin"
      ]
    }
  },
  "E_Relations": [
    {
      "head": "Adam",
      "head_type": "Method",
      "relation": "BUILDS_ON",
      "tail": "AdaGrad",
      "tail_type": "Method",
      "evidence": "Our method is designed to combine the advantages of two recently popular methods: AdaGrad (Duchi et al., 2011), which works well with sparse gradients, and RMSProp (Fieleman & Hinton, 2012), which works well in on-line and non-stationary settings",
      "confidence": "high",
      "category": "controlled",
      "source": "semantic"
    },
    {
      "head": "Adam",
      "head_type": "Method",
      "relation": "BUILDS_ON",
      "tail": "RMSProp",
      "tail_type": "Method",
      "evidence": "Our method is designed to combine the advantages of two recently popular methods: AdaGrad (Duchi et al., 2011), which works well with sparse gradients, and RMSProp (Fieleman & Hinton, 2012), which works well in on-line and non-stationary settings",
      "confidence": "high",
      "category": "controlled",
      "source": "semantic"
    },
    {
      "head": "Adam",
      "head_type": "Method",
      "relation": "SOLVES",
      "tail": "Stochastic Optimization",
      "tail_type": "Task",
      "evidence": "We propose Adam, a method for efficient stochastic optimization that only requires first-order gradients with little memory requirement",
      "confidence": "high",
      "category": "controlled",
      "source": "semantic"
    },
    {
      "head": "Adam",
      "head_type": "Method",
      "relation": "USES_TECHNIQUE",
      "tail": "Bias-Corrected Moment Estimates",
      "tail_type": "Domain_Term",
      "evidence": "The method computes individual adaptive learning rates for different parameters from estimates of first and second moments of the gradients",
      "confidence": "high",
      "category": "open",
      "source": "semantic"
    },
    {
      "head": "Adam",
      "head_type": "Method",
      "relation": "REQUIRES",
      "tail": "first-order gradients",
      "tail_type": "Requirement",
      "evidence": "We propose Adam, a method for efficient stochastic optimization that only requires first-order gradients with little memory requirement",
      "confidence": "high",
      "category": "open",
      "source": "semantic"
    },
    {
      "head": "Adam",
      "head_type": "Method",
      "relation": "HAS_LIMITATION",
      "tail": "restricted to first-order methods",
      "tail_type": "Limitation",
      "evidence": "discussion in this paper will be restricted to first-order methods",
      "confidence": "medium",
      "category": "open",
      "source": "semantic"
    },
    {
      "head": "Adam",
      "head_type": "Method",
      "relation": "USES_TECHNIQUE",
      "tail": "exponential moving averages",
      "tail_type": "Technique",
      "evidence": "The algorithm updates exponential moving averages of the gradient (m_t) and the squared gradient (v_t)",
      "confidence": "high",
      "category": "open",
      "source": "semantic"
    },
    {
      "head": "Adam",
      "head_type": "Method",
      "relation": "USES_TECHNIQUE",
      "tail": "bias-corrected moment estimates",
      "tail_type": "Domain_Term",
      "evidence": "this initialization bias can be easily counteracted, resulting in bias-corrected estimates m_hat_t and v_hat_t",
      "confidence": "high",
      "category": "open",
      "source": "semantic"
    },
    {
      "head": "Adam",
      "head_type": "Method",
      "relation": "USES_TECHNIQUE",
      "tail": "adaptive stepsize selection",
      "tail_type": "Technique",
      "evidence": "An important property of Adam's update rule is its careful choice of stepsizes",
      "confidence": "high",
      "category": "open",
      "source": "semantic"
    },
    {
      "head": "Adam",
      "head_type": "Method",
      "relation": "REQUIRES",
      "tail": "Signal-to-Noise Ratio (SNR)",
      "tail_type": "Domain_Term",
      "evidence": "The stochasticity might come from the evaluation at random subsamples (minibatches) of datapoints, or arise from inherent function noise",
      "confidence": "medium",
      "category": "open",
      "source": "semantic"
    },
    {
      "head": "Adam",
      "head_type": "Method",
      "relation": "USES_COMPONENT",
      "tail": "Bias-Corrected Moment Estimates",
      "tail_type": "Domain_Term",
      "evidence": "Adam utilizes initialization bias correction terms.",
      "confidence": "high",
      "category": "controlled",
      "source": "semantic"
    },
    {
      "head": "Adam",
      "head_type": "Method",
      "relation": "USES_TECHNIQUE",
      "tail": "exponential moving average",
      "tail_type": "Technique",
      "evidence": "Adam utilizes initialization bias correction terms.",
      "confidence": "medium",
      "category": "open",
      "source": "semantic"
    },
    {
      "head": "Signal-to-Noise Ratio (SNR)",
      "head_type": "Domain_Term",
      "relation": "HAS_PROPERTY",
      "tail": "bounded by stepsize setting alpha",
      "tail_type": "Property",
      "evidence": "The effective magnitude of the steps taken in parameter space at each timestep are approximately bounded by the stepsize setting alpha, i.e., |Delta_t| approximately <= alpha.",
      "confidence": "high",
      "category": "open",
      "source": "semantic"
    },
    {
      "head": "Signal-to-Noise Ratio (SNR)",
      "head_type": "Domain_Term",
      "relation": "MODULATES",
      "tail": "effective stepsize Delta_t",
      "tail_type": "Property",
      "evidence": "With a smaller SNR the effective stepsize Delta_t will be closer to zero.",
      "confidence": "high",
      "category": "open",
      "source": "semantic"
    },
    {
      "head": "Signal-to-Noise Ratio (SNR)",
      "head_type": "Domain_Term",
      "relation": "CORRELATED_WITH",
      "tail": "uncertainty in gradient direction",
      "tail_type": "Property",
      "evidence": "a smaller SNR means that there is greater uncertainty about whether the direction of m_hat_t corresponds to the direction of the true gradient.",
      "confidence": "high",
      "category": "open",
      "source": "semantic"
    },
    {
      "head": "Signal-to-Noise Ratio (SNR)",
      "head_type": "Domain_Term",
      "relation": "CORRELATED_WITH",
      "tail": "automatic annealing",
      "tail_type": "Process",
      "evidence": "the SNR value typically becomes closer to 0 towards an optimum, leading to smaller effective steps in parameter space: a form of automatic annealing.",
      "confidence": "high",
      "category": "open",
      "source": "semantic"
    },
    {
      "head": "effective stepsize Delta_t",
      "head_type": "Property",
      "relation": "INHIBITS",
      "tail": "large parameter updates",
      "tail_type": "Process",
      "evidence": "With a smaller SNR the effective stepsize Delta_t will be closer to zero.",
      "confidence": "medium",
      "category": "open",
      "source": "semantic"
    },
    {
      "head": "Adam",
      "head_type": "Method",
      "relation": "REQUIRES",
      "tail": "initialization bias correction",
      "tail_type": "Requirement",
      "evidence": "Adam utilizes initialization bias correction terms.",
      "confidence": "high",
      "category": "open",
      "source": "semantic"
    },
    {
      "head": "Bias-Corrected Moment Estimates",
      "head_type": "Domain_Term",
      "relation": "IMPLEMENTED_FOR",
      "tail": "second moment estimate",
      "tail_type": "Component",
      "evidence": "We will here derive the term for the second moment estimate; the derivation for the first moment estimate is completely analogous.",
      "confidence": "medium",
      "category": "open",
      "source": "semantic_unknown_type"
    },
    {
      "head": "Bias-Corrected Moment Estimates",
      "head_type": "Domain_Term",
      "relation": "REQUIRES",
      "tail": "stationary true second moment",
      "tail_type": "Condition",
      "evidence": "zeta = 0 if the true second moment E[g_t^2] is stationary; otherwise zeta can be kept small since the exponential decay rate beta_1 can (and should) be chosen such that the exponential moving average assigns small weights to gradients too far in the past.",
      "confidence": "medium",
      "category": "open",
      "source": "semantic"
    },
    {
      "head": "Adam",
      "head_type": "Method",
      "relation": "HAS_LIMITATION",
      "tail": "large initial steps without bias correction",
      "tail_type": "Limitation",
      "evidence": "In case of sparse gradients, for a reliable estimate of the second moment one needs to average over many gradients by choosing a small value of beta_2; however it is exactly this case of small beta_2 where a lack of initialisation bias correction would lead to initial steps that are much larger.",
      "confidence": "high",
      "category": "open",
      "source": "semantic"
    },
    {
      "head": "Adam",
      "head_type": "Method",
      "relation": "HAS_LIMITATION",
      "tail": "requires decaying learning rate at t^{-1/2} and decaying first moment coefficient",
      "tail_type": "Limitation",
      "evidence": "Our following theorem holds when the learning rate alpha_t is decaying at a rate of t^{-1/2} and first moment running average coefficient beta_{1,t} decay exponentially with lambda, that is typically close to 1, e.g. 1-10^{-8}.",
      "confidence": "high",
      "category": "open",
      "source": "semantic"
    },
    {
      "head": "Adam",
      "head_type": "Method",
      "relation": "REQUIRES",
      "tail": "bounded gradients of functions",
      "tail_type": "Requirement",
      "evidence": "Assume that the function f_t has bounded gradients, ||nabla f_t(theta)||_2 <= G, ||nabla f_t(theta)||_infty <= G_infty for all theta in R^d",
      "confidence": "high",
      "category": "open",
      "source": "semantic"
    },
    {
      "head": "Adam",
      "head_type": "Method",
      "relation": "REQUIRES",
      "tail": "bounded distance between generated parameters",
      "tail_type": "Requirement",
      "evidence": "distance between any theta_t generated by Adam is bounded ||theta_n - theta_m||_2 <= D, ||theta_m - theta_n||_infty <= D_infty for any m,n in {1,...,T}",
      "confidence": "high",
      "category": "open",
      "source": "semantic"
    },
    {
      "head": "Adam",
      "head_type": "Method",
      "relation": "REQUIRES",
      "tail": "beta_1 and beta_2 in [0,1) satisfying beta_1 / sqrt(beta_2) < 1",
      "tail_type": "Requirement",
      "evidence": "beta_1, beta_2 in [0,1) satisfy beta_1 / sqrt(beta_2) < 1",
      "confidence": "high",
      "category": "open",
      "source": "semantic"
    },
    {
      "head": "Adam",
      "head_type": "Method",
      "relation": "ACHIEVES",
      "tail": "O(log d sqrt(T)) regret bound",
      "tail_type": "Metric",
      "evidence": "adaptive method, such as Adam and Adagrad, can achieve O(log d sqrt(T))",
      "confidence": "high",
      "category": "open",
      "source": "semantic_unknown_type"
    },
    {
      "head": "Adam",
      "head_type": "Method",
      "relation": "IMPROVES_OVER",
      "tail": "non-adaptive methods with O(sqrt(dT)) regret bound",
      "tail_type": "Method",
      "evidence": "an improvement over O(sqrt(dT)) for the non-adaptive method",
      "confidence": "high",
      "category": "open",
      "source": "semantic_unknown_type"
    },
    {
      "head": "Adam",
      "head_type": "Method",
      "relation": "USES_TECHNIQUE",
      "tail": "decaying first moment coefficient beta_1,t",
      "tail_type": "Technique",
      "evidence": "Decaying beta_1,t towards zero is important in our theoretical analysis",
      "confidence": "high",
      "category": "open",
      "source": "semantic"
    },
    {
      "head": "Adam",
      "head_type": "Method",
      "relation": "USES_TECHNIQUE",
      "tail": "running average of first and second moment of the gradient",
      "tail_type": "Technique",
      "evidence": "Adam updates are directly estimated using a running average of first and second moment of the gradient.",
      "confidence": "high",
      "category": "open",
      "source": "semantic"
    },
    {
      "head": "RMSProp",
      "head_type": "Method",
      "relation": "HAS_LIMITATION",
      "tail": "lacks bias-correction term",
      "tail_type": "Limitation",
      "evidence": "RMSProp also lacks a bias-correction term; this matters most in case of a value of beta2 close to 1 (required in case of sparse gradients), since in that case not correcting the bias leads to very large stepsizes and often divergence, as we also empirically demonstrate in section 6.4.",
      "confidence": "high",
      "category": "open",
      "source": "semantic"
    },
    {
      "head": "Adam",
      "head_type": "Method",
      "relation": "DIFFERS_FROM",
      "tail": "RMSProp",
      "tail_type": "Method",
      "evidence": "There are a few important differences between RMSProp with momentum and Adam: RMSProp with momentum generates its parameter updates using a momentum on the rescaled gradient, whereas Adam updates are directly estimated using a running average of first and second moment of the gradient.",
      "confidence": "high",
      "category": "open",
      "source": "semantic"
    },
    {
      "head": "Adam",
      "head_type": "Method",
      "relation": "USES_TECHNIQUE",
      "tail": "preconditioner that adapts to the geometry of the data",
      "tail_type": "Technique",
      "evidence": "Adam employs a preconditioner that adapts to the geometry of the data, since v_hat_t is an approximation to the diagonal of the Fisher information matrix (Pascanu & Bengio, 2013);",
      "confidence": "high",
      "category": "open",
      "source": "semantic"
    },
    {
      "head": "Adam",
      "head_type": "Method",
      "relation": "DERIVED_FROM",
      "tail": "natural gradient descent",
      "tail_type": "Method",
      "evidence": "Like natural gradient descent (NGD) (Amari, 1998), Adam employs a preconditioner that adapts to the geometry of the data, since v_hat_t is an approximation to the diagonal of the Fisher information matrix (Pascanu & Bengio, 2013);",
      "confidence": "high",
      "category": "open",
      "source": "semantic"
    },
    {
      "head": "AdaGrad",
      "head_type": "Method",
      "relation": "BUILDS_ON",
      "tail": "Adam",
      "tail_type": "Method",
      "evidence": "AdaGrad corresponds to a version of Adam with beta1=0, infinitesimal (1-beta^2) and a replacement of alpha by an annealed version alpha_t=alpha*t^{-1/2}",
      "confidence": "high",
      "category": "controlled",
      "source": "semantic"
    },
    {
      "head": "AdaGrad",
      "head_type": "Method",
      "relation": "ALTERNATIVE_TO",
      "tail": "RMSProp",
      "tail_type": "Method",
      "evidence": "Note that this direct correspondence between Adam and Adagrad does not hold when removing the bias-correction terms; without bias correction, like in RMSProp, a beta2 infinitesimally close to 1 would lead to infinitely large bias",
      "confidence": "high",
      "category": "controlled",
      "source": "semantic"
    },
    {
      "head": "RMSProp",
      "head_type": "Method",
      "relation": "HAS_LIMITATION",
      "tail": "Infinitely Large Bias",
      "tail_type": "Limitation",
      "evidence": "without bias correction, like in RMSProp, a beta2 infinitesimally close to 1 would lead to infinitely large bias",
      "confidence": "high",
      "category": "open",
      "source": "semantic"
    },
    {
      "head": "Adam",
      "head_type": "Method",
      "relation": "REQUIRES",
      "tail": "Bias-Corrected Moment Estimates",
      "tail_type": "Domain_Term",
      "evidence": "Note that this direct correspondence between Adam and Adagrad does not hold when removing the bias-correction terms",
      "confidence": "high",
      "category": "open",
      "source": "semantic"
    },
    {
      "head": "AdaGrad",
      "head_type": "Method",
      "relation": "BUILDS_ON",
      "tail": "Stochastic Optimization",
      "tail_type": "Task",
      "evidence": "Adagrad can efficiently deal with sparse features and gradients as one of its main theoretical results whereas SGD is low at learning rare features.",
      "confidence": "high",
      "category": "controlled",
      "source": "semantic"
    },
    {
      "head": "Adam",
      "head_type": "Method",
      "relation": "BUILDS_ON",
      "tail": "Stochastic Optimization",
      "tail_type": "Task",
      "evidence": "Adam with $1/\\sqrt{t}$ decay on its stepsize should theoretically match the performance of Adagrad.",
      "confidence": "high",
      "category": "controlled",
      "source": "semantic"
    },
    {
      "head": "Adam",
      "head_type": "Method",
      "relation": "USES_TECHNIQUE",
      "tail": "1/sqrt(t) stepsize decay",
      "tail_type": "Technique",
      "evidence": "Adam with $1/\\sqrt{t}$ decay on its stepsize should theoretically match the performance of Adagrad.",
      "confidence": "high",
      "category": "open",
      "source": "semantic"
    },
    {
      "head": "Adam",
      "head_type": "Method",
      "relation": "CONSISTS_OF",
      "tail": "Bias-Corrected Moment Estimates",
      "tail_type": "Domain_Term",
      "evidence": "Similar to Adagrad, Adam can take advantage of sparse features and obtain faster convergence rate than normal SGD with momentum.",
      "confidence": "medium",
      "category": "open",
      "source": "semantic"
    },
    {
      "head": "SFO",
      "head_type": "Method",
      "relation": "REQUIRES",
      "tail": "deterministic subfunctions",
      "tail_type": "Requirement",
      "evidence": "SFO assumes deterministic subfunctions, and indeed failed to converge on cost functions with stochastic regularization.",
      "confidence": "high",
      "category": "open",
      "source": "semantic"
    },
    {
      "head": "SFO",
      "head_type": "Method",
      "relation": "HAS_LIMITATION",
      "tail": "incompatibility with stochastic regularization",
      "tail_type": "Limitation",
      "evidence": "SFO assumes deterministic subfunctions, and indeed failed to converge on cost functions with stochastic regularization.",
      "confidence": "high",
      "category": "open",
      "source": "semantic"
    },
    {
      "head": "Adam",
      "head_type": "Method",
      "relation": "USES_TECHNIQUE",
      "tail": "dropout noise",
      "tail_type": "Technique",
      "evidence": "Adam shows better convergence than other methods on multi-layer neural networks trained with dropout noise.",
      "confidence": "medium",
      "category": "open",
      "source": "semantic"
    },
    {
      "head": "Adagrad",
      "head_type": "Method",
      "relation": "HAS_LIMITATION",
      "tail": "vanishing second moment estimate in CNNs",
      "tail_type": "Limitation",
      "evidence": "The second moment estimate $\\widehat{v}_{t}$ vanishes to zeros after a few epochs and is dominated by the $\\epsilon$ in algorithm 1.",
      "confidence": "high",
      "category": "open",
      "source": "semantic"
    },
    {
      "head": "Adam",
      "head_type": "Method",
      "relation": "REQUIRES",
      "tail": "bias correction term for robustness to sparse gradients",
      "tail_type": "Requirement",
      "evidence": "Values of $\\beta_{2}$ close to 1, required for robustness to sparse gradients, results in larger initialization bias; therefore we expect the bias correction term is important in such cases of slow decay, preventing an adverse effect on optimization.",
      "confidence": "high",
      "category": "open",
      "source": "semantic"
    },
    {
      "head": "Adam",
      "head_type": "Method",
      "relation": "COMBINES",
      "tail": "first moment variance reduction with second moment estimation",
      "tail_type": "Technique",
      "evidence": "Whereas, reducing the minibatch variance through the first moment is more important in CNNs and contributes to the speed-up.",
      "confidence": "medium",
      "category": "open",
      "source": "semantic"
    },
    {
      "head": "Adam",
      "head_type": "Method",
      "relation": "DERIVED_FROM",
      "tail": "RMSProp",
      "tail_type": "Method",
      "evidence": "Discussed in section 5, removal of the bias correction terms results in a version of RMSProp (Tieleman & Hinton, 2012) with momentum.",
      "confidence": "high",
      "category": "open",
      "source": "semantic"
    },
    {
      "head": "Adam",
      "head_type": "Method",
      "relation": "USES_TECHNIQUE",
      "tail": "bias correction in moment estimates",
      "tail_type": "Technique",
      "evidence": "Values of $\\beta_{2}$ close to 1, required for robustness to sparse gradients, results in larger initialization bias; therefore we expect the bias correction term is important in such cases of slow decay, preventing an adverse effect on optimization.",
      "confidence": "high",
      "category": "open",
      "source": "semantic"
    },
    {
      "head": "AdaMax",
      "head_type": "Method",
      "relation": "BUILDS_ON",
      "tail": "Adam",
      "tail_type": "Method",
      "evidence": "AdaMax, a variant of Adam based on the infinity norm.",
      "confidence": "high",
      "category": "controlled",
      "source": "semantic"
    },
    {
      "head": "AdaMax",
      "head_type": "Method",
      "relation": "USES_TECHNIQUE",
      "tail": "infinity norm",
      "tail_type": "Technique",
      "evidence": "AdaMax, a variant of Adam based on the infinity norm.",
      "confidence": "high",
      "category": "open",
      "source": "semantic"
    },
    {
      "head": "Adam",
      "head_type": "Method",
      "relation": "USES_TECHNIQUE",
      "tail": "L2 norm",
      "tail_type": "Technique",
      "evidence": "In Adam, the update rule for individual weights is to scale their gradients inversely proportional to a (scaled) $L^{2}$ norm of their individual current and past gradients.",
      "confidence": "high",
      "category": "open",
      "source": "semantic"
    },
    {
      "head": "AdaMax",
      "head_type": "Method",
      "relation": "REQUIRES",
      "tail": "bias-correction term",
      "tail_type": "Technique",
      "evidence": "($\\alpha/(1-\\beta_{1}^{t})$) is the learning rate with the bias-correction term for the first moment.",
      "confidence": "high",
      "category": "open",
      "source": "semantic"
    },
    {
      "head": "Adam",
      "head_type": "Method",
      "relation": "REQUIRES",
      "tail": "bias-correction term",
      "tail_type": "Technique",
      "evidence": "The best results were achieved with small values of $(1-\\beta_{2})$ and bias correction;",
      "confidence": "high",
      "category": "open",
      "source": "semantic"
    },
    {
      "head": "Adam",
      "head_type": "Method",
      "relation": "HAS_LIMITATION",
      "tail": "instability without bias correction",
      "tail_type": "Limitation",
      "evidence": "values $\\beta_{2}$ close to 1 indeed lead to instabilities in training when no bias correction term was present",
      "confidence": "high",
      "category": "open",
      "source": "semantic"
    },
    {
      "head": "AdaMax",
      "head_type": "Method",
      "relation": "HAS_LIMITATION",
      "tail": "Initialization bias",
      "tail_type": "Limitation",
      "evidence": "Initialization bias can again be corrected by the estimator",
      "confidence": "medium",
      "category": "open",
      "source": "semantic"
    },
    {
      "head": "AdaMax",
      "head_type": "Method",
      "relation": "USES_TECHNIQUE",
      "tail": "exponential moving average",
      "tail_type": "Technique",
      "evidence": "an exponential moving average over the parameters can be used",
      "confidence": "high",
      "category": "open",
      "source": "semantic"
    },
    {
      "head": "AdaMax",
      "head_type": "Method",
      "relation": "REQUIRES",
      "tail": "parameter initialization",
      "tail_type": "Requirement",
      "evidence": "with initial value $u_{0} = 0$",
      "confidence": "medium",
      "category": "open",
      "source": "semantic"
    },
    {
      "head": "AdaMax",
      "head_type": "Method",
      "relation": "DERIVED_FROM",
      "tail": "Adam",
      "tail_type": "Method",
      "evidence": "the magnitude of parameter updates has a simpler bound with AdaMax than Adam",
      "confidence": "high",
      "category": "open",
      "source": "semantic"
    },
    {
      "head": "AdaMax",
      "head_type": "Method",
      "relation": "USES_TECHNIQUE",
      "tail": "recursive formula",
      "tail_type": "Technique",
      "evidence": "Which corresponds to the remarkably simple recursive formula",
      "confidence": "high",
      "category": "open",
      "source": "semantic"
    },
    {
      "head": "AdaMax",
      "head_type": "Method",
      "relation": "HAS_LIMITATION",
      "tail": "noisy last iterate",
      "tail_type": "Limitation",
      "evidence": "Since the last iterate is noisy due to stochastic approximation",
      "confidence": "medium",
      "category": "open",
      "source": "semantic"
    },
    {
      "head": "AdaMax",
      "head_type": "Method",
      "relation": "USES_TECHNIQUE",
      "tail": "Polyak-Ruppert averaging",
      "tail_type": "Technique",
      "evidence": "Previously in Moulines & Bach (2011), Polyak-Ruppert averaging has been shown to improve the convergence of standard SGD",
      "confidence": "medium",
      "category": "open",
      "source": "semantic"
    },
    {
      "head": "AdaMax",
      "head_type": "Method",
      "relation": "USES_TECHNIQUE",
      "tail": "exponential moving average estimator",
      "tail_type": "Technique",
      "evidence": "Initialization bias can again be corrected by the estimator $\\widehat{\\theta}_{t} = \\widehat{\\theta}_{t} / (1 - \\beta_{2}^{t})$",
      "confidence": "medium",
      "category": "open",
      "source": "semantic"
    },
    {
      "head": "Adam",
      "head_type": "Method",
      "relation": "APPLIED_TO",
      "tail": "non-convex optimization problems",
      "tail_type": "Task",
      "evidence": "we found Adam to be robust and well-suited to a wide range of non-convex optimization problems in the field machine learning",
      "confidence": "high",
      "category": "controlled",
      "source": "semantic"
    },
    {
      "head": "Adam",
      "head_type": "Method",
      "relation": "REQUIRES",
      "tail": "little memory",
      "tail_type": "Requirement",
      "evidence": "The method is straightforward to implement and requires little memory",
      "confidence": "high",
      "category": "open",
      "source": "semantic"
    },
    {
      "head": "Adam",
      "head_type": "Method",
      "relation": "REQUIRES",
      "tail": "computationally efficient algorithm",
      "tail_type": "Requirement",
      "evidence": "We have introduced a simple and computationally efficient algorithm for gradient-based optimization of stochastic objective functions",
      "confidence": "medium",
      "category": "open",
      "source": "semantic"
    }
  ],
  "entity_aliases": {
    "adaptive moment estimation": "Adam",
    "Adam optimizer": "Adam",
    "Adaptive Gradient Algorithm": "AdaGrad",
    "Root Mean Square Propagation": "RMSProp",
    "Adaptive Moment Estimation with Infinity Norm": "AdaMax",
    "stochastic gradient descent": "Stochastic Optimization",
    "adaptive optimization": "Stochastic Optimization",
    "Regret": "Regret Bound",
    "SNR": "Signal-to-Noise Ratio (SNR)",
    "Bias Correction": "Bias-Corrected Moment Estimates"
  }
}
\end{lstlisting}

\end{document}